%% file: main.tex
\documentclass[acmsmall]{acmart}
\usepackage{multirow}
\usepackage{booktabs}
\usepackage{adjustbox}
\usepackage{listings}
\renewcommand{\refname}{{}}
\usepackage{graphicx}
\usepackage{tabularx}
\usepackage{soul}

\usepackage{IEEEtrantools}
\usepackage{graphicx}
\usepackage{amsmath}
\usepackage{amsthm}
\usepackage{booktabs}
\usepackage{algorithm}
\usepackage{algorithmic}
\usepackage{multirow}
\usepackage{boldline}
\usepackage{hhline}
\usepackage{xcolor}
\usepackage{pifont}
\usepackage{makecell}
\usepackage{utfsym}
\usepackage[normalem]{ulem}
\useunder{\uline}{\ul}{}
\usepackage[inline]{enumitem}
\usepackage{enumitem}
\usepackage{IEEEtrantools}
\usepackage{stmaryrd}
\usepackage[edges]{forest}
\usepackage{soul}
\sethlcolor{red}
\usepackage{color,xcolor}
\usepackage{tikz}
\usepackage{subcaption}
\usepackage{utfsym}
\usepackage[normalem]{ulem}
\useunder{\uline}{\ul}{}
\usepackage[inline]{enumitem}
\usepackage{enumitem}
\usepackage{IEEEtrantools}
\usepackage{stmaryrd}
\usepackage{array}

\definecolor{hidden-red}{RGB}{205, 44, 36}
\definecolor{hidden-blue}{RGB}{194,232,247}
\definecolor{hidden-orange}{RGB}{243,202,120}
\definecolor{hidden-green}{RGB}{34,139,34}
\definecolor{hidden-pink}{RGB}{255,245,247}
\definecolor{hidden-black}{RGB}{20,68,106}

\newcommand{\cmark}{\ding{51}}
\newcommand{\xmark}{\ding{55}}

\newcommand{\highlight}[1]{\textcolor{black}{#1}}
\newcommand\encircle[2][]{\tikz[overlay]\node[fill=blue!20,inner sep=2pt, anchor=text, rectangle, rounded corners=1.5mm,#1] {#2};\phantom{#2}}

\definecolor{hidden-red}{RGB}{205, 44, 36}
\definecolor{hidden-blue}{RGB}{194,232,247}
\definecolor{hidden-orange}{RGB}{243,202,120}
\definecolor{hidden-green}{RGB}{34,139,34}
\definecolor{hidden-pink}{RGB}{255,245,247}
\definecolor{hidden-black}{RGB}{20,68,106}

\definecolor{myGreen}{RGB}{127,210,85}
\definecolor{myOrange}{RGB}{242,154,66}
\definecolor{myYellow}{RGB}{247,223,65}
\definecolor{myRed}{RGB}{232,80,43}
\definecolor{myViolet}{RGB}{162,57,102}
\definecolor{myBlue}{HTML}{4686f3}
\definecolor{myYellowv2}{HTML}{E6C802}
\definecolor{myOrangev2}{HTML}{ED8E55}
\definecolor{MyGreenv2}{HTML}{009B55}
\definecolor{MyRedv2}{HTML}{c22f2f}
\definecolor{DarkRed}{RGB}{130,25,0}
\definecolor{PurpleRed}{RGB}{204,0,102}
\definecolor{DarkGreen}{RGB}{30,130,30}
\definecolor{DarkBlue}{RGB}{0,0,250}
\definecolor{DarkYellow}{RGB}{255,128,0}

\AtBeginDocument{%
  }

\setcopyright{acmlicensed}
\copyrightyear{2024}
\acmYear{2024}
\acmDOI{XXXXXXX.XXXXXXX}

\acmJournal{JACM}
\acmVolume{37}
\acmNumber{4}
\acmArticle{111}
\acmMonth{8}

\begin{document}

\title{Mathematical Language Models: A Survey}




\author{Wentao Liu}
\authornote{Both authors contributed equally to this research.}
\author{Hanglei Hu}
\authornotemark[1]

\author{Jie Zhou}
\authornote{Corresponding authors.}
\email{jzhou@cs.ecnu.edu.cn}

\author{Yuyang Ding}

\author{Junsong Li}

\author{Jiayi Zeng}

\author{Mengliang He}

\author{Qin Chen}
\affiliation{%
  \institution{School of Computer Science and Technology, East China Normal University}
  \city{Shanghai}
  \country{China}
}

\author{Bo Jiang}
\affiliation{%
  \institution{Lab of Artificial Intelligence for Education, East China Normal University}
  \city{Shanghai}
  \country{China}
}

\author{Aimin Zhou}
\authornotemark[2]
\email{amzhou@cs.ecnu.edu.cn}

\author{Liang He}
\affiliation{%
  \institution{School of Computer Science and Technology, East China Normal University}
  \city{Shanghai}
  \country{China}
}
\affiliation{%
  \institution{Lab of Artificial Intelligence for Education, East China Normal University}
  \city{Shanghai}
  \country{China}
}

\renewcommand{\shortauthors}{Liu et al.}

\begin{abstract}
In recent years, there has been remarkable progress in leveraging Language Models (LMs), encompassing Pre-trained Language Models (PLMs) and Large-scale Language Models (LLMs), within the domain of mathematics. This paper conducts a comprehensive survey of mathematical LMs, systematically categorizing pivotal research endeavors from two distinct perspectives: tasks and methodologies. 
The landscape reveals a large number of proposed mathematical LLMs, which are further delineated into instruction learning, tool-based methods, fundamental CoT techniques, \highlight{advanced CoT methodologies and multi-modal methods.}
\highlight{To comprehend the benefits of mathematical LMs more thoroughly, we carry out an in-depth contrast of their characteristics and performance.}
In addition, our survey entails the compilation of over 60 mathematical datasets, including training datasets, benchmark datasets, and augmented datasets. 
Addressing the primary challenges and delineating future trajectories within the field of mathematical LMs, \highlight{this survey is poised to facilitate} and inspire future innovation among researchers invested in advancing this domain.
\end{abstract}



\received{20 February 2007}
\received[revised]{12 March 2009}
\received[accepted]{5 June 2009}

\maketitle

\input{secs/01Introduction}

\input{secs/02Tasks}
\input{secs/03PLMs}

\input{secs/04LLMs}

\input{secs/05Datasets}
\input{secs/06Disscusion}

\input{secs/07Challenges}

\section{Conclusions}
\label{Conclusions}
The survey elucidates the pivotal role of mathematical LMs in reshaping the landscape of mathematical problem-solving, leveraging a spectrum of models, \highlight{from PLMs to LLMs}, to address diverse mathematical tasks. 
Our taxonomical delineation of mathematical tasks and methods provides a systematic framework for comprehending the intricacies of LMs-based methodologies, distinguishing between arithmetic calculation, mathematical reasoning, and various algorithmic approaches employed in these models.
\highlight{
We also outline the features and effectiveness of current mathematical LMs to gain a deeper understanding of these techniques.
We gather over 60 diverse mathematical datasets, carefully organized into training, benchmark, and augmented categories, highlighting how crucial data is in pushing forward mathematical research and supporting insightful inquiry in various mathematical fields. Additionally, by thoughtfully tackling challenges like faithfulness, multi-modality, uncertainty, evaluation, theorem creation, application, and data scarcity, this survey opens doors for future studies aimed at enhancing the capabilities of mathematical LMs. By exploring the current advancements, obstacles, and paths for future research, we see this thorough overview as a key resource for fostering innovation and guiding the future of mathematical LMs research, all while contributing to the exciting developments in mathematics and artificial intelligence.}

\begin{acks}
The authors wish to thank the reviewers for their helpful comments and suggestions.
This research is funded by the National Science and Technology Major Project (No. 2021ZD0114002), the National Nature Science Foundation of China (No. 62477010, No.62307028), the Science and Technology Commission of Shanghai Municipality Grant (No. 22511105901, No. 21511100402), Shanghai Science and Technology Innovation Action Plan (No. 24YF2710100) and Shanghai Special Project to Promote High-quality Industrial Development (No. RZ-CYAI-01-24-0288).
\end{acks}

\bibliographystyle{ACM-Reference-Format}
\bibliography{References}


\end{document}

%% file: secs/01Introduction.tex
\section{Introduction}
Mathematics is the queen of sciences.

\rightline{-- Carl Friedrich Gauss}

Mathematics stands as a foundational skill integral to human intelligence, wielding significance across diverse fields such as natural sciences, engineering, medicine, finance, computer science, and social sciences. 
Within the field of natural language processing (NLP), the development of computer models aimed at autonomously resolving mathematical word problems has captivated researchers since as early as 1963 \cite{feigenbaum1963computers,bobrow1964natural,briars1984integrated,fletcher1985understanding}. 
We believe that addressing this problem is a potential way toward illuminating pathways for general reasoning mechanisms, consequently advancing the pursuit of \highlight{artificial general intelligence (AGI)}.

One main traditional solution to math word problems is statistical learning-based methodologies \cite{hosseini-etal-2014-learning,kushman-etal-2014-learning,zhou2015learn,mitra2016learning}. Notably, machine learning techniques \cite{kushman-etal-2014-learning,roy2015reasoning,roy2015solving} alongside semantic parsing methods \cite{shi2015automatically,koncel2015parsing} have been employed to address this challenge, showcasing promising outcomes on certain datasets.
The evolution of deep learning has prompted considerable interest in crafting neural networks capable of resolving mathematical problems \cite{wang2017deep,couperus2023large}. 
Notably, recent years have witnessed significant advancements in the domain of mathematical AI \cite{wei2022chain,matzakos2023learning}, propelled by the emergence of powerful language models (LMs) \cite{kenton2019bert,touvron2023llama}. 
These language models, comprising pre-trained language models (PLMs) and large-scale language models (LLMs), have assumed a central role in reshaping the landscape of mathematical exploration and practical applications. 
The paramount focus of this comprehensive survey lies in assessing the impact of these models on the field of mathematics. The survey endeavors to provide a thorough overview of extant research and its consequential implications.

PLMs such as BERT \cite{devlinBERTPretrainingDeep2019}, RoBERTa \cite{liuRoBERTaRobustlyOptimized2019}, BART \cite{lewis2020bart}, GPT-1 \cite{radford_gpt-1_2018} and GPT-2  \cite{radford_gpt-2_2019} undergo pre-training on extensive textual corpora to assimilate worldly knowledge. 
To enhance mathematical performance, certain endeavors focus on either pre-training or fine-tuning PLMs using mathematical datasets \cite{clarkRegentsScienceExams2021,gevaInjectingNumericalReasoning2020,fengInjectingNumericalReasoning2022}.
For instance, methodologies like GenBERT \cite{gevaInjectingNumericalReasoning2020}, NF-NSM \cite{fengInjectingNumericalReasoning2022}, MathBERT\cite{pengMathBERTPretrainedModel2021} and LISA \cite{jiang2021lisa} incorporate numerical data or mathematical formulas into PLMs to augment their capabilities. Moreover, specialized modules or tailored loss functions are devised, leveraging existing PLMs to learn mathematical reasoning and operations \cite{zhu2021tat,zhao2022multihiertt,jieLearningReasonDeductively2022,liSeekingPatternsNot2022}.

The recent advent of LLMs, exemplified by OpenAI's GPT-4 \cite{openai2023gpt4}, has catalyzed an unforeseen surge in innovation, underscoring the multifaceted potential of AI within the domain of mathematics. 
These models (\cite{touvron2023llama,touvron2023llama2}) have demonstrated remarkable success across diverse Natural Language Processing (NLP) tasks, leveraging in-context learning \cite{min2022metaicl,brown2020language,chen2022meta} and instruction learning \cite{yang2023gpt,liu_goat_2023}.
Recent studies by Wang et al. \cite{wang2022self} indicate that LLMs featuring over 100 billion parameters (e.g., GPT-3 \cite{brown2020language} with 175 billion, PaLM \cite{chowdhery2022palm} with 540 billion) exhibit the capability to address intricate tasks by employing a chain-of-thought (CoT) mechanism \cite{wei2022chain} when furnished with a limited set of reasoning examples as demonstrations. 
Advancements in CoT frameworks \cite{chen2022program,lu2022dynamic,zhang2023automatic,fu2022complexity} have been tailored to enhance mathematical performance, incorporating tools and programs \cite{gao2023pal,drori2022neural,he2023solving}.
\highlight{Moreover, openly accessible LLMs are specifically designed for mathematical tasks, such as LLEMMA \cite{azerbayev8llemma}, Qwen-Math \cite{yang2024qwen2} and InternLM-Math \cite{ying2024internlmmath}. 
Recently, o1 \cite{openai_o1_system_card} and o3 obtained state-of-the-art performance on mathematical reasoning by integrating reinforcement learning with the Monte Carlo tree.
}

We mainly review the most relevant surveys about mathematical reasoning and LMs \cite{10.1145/3605943}. \highlight{For surveys focused on mathematical reasoning,} Lu et al. \cite{lu-etal-2023-survey} delineate the landscape of deep learning applications specifically pertaining to mathematical reasoning \highlight{(DL4Math)}. Qiao et al. \cite{qiao-etal-2023-reasoning} predominantly delve \highlight{into general reasoning mechanisms} centered around language models, encompassing arithmetic, commonsense, logical, symbolic, and multi-modal reasoning \highlight{(LM4Reasoning)}. Additionally, Chu et al. \cite{chuCoTReasoningSurvey2023} offer an exploration of studies focused on \highlight{CoT} reasoning. \highlight{For surveys related to LMs,} Qiu et al. \cite{qiu2020pre} and Zhao et al. \cite{zhao2023survey} contribute comprehensive reviews elucidating the realm of existing PLMs and LLMs, respectively. \highlight{Yan et al. \cite{yan2024survey} summary the related studies about multimodal LLMs-based mathematical reasoning (MMLM4Math).}


\highlight{In contrast to these existing works, our study shifts focus to the emerging field of mathematical language models (MLMs), presenting a systematic and in-depth examination of the novel methods, tasks, and datasets that have emerged specifically within this domain. While previous surveys primarily address general reasoning capabilities of LMs or broader trends in LMs, our work distinguishes itself by offering a detailed exploration of how LMs are being uniquely applied to advanced mathematical tasks. We not only discuss the successes and challenges of existing approaches but also identify critical gaps and emerging trends that are shaping the future of mathematical reasoning.
By providing a comprehensive overview of mathematical language models, including their applications, challenges, and future directions, our work aims to inspire further research and innovation in this rapidly developing area. We believe that highlighting the intersections between mathematics and LMs will encourage new approaches for solving complex mathematical problems, and ultimately, contribute to the broader goal of leveraging language models to revolutionize mathematics and its real-world applications.}

Specifically, our categorization of mathematical tasks (\S\ref{Tasks}) contain two primary domains: mathematical calculation (\S\ref{Mathematical Calculation}), consisting of arithmetic representation and arithmetic calculation, and mathematical reasoning (\S\ref{Mathematical Reasoning}), consisting of \highlight{math} problem-solving and theorem proving. Furthermore, our taxonomy of existing algorithms segregates them into PLMs-based approaches (\S\ref{PLMs}), including autoregression (\S\ref{Autoregression LMs}) and non-autoregression (\S\ref{Non-Autoregression LMs}) LMs, and LLMs-based methodologies (\S\ref{LLMs}), which include instruction learning (\S\ref{Instruction Learning}), tool-based strategies (\S\ref{Tool-based Methods}), fundamental CoT techniques (\S\ref{Fundamental CoT Methods}), \highlight{advanced CoT methodologies (\S\ref{Advanced CoT Methods}) and multi-modal methods (\S\ref{Multi-modal Methods}).}
Moreover, we provide a comprehensive compilation of over 60 mathematical datasets (\S\ref{Datasets}), systematically categorized into training (\S\ref{Training Datasets}), benchmark (\S\ref{Benchmark Datasets}), and augmented datasets (\S\ref{Augmented Datasets}). 
This classification aims to facilitate research by delineating the utility and applicability of these datasets within distinct research contexts. 
\highlight{Also, we analyze and discuss the characteristics and performance of various mathematical LMs (\S\ref{Analysis and Discussion}).}
Lastly, our survey meticulously addresses the primary challenges and explores the future directions of this domain (\S\ref{Challenges and Further Directions}), including faithfulness, multi-modal, uncertainty, evaluation, creation, application, and data scarcity. In conclusion (\S\ref{Conclusions}), we believe that this comprehensive survey will serve as a valuable asset for researchers and practitioners alike, seeking to push the boundaries of innovation in the field of mathematical language models.

%% file: secs/02Tasks.tex
\section{Mathematical Tasks}
\label{Tasks}
\begin{figure}[t]
\begin{center}
\includegraphics[width=0.85\textwidth]{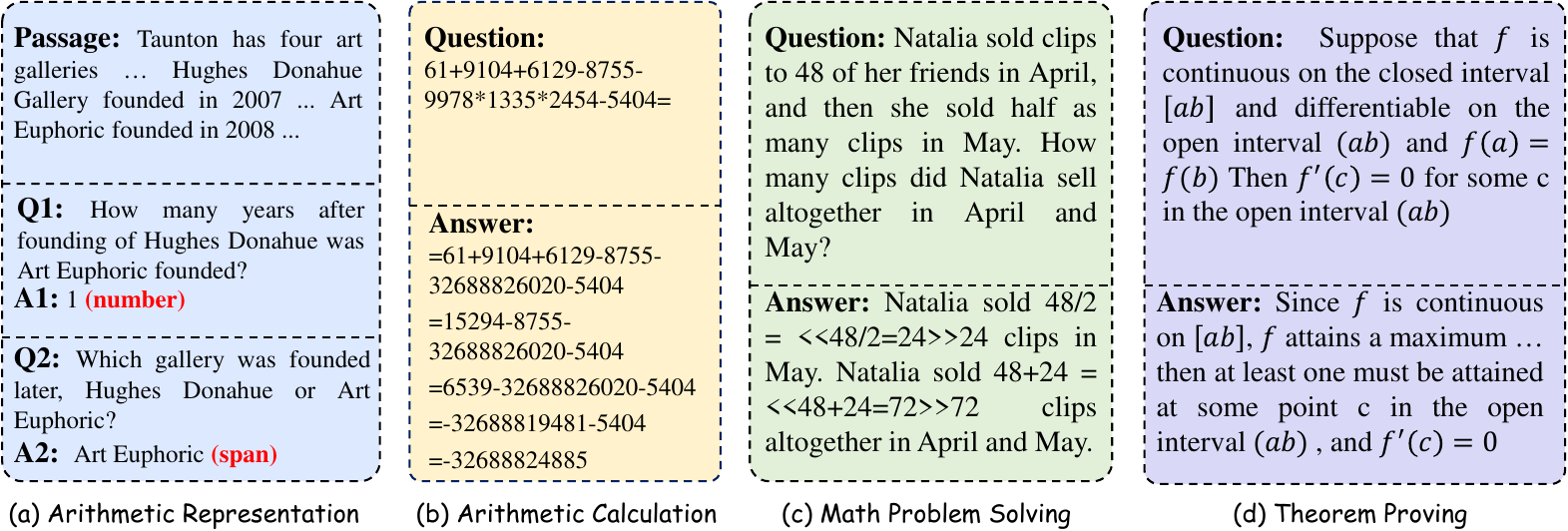}
\end{center}
\vspace{-3mm}
\caption{\highlight{Examples of mathematical tasks, where (a) illustrates responses with different representation types.}}
\label{fig:Tasks}
\vspace{-5mm}
\end{figure}

In this section, we summarize the existing mathematical tasks into mathematical calculation and mathematical reasoning (Figure \ref{fig:task_taxonomy}, \highlight{\ref{fig:Tasks}}). 

\input{imgs/taxonomy_Tasks}

\subsection{Mathematical Calculation}
\label{Mathematical Calculation}
The advent of \highlight{LMs} has ushered in a new era of exploration into their computational capabilities, particularly in arithmetic. Initially, these models demonstrated basic computational abilities by representing numbers in textual formats. With the increasing capabilities of LMs, it has been observed that they can acquire arithmetic skills through methods such as fine-tuning, even without carefully crafted numeric representations.

\subsubsection{Arithmetic Representation}
\label{Arithmetic Representation}
In the early stages of the research, numerical values were either omitted or oversimplified, treated as ordinary text, or categorized as ``unknown". \highlight{These approaches}, however, proved inadequate for \highlight{mathematical} tasks. For instance, BERT performs five times worse when the answer is a number rather than a textual span \highlight{on the DROP benchmark} \cite{dua2019drop}.

Recent literature proposes various methods for numerical representation. 
Geva et al. \cite{gevaInjectingNumericalReasoning2020} introduce GenBERT, which tokenizes numbers at the digit level and undergoes fine-tuning with arithmetic word problems and simple arithmetic tasks.
Zhang et al. \cite{zhang2020language} experiment with a retrained BERT model by converting numbers into scientific notation (e.g., 314.1 as 3141[EXP]2).
Spithourakis and Riedel \cite{spithourakis2018numeracy}, along with Wallace et al. \cite{wallace2019nlp}, explore the integration of digit embeddings into a singular embedding that represents the entire number. Berg-Kirkpatrick and Spokoyny \cite{berg-kirkpatrickEmpiricalInvestigationContextualized2020} propose a method with digit-RNN and exponent embeddings. This method specifically emphasizes the exponent while disregarding the mantissa. Goat \cite{liu_goat_2023} introduces Consistent Tokenization, enhancing the relationships between similar numerical values.

\subsubsection{Arithmetic Calculation}
\label{Arithmetic Calculation}
There has been significant research into the arithmetic capabilities of LLMs. Nogueira et al. \cite{nogueira2021investigating} and Wang et al. \cite{wang2021exploring} assessed addition and subtraction tasks. Muffo et al. \cite{muffo_evaluating_nodate} evaluated two-digit multiplication. 
Yuan et al. \cite{yuan2023well} evaluated the arithmetic operation capabilities of different models, including GPT-4 \cite{openai2023gpt4}, Galactica \cite{taylor_galactica_2022}, and LLaMA \cite{touvron2023llama}.

Traditionally, it was presumed that LLMs could not accurately perform complex arithmetic, especially multiplication involving more than eight digits. However, recent approaches challenge this assumption. Zhou et al. \cite{zhouTeachingAlgorithmicReasoning2022} apply specialized prompt engineering to enhance addition capabilities but note limitations in multiplication beyond seven digits. Jelassi et al. \cite{jelassiLengthGeneralizationArithmetic2023} investigate length generalization in basic arithmetic tasks using techniques like relative position embeddings and training set priming. ScratchpadGPT \cite{nye_show_2021} demonstrates the effectiveness of pre-generating a Chain of Thought (CoT) before producing an answer in 8-digit addition tasks. Goat \cite{liu_goat_2023} utilizes supervised instruction for fine-tuning elementary arithmetic operations with large integers. MathGLM \cite{yang2023gpt} excels in intricate arithmetic tasks by pre-training on datasets that decompose complex arithmetic expressions into simpler steps. 

\subsection{Mathematical Reasoning}
\label{Mathematical Reasoning}


Mathematical reasoning is a pivotal aspect of artificial intelligence that facilitates the understanding and solving of complex mathematical problems. The integration of LLMs in this domain has been significant, owing to their ability to interpret, process, and generate complex natural text. This section delves into the current research and developments within this field, focusing on two main areas: math problem solving and theorem proving.

\subsubsection{Math Problem Solving}
\label{Math Problem Solving}
\highlight{Math problem solving} refers to the process of using algorithms, computational models, and increasingly LLMs to understand, explain, and solve mathematical problems. \highlight{These approaches} to solving problems spans all levels from basic arithmetic to advanced mathematics, including but not limited to algebra, geometry, statistics, and calculus. \highlight{Generally, mathematical problem solving can be categorized into two primary forms: the first involves contextualized problems, where the solver must comprehend the given scenario, translate it into a purely mathematical problem, and then proceed to solve it with solutions; the second form presents the solver with a direct mathematical problem, requiring only the provision of a specific answer.}




Math Word Problems (MWPs) refer to problems that present mathematical concepts and calculations in the form of written descriptions. Such questions usually include one or more story situations that describe a series of events or conditions from which the solver needs to extract relevant mathematical information and apply appropriate mathematical principles to solve the problem. Over the years, the use of various efficient and intelligent algorithms to solve MWPs has also been a research focus in the fields of computer science and artificial intelligence \cite{bobrow1964natural}. 
Recent research has highlighted the growing capabilities of LLMs in \highlight{MWPs task}, emphasizing the trend toward more nuanced and sophisticated AI-driven mathematical analysis. \highlight{MathPrompter \cite{imani2023mathprompter} uses a GPT3 DaVinci to solve MWPs with excellent results, demonstrating the potential of LLMs to not only explain but also generate complex mathematical reasoning. 
Yuan et al. \cite{yuan2023scaling} investigated the interaction between various factors such as pre-training loss, amount of supervised data, and augmented data to improve the mathematical inference performance of LLMs. 
They proposed Rejection sampling Fine-Tuning (RFT) to enhance performance by using supervised models to generate and collect correct reasoning paths, thereby achieving better inference generalization. 
}

MetaMath \cite{yu2023metamath} further extends the practicality of LLMs, proposing a paradigm by which LLMs generate their mathematical problems, thus creating a self-sustaining learning environment that encourages continuous improvement of the model's problem-solving acuity. 
Similarly, WizardMath \cite{luo2023wizardmath} explores new ways to enhance LLMs’ mathematical reasoning capabilities by reinforcing evolutionary instructions, indicating an important step towards the autonomous \highlight{self-improvement of models.} Instead, MathAttack \cite{zhou2023mathattack} introduces a critical perspective by exploring the sensitivity of LLMs to specialized adversarial input designed to test their mathematical problem-solving abilities. Such investigations are crucial to developing more resilient and reliable LLMs that can withstand and adapt to a variety of problem-solving situations. 
Furthermore, LLEMMA \cite{azerbayev8llemma}, an \highlight{open-source} LM dedicated to mathematics, outlines ongoing efforts to \highlight{make LLMs accessible to} a wide range of mathematical inquiries, from basic problem solving to advanced theorem proving. 

 

\highlight{Correspondingly, if the problems are non-contextualized and purely mathematical, the focus shifts more towards the Math Question-Answer (MQA) system's ability to understand and automatically solve the related computational tasks.} These questions can range from simple arithmetic problems to complex high-school or college-level mathematics, including algebra, calculus, and geometry. The challenge for MQA systems is to accurately interpret the text, convert it into an appropriate mathematical representation, perform the necessary computations, and generate a correct and often step-by-step solution, mimicking the problem-solving process of a human.

Sachan et al. \cite{saxton2018analysing} introduce a framework for evaluating neural architectures through a suite of math problems, highlighting the importance of algebraic generalization. This is complemented by GEOS \cite{seo-etal-2015-solving}, an innovative system that solves SAT geometry problems by integrating text understanding with diagram interpretation. The introduction of datasets and systems like Inter-GPS \cite{lu2021inter}, IconQA \cite{lu2021iconqa}, and PGDP5K \cite{Hao2022PGDP5KAD, Zhang2022, Zhang2023PGPS} further advances the field, offering new benchmarks for abstract diagram understanding and visual language reasoning. Additionally, Scharpf et al. \cite{scharpf2022mining} indicate that unsupervised methods for formula labeling contribute to the automated understanding of mathematical documents. Collectively, these works demonstrate the evolving landscape of MQA, where the synergy between natural language processing, computer vision, and symbolic reasoning opens new avenues for educational technology and AI research.


\subsubsection{Theorem Proving}
\label{Theorem Proving}
Theorem proving (TP) refers to the process of demonstrating that a statement is correct based on existing axioms and facts, which is a long-term challenge in AI \cite{10.1145/1455567.1455605}. In the context of LLMs, people began to try to use the huge knowledge base of LLMs and some manual guidance to solve this task. LLMs like GPT-3 \cite{brown2020language} can process natural text to understand the premises of the theorem, and apply certain logical reasoning rules to complete the proof requirements. This can then be done by calling relevant tools or using manual inspection to check the correctness of the process. The advanced approach aims to leverage LLM's superior computing power and vast knowledge base relative to humans to automate the theorem-proving process, a task that previously required professional mathematicians.

In the field of AI, research on theorem proving has progressed from data sets containing large numbers of human-written proofs (such as CoqGym \cite{yang2019learning}) to complex models that can autonomously generate proof strategies (such as ASTactic \cite{yang2019learning}). This improvement is reflected in the application of some language models that are based on Transformer \cite{vaswani2017attention} to TP tasks, especially in recent research such as GPT-f \cite{polu2020generative}. Most notably, some of the proofs generated by these models have been formally recognized by the mathematical community. Jiang et al. \cite{jiang2022draft} further refined this process by using informal proofs to guide TP models. NaturalProofs \cite{welleck2021naturalproofs} extends these capabilities by leveraging the language of natural mathematics to create a rich corpus of model training and evaluation. At the same time, DeepMath \cite{irving2016deepmath} and INT \cite{wu2020int} have promoted the development of this field by demonstrating the effectiveness of neural sequence models for premise selection and evaluating the LLMs' generalization ability in theorem proving.


Ongoing research is delving into the integration of \highlight{LMs} and interactive proof assistants within the realm of mathematics. This involves generating proofs through tactic prediction \cite{DBLP:journals/corr/abs-2009-03393,han2021proof,lample2022hypertree,jiang2022thor}, automating formalization processes \cite{DBLP:conf/nips/WuJLRSJS22,jiang2022draft}, and developing unified tools \cite{llmstep}. Due to the substantial computational demands of exploration, \highlight{LMs} applied in this domain have traditionally been limited in size. However, recent advancements have showcased potential in employing larger models \cite{first2023baldur,jiang2022draft}. LLEMMA \cite{azerbayev8llemma} demonstrates few-shot proof autoformalization and tactic prediction, offering a vast dataset of formal mathematics and an openly accessible model for further investigation in these directions. These endeavors establish crucial theoretical and practical frameworks for automating and enhancing tactic prediction.

The above series of progressive research work shows that utilizing the powerful computing power and extensive knowledge base of LLMs is the main development direction for future TP tasks. The more excellent models (such as GPT-4 \cite{openai2023gpt4}) obviously have better reasoning capabilities without any human intervention, but we also need to be clear about one thing, that is the inevitable hallucination problem in neural text generation models \cite{maynez-etal-2020-faithfulness}. The hallucination problem refers to the generated content that is nonsensical or unfaithful to the provided source content \cite{ji2023survey}. The occurrence of hallucinations at any step in the reasoning process will lead to errors in the final reasoning results. Therefore, solving the hallucination problem is also particularly important in future work related to TP.


%% file: imgs/taxonomy_Tasks.tex
\tikzstyle{my-box}=[
    rectangle,
    draw=hidden-black,
    rounded corners,
    text opacity=1,
    minimum height=1.5em,
    minimum width=5em,
    inner sep=2pt,
    align=center,
    fill opacity=.5,
]
\tikzstyle{leaf}=[
    my-box, 
    minimum height=1.5em,
    fill=hidden-blue!90, 
    text=black,
    align=left,
    font=\normalsize,
    inner xsep=2pt,
    inner ysep=4pt,
]
\begin{figure*}[t]
    \vspace{-2mm}
    \centering
    \resizebox{\textwidth}{!}{
        \begin{forest}
            forked edges,
            for tree={
                grow=east,
                reversed=true,
                anchor=base west,
                parent anchor=east,
                child anchor=west,
                base=left,
                font=\large,
                rectangle,
                draw=hidden-black,
                rounded corners,
                align=left,
                minimum width=4em,
                edge+={darkgray, line width=1pt},
                s sep=6pt,
                inner xsep=2pt,
                inner ysep=3pt,
                line width=0.8pt,
                ver/.style={rotate=90, child anchor=north, parent anchor=south, anchor=center},
            },
            where level=1{text width=7.6em,font=\normalsize,}{},
            where level=2{text width=9.4em,font=\normalsize,}{},
            where level=3{text width=9.5em,font=\normalsize,}{},
            where level=4{text width=12em,font=\normalsize,}{},
            [
                Mathematical Tasks, ver
                [
                    Mathematical \\ Calculation ~(\S\ref{Mathematical Calculation})
                    [
                        Arithmetic \\ Representation 
                        [
                            DigitRNN ~\cite{spithourakis2018numeracy}{,}
                            DigitCNN ~\cite{wallace2019nlp}{,}
                            GenBERT ~\cite{gevaInjectingNumericalReasoning2020}{,}
                            NumBERT ~\cite{zhangLanguageEmbeddingsCapture2020}
                            , leaf, text width=44.6em
                        ]
                    ]
                    [
                        Arithmetic \\ Calculation 
                        [
                            ScratchpadGPT ~\cite{nye_show_2021}{,}
                            Goat ~\cite{liu_goat_2023}{,}
                            MathGLM ~\cite{yang2023gpt}{,}
                            , leaf, text width=44.6em
                        ]
                    ]
                ]
                [
                    Mathematical \\ Reasoning ~(\S\ref{Mathematical Reasoning})
                    [
                        \highlight{Math Problem} \\ \highlight{Solving} 
                        [
                            \highlight{MathPrompter ~\cite{imani2023mathprompter}{,}
                            MetaMath ~\cite{yu2023metamath}{,}
                            WizardMath ~\cite{luo2023wizardmath}{,}
                            MathAttack ~\cite{zhou2023mathattack}{,}
                            LLEMMA ~\cite{azerbayev8llemma}{,}
                            GEOS ~\cite{seo-etal-2015-solving}{,}}\\
                            \highlight{DROP ~\cite{dua2019drop}{,}
                            Mathematics ~\cite{saxton2018analysing}{,}
                            Lila ~\cite{mishra2022lila}}
                            , leaf, text width=44.6em
                        ]
                    ]
                    [
                        Theorem \\ Proving 
                        [
                            DeepMath ~\cite{irving2016deepmath}{,}
                            ASTactic ~\cite{yang2019learning}{,}
                            NaturalProofs ~\cite{welleck2021naturalproofs}{,}
                            INT ~\cite{wu2020int}
                            , leaf, text width=44.6em
                        ]
                    ]
                ]    
            ]
        \end{forest}
    }
    \vspace{-5mm}
    \caption{Taxonomy of mathematical tasks.}
    \label{fig:task_taxonomy}
    \vspace{-4mm}
\end{figure*}
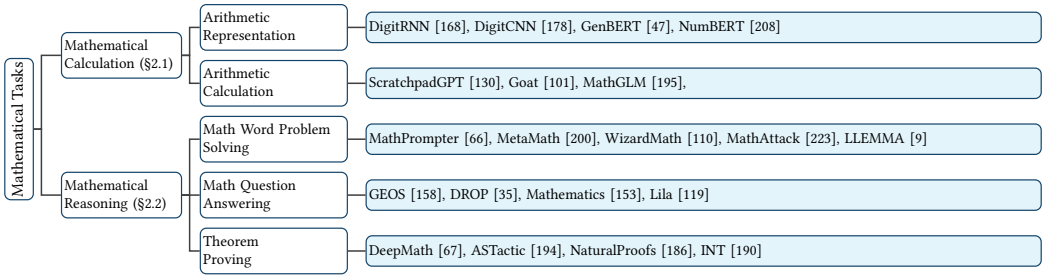

%% file: secs/03PLMs.tex
\section{PLMs-based Methods}
\label{PLMs}

Based on the transformer architecture \cite{vaswani2017attention} with self-attention mechanisms, the ``pre-training and fine-tuning" paradigm has revolutionized the field of natural language processing. Within the context of Pre-trained Language Models (PLMs), numerous approaches have been proposed to tackle text generation problems \cite{noorbakhsh2021pretrained}. Autoregressive LMs (ALMs, e.g., GPT-1 \cite{radford_gpt-1_2018} and T5 \cite{rael_exploring_nodate}) and Non-Autoregressive LMs (NALMs, e.g., BERT \cite{kenton2019bert} and Roberta \cite{liuRoBERTaRobustlyOptimized2019}) are the two primary strategies for mathematics (Figure \ref{fig:taxonomy}). 

\input{imgs/taxonomy_PLLMs}

\subsection{Autoregression LMs}
\label{Autoregression LMs}

\begin{figure}[t]
\begin{center}
\includegraphics[width=0.7\textwidth]{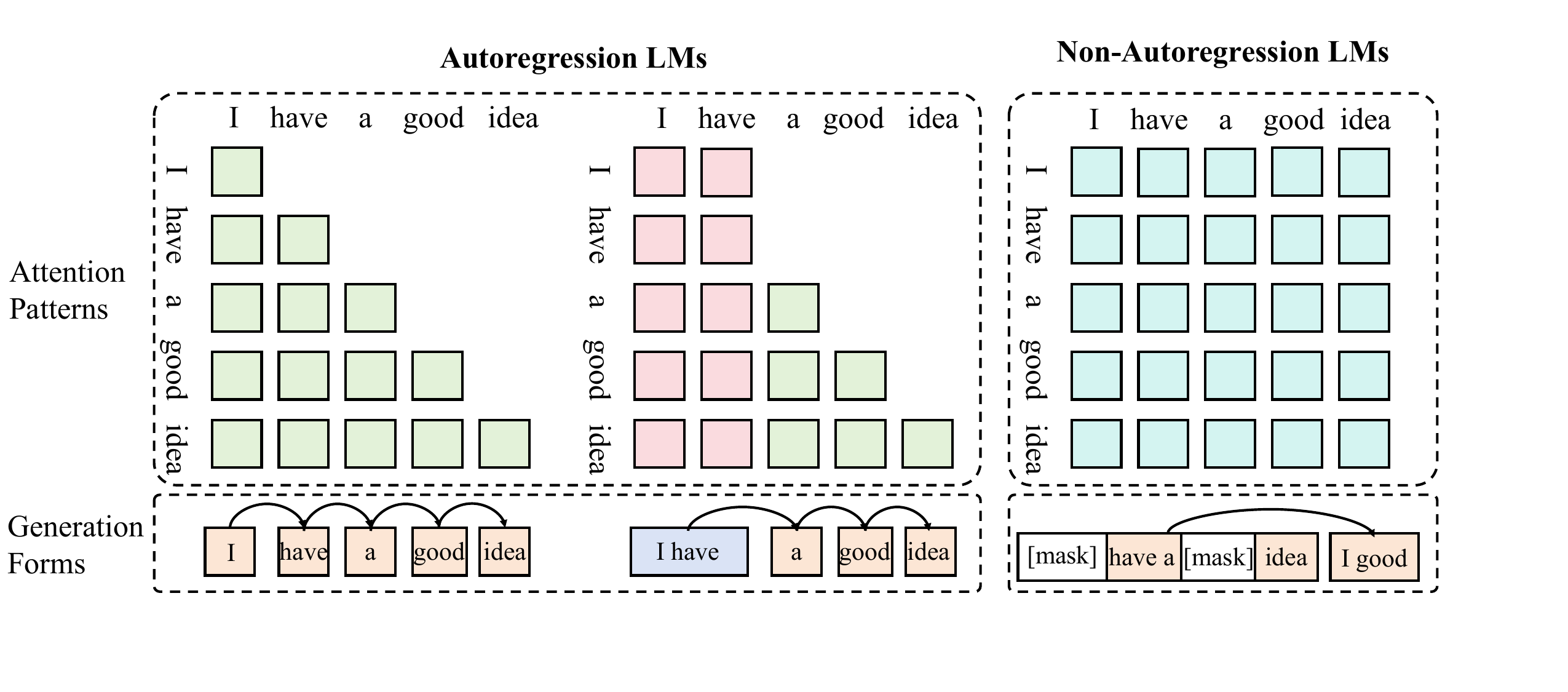}
\end{center}
\vspace{-3mm}
\caption{The distinctions between Autoregression LMs and Non-Autoregression LMs.}
\label{PLM_D}
\vspace{-5mm}
\end{figure}

The distinctions between \highlight{Autoregression LMs and Non-Autoregressive LMs} in sequence generation tasks are depicted in Figure \ref{PLM_D}. Furthermore, \highlight{Autoregression LMs} can be categorized into \highlight{two parts}: causal decoder (e.g., GPT-1 \cite{radford_gpt-1_2018} and GPT-2 \cite{radford_gpt-2_2019}) and encoder-decoder (e.g., T5 \cite{rael_exploring_nodate}). The generation mechanisms of these two structures for sequences are depicted on the left and in the middle of Figure \ref{PLM_D}, respectively. These two architectures have not only significantly propelled the advancement of PLMs but have also become the predominant foundational architectures for LLMs.

\highlight{Many improved Autoregression LMs-based methods have been proposed to enhance the performance of PLMs in mathematical problems. }
GPT-$f$ \cite{polu2020generative} presents an automated prover and proof assistant to explore the performance of LMs to automated theorem proving. 
To resolve challenges related to expression fragmentation and operand-context separation, Kim et al. \cite{kim_point_2020} introduce the Expression-Pointer Transformer (EPT), a pure neural model that employs both an `Expression' token and operand-context pointers during the generation of solution equations. 
Generate $\&$ Rank \cite{shen_generate_2021} introduces a novel ranking task for MWPs, a multi-task framework built upon a generative PLM, effectively addressing the challenge of minor mistakes in mathematical expressions. 
Thor \cite{jiang2022thor} introduces a framework that integrates language models and automated TP. The latter is employed to selectively choose relevant premises from an extensive library to assist the language model in TP. 
HTPS \cite{lample_hypertree_nodate} presents HyperTree Proof Search, a novel algorithm leveraging online training from past proof searches, enhancing model performance and surpassing the previous SOTA GPT-$f$. 
Galactica \cite{taylor_galactica_2022} proposes a working memory token approach that can achieve strong performance over existing methods on mathematical MMLU \cite{hendryckstest2021} and MATH \cite{hendrycks2021measuring} benchmarks.

Furthermore, MATH-PLM \cite{hendrycks2021measuring} introduces a new dataset named MATH, consisting of challenging competition mathematics problems. It is found that relying solely on increasing budgets and model parameter counts would be impractical for achieving robust mathematical reasoning. 
LISA \cite{jiang2021lisa} gathers an extensive collection of lemmas and theorems extracted from the Isabelle standard library and the Archive of Formal Proofs (AFP). Using this vast corpus, it builds LMs that prove theorems effectively within the AFP. 
PACT \cite{han2021proof} extracts abundant self-supervised data from kernel-level proof terms, enabling joint training alongside the standard tactic prediction objective. This effectively enhances the model's ability for TP. 
Minerva \cite{lewkowycz2022solving} trains a language model, initially pre-trained on general natural language data, through additional training on technical content to address Quantitative Reasoning Problems. It achieves a SOTA performance on technical benchmarks including MATH \cite{hendrycks2021measuring}. 
LIME \cite{wu_lime_nodate} introduces a novel pre-training methodology that specifically learns inductive bias for mathematical reasoning. Observations indicate that models trained using LIME exhibit superior performance compared to standard transformers. 

\subsection{Non-Autoregression LMs}
\label{Non-Autoregression LMs}
\highlight{Unlike Autoregression LMs, Non-Autoregression LMs enable the model} to generate all parts of a sequence simultaneously, without depending on previously generated content, as depicted on the right side \highlight{of Figure \ref{PLM_D}. Notably, models like BERT \cite{kenton2019bert}, Roberta \cite{liuRoBERTaRobustlyOptimized2019}}, and related architectures utilize masked word representations, leveraging pre-trained context-aware embeddings as comprehensive semantic features. This significantly elevates the performance benchmarks of NLP tasks.

In the field of mathematics, researchers have proposed various methods for designing models to address mathematical reasoning and computational problems. For example, Aristo \cite{clarkRegentsScienceExams2021} fine-tunes BERT using scientific curriculum data, yielding promising results on science exams. GenBERT \cite{gevaInjectingNumericalReasoning2020} and NF-NSM \cite{fengInjectingNumericalReasoning2022} enhance the numerical reasoning capabilities of models by incorporating numerical data into the training process of PLMs. MWP-BERT \cite{liangMWPBERTNumeracyaugmentedPretraining2022} further enhances the model's capacity to represent and calculate numerical values by incorporating numeric attributes into symbol placeholders. MathBERT\cite{pengMathBERTPretrainedModel2021} employs additional joint training of text and formulas to effectively capture the semantic-level structural information of formulas. TAGOP \cite{zhu2021tat}, MT2Net \cite{zhao2022multihiertt}, and DeductReasoner \cite{jieLearningReasonDeductively2022} utilize BERT or RoBERTa to extract the fundamental arithmetic relationships between quantities, enabling mathematical reasoning and operations. BERT-TD \cite{liSeekingPatternsNot2022} utilizes semantic encoding and contrastive learning to cluster problems with similar prototype equations, thereby enhancing the understanding of MWP patterns.

%% file: imgs/taxonomy_PLLMs.tex
\tikzstyle{my-box}=[
    rectangle,
    draw=hidden-black,
    rounded corners,
    text opacity=1,
    minimum height=1.5em,
    minimum width=5em,
    inner sep=2pt,
    align=center,
    fill opacity=.5,
]
\tikzstyle{leaf}=[
    my-box, 
    minimum height=1.5em,
    fill=hidden-blue!90, 
    text=black,
    align=left,
    font=\normalsize,
    inner xsep=2pt,
    inner ysep=4pt,
]
\begin{figure*}[t]
    \vspace{-2mm}
    \centering
    \resizebox{\textwidth}{!}{
        \begin{forest}
            forked edges,
            for tree={
                grow=east,
                reversed=true,
                anchor=base west,
                parent anchor=east,
                child anchor=west,
                base=left,
                font=\large,
                rectangle,
                draw=hidden-black,
                rounded corners,
                align=left,
                minimum width=4em,
                edge+={darkgray, line width=1pt},
                s sep=3pt,
                inner xsep=2pt,
                inner ysep=3pt,
                line width=0.8pt,
                ver/.style={rotate=90, child anchor=north, parent anchor=south, anchor=center},
            },
            where level=1{text width=7.4em,font=\normalsize,}{},
            where level=2{text width=8.4em,font=\normalsize,}{},
            where level=3{text width=8.0em,font=\normalsize,}{},
            where level=4{text width=12em,font=\normalsize,}{},
            [
                Language Models, ver
                [
                    Pre-trained \\ Language \\ Models ~(\S\ref{PLMs})
                    [
                        Autoregression \\ LMs ~(\S\ref{Autoregression LMs})
                        [
                           GPT-$f$~\cite{polu2020generative}{,}
                            EPT~\cite{kim_point_2020}{,}
                            Generate $\&$ Rank~\cite{shen_generate_2021}{,}
                            Thor~\cite{jiang2022thor}{,}
                            HTPS~\cite{lample_hypertree_nodate}{,}
                            Galactica~\cite{taylor_galactica_2022}{,}
                            MATH-PLM~\cite{hendrycks2021measuring}{,}
                            LISA~\cite{jiang2021lisa}{,}
                            \\
                            PACT~\cite{han2021proof}{,}
                            Minerva~\cite{lewkowycz2022solving}{,} 
                            LIME~\cite{wu_lime_nodate}
                            , leaf, text width=54.2em
                        ]
                    ]
                    [
                        Non-Autoregression \\ LMs ~(\S\ref{Non-Autoregression LMs})
                        [
                            Aristo~\cite{clarkRegentsScienceExams2021}{,}
                            GenBERT~\cite{gevaInjectingNumericalReasoning2020}{,}
                            NF-NSM~\cite{fengInjectingNumericalReasoning2022}{,}
                            MWP-BERT~\cite{liangMWPBERTNumeracyaugmentedPretraining2022}{,} MathBERT\cite{pengMathBERTPretrainedModel2021}{,}
                            TAGOP~\cite{zhu2021tat}{,}
                            MT2Net~\cite{zhao2022multihiertt}{,}
                            \\
                            DeductReasoner~\cite{jieLearningReasonDeductively2022}{,} 
                            BERT-TD~\cite{liSeekingPatternsNot2022}
                            , leaf, text width=54.2em
                        ]
                    ]
                ]
                [
                     Large \\ Language \\ Models ~(\S\ref{LLMs})
                    [
                        Instruction \\ Learning ~(\S\ref{Instruction Learning})
                        [
                            Instruction \\ Building 
                            [
                                Auto-explanation~\cite{nakamoto2023enhancing}{,}
                                ProofNet~\cite{azerbayev2023proofnet}{,}
                                WizardLM~\cite{xu_wizardlm_2023}{,}
                                Wizardmath~\cite{luo2023wizardmath}{,}
                                \highlight{ControlMath~\cite{chen2024controlmath}{,}} \\
                                \highlight{EURUS~\cite{yuan2024advancing}{,}}
                                \highlight{RefAug~\cite{zhang2024learn}}
                                , leaf, text width=44.6em
                            ]
                        ]
                        [
                            Instruct \\ Tuning 
                            [
                                MathGLM~\cite{yang2023gpt}{,}
                                Goat~\cite{liu_goat_2023}{,}
                                Calculon~\cite{muffo_evaluating_nodate}{,}
                                PaLM 2-L-Math~\cite{liu_improving_2023}{,}
                                LLEMMA~\cite{azerbayev8llemma}{,}
                                \highlight{
                                Qwen-Math~\cite{yang2024qwen2}{,}} \\
                                \highlight{
                                REFT~\cite{luong2024reft}{,}
                                InternLM-Math~\cite{ying2024internlmmath}}
                                , leaf, text width=44.6em
                            ]
                        ]
                        [
                            In-context \\Learning 
                            [
                                ScratchpadGPT~\cite{nye_show_2021}{,}
                                Codex-math~\cite{drori2022neural}{,}
                                In-Context Sampling~\cite{liu2022makes}{,}
                                ComplexPrompt~\cite{fu2022complexity}{,} 
                                \\
                                ART~\cite{zhang2023automatic}{,}
                                \highlight{
                                LAMBADA~\cite{kazemi2022lambada} {,}
                                LogicSolver~\cite{yang2022logicsolver}}
                                , leaf, text width=44.6em
                            ]
                        ]
                    ]
                    [
                        Tool-based \\ Methods ~(\S\ref{Tool-based Methods})
                        [
                            Single-tool \\ Methods 
                            [
                                SymbLLM~\cite{he2023solving}{,}
                                 PoT~\cite{chen2022program}{,}
                                 Codex-math~\cite{drori2022neural}{,}
                                 MathPrompter~\cite{imani2023mathprompter}{,}
                                 PAL~\cite{gao2023pal}{,}
                                 Code-scratchpad~\cite{upadhyay-etal-2023-improving}{,}\\
                                 \highlight{
                                 MuMath-Code~\cite{yin2024mumath}{,}
                                 LeanReasoner~\cite{jiang2024leanreasoner}{,}
                                 LINC~\cite{olausson2023linc}{,}
                                 }
                                , leaf, text width=44.6em
                            ]
                        ]
                        [
                            Multi-tool \\ Methods 
                            [
                                Toolformer~\cite{schick2023toolformer}{,}
                                Chameleon~\cite{lu2023chameleon}{,}
                                ART~\cite{paranjape2023art}{,}
                                ToolkenGPT~\cite{hao2023toolkengpt}{,}
                                CRITIC~\cite{gou2023critic}{,}     
                                TALM~\cite{parisi2022talm}{,} \\
                                Tool-Documentation~\cite{hsieh2023tool} 
                                , leaf, text width=44.6em
                            ]
                        ]
                    ]
                    [
                       Fundamental CoT \\ Methods ~(\S\ref{Fundamental CoT Methods})
                        [
                            Foundation of \\ CoT 
                            [
                                Answer Rationales~\cite{ling2017program}{,}
                                MathQA~\cite{amini2019mathqa}{,}
                                CoT~\cite{wei2022chain}{,}
                                Complexity-based CoT~\cite{fu2022complexity}{,}
                                RFT~\cite{yuan2023scaling}{,}
                                AFT~\cite{wang2023making}{,}
                                \\
                                Thought Propagation~\cite{yu2023thought}{,}
                                PoT~\cite{chen2022program}{,}
                                AOT~\cite{sel2023algorithm}{,}
                                MAmmoTH~\cite{yue2023mammoth}{,}
                                \highlight{SymbCoT~\cite{xu2024faithful}}
                                , leaf, text width=44.6em
                            ]
                        ]
                        [
                            Construction of \\ CoT 
                            [
                                Zero-shot CoT~\cite{kojima_large_nodate}{,}
                                Auto-CoT~\cite{zhang2023automatic}{,}
                                Complexity-based CoT~\cite{fu2022complexity}{,}
                               PromptPG-CoT~\cite{lu2022dynamic}{,} \\
                               AutoMate CoT~\cite{shum_automatic_2023}{,}
                                BoostedPrompt~\cite{pitis_boosted_2023}{,}
                                \highlight{CoT-Influx~\cite{huang2024fewer}}
                                , leaf, text width=44.6em
                            ]
                        ]
                    ]
                    [
                        Advanced CoT \\ Methods ~(\S\ref{Advanced CoT Methods})
                        [
                            Verify-based \\ Methods 
                            [
                                Code-based self-verification~\cite{zhou2023solving}{,}
                                VerifyCoT~\cite{ling2023deductive}{,}
                                DIVERSE~\cite{li2023making}{,}
                                Verify-and-Edit~\cite{zhao2023verify}{,}
                                \\
                                Retrieval-CoT~\cite{he_rethinking_2022}{,}
                                SCREWS~\cite{shridhar2023screws}{,}
                                \highlight{SSC-CoT~\cite{zhao2024stepwise}}
                                , leaf, text width=44.6em
                            ]
                        ]
                        [
                            Ensemble-based \\ Methods 
                            [
                                Self-Consistency~\cite{wang2022self}{,}
                                Diversity-of-Thought~\cite{naik2023diversity}{,}
                                Complexity-based CoT~\cite{fu2022complexity}{,}
                                DIVERSE~\cite{li2023making}{,}
                                \\
                                Self-check~\cite{miao2023selfcheck}{,}
                                MCR~\cite{yoran2023answering}{,}
                                Rank-verifier~\cite{cobbe2021training}{,}
                                GRACE~\cite{khalifa2023discriminator}
                                , leaf, text width=44.6em
                            ]
                        ]
                        [
                            Planing-based \\ Methods 
                            [
                                ToT~\cite{yao2023tree}{,}
                                TouT~\cite{mo2023tree}{,}
                                ToT-system~\cite{long2023large}{,}
                                GoT~\cite{yao2023beyond,besta_graph_2023,lei2023boosting}{,}
                                RESPROMPT~\cite{jiang2023resprompt}{,} 
                                RAP~\cite{hao2023reasoning}{,} 
                                LATS~\cite{zhou2023language}{,}  \\
                                LLM+P~\cite{liu2023llm+}{,}
                                LLM+DP~\cite{dagan2023dynamic}{,}
                                Self-refine~\cite{madaan2023self,sun2023adaplanner}{,}
                                ISR-LLM~\cite{zhou2023isr}{,}
                                Reflexion~\cite{shinn2023reflexion}{,} 
                                \highlight{CR~\cite{zhang2023cumulative}} \\
                                \highlight{Step-Plan~\cite{zhang2023interpretable}}
                                , leaf, text width=44.6em
                            ]
                        ]
                        [
                            \highlight{Self-correct} \\ \highlight{Methods} 
                            [
                                \highlight{
                                o1~\cite{openai_o1_system_card}{,}
                                QWQ~\cite{qwq-32b-preview}{,}
                                STaR~\cite{zelikman2022star}{,}
                                V-STaR~\cite{hosseini2024v}{,}
                                Quiet-STaR~\cite{zelikman2024quiet}{,}
                                REFT~\cite{trung2024reft}{,}
                                SCoRe~\cite{kumar2024training}{,}
                                HGS-PRM~\cite{ma2023let}
                                } \\
                                \highlight{
                                MCTSr~\cite{zhang2024accessing}{,}
                                CoRe~\cite{zhu2022solving}{,}
                                Marco-o1~\cite{zhao2024marco}{,}
                                PRM~\cite{DBLP:conf/iclr/LightmanKBEBLLS24}{,}
                                Step-DPO~\cite{lai2024step}{,}
                                Flow-DPO~\cite{deng2024flow}{,}
                                OmegaPRM~\cite{luo2024improve}
                                } \\
                                , leaf, text width=44.6em
                            ]
                        ]
                        [
                            Socratic Teaching \\ Methods 
                            [
                                Socratic Models~\cite{zeng2022socratic}{,}
                                SOCRATIC QUESTIONING~\cite{qi2023art}{,}
                                Socratic prompt~\cite{chang2023prompting}{,}
                                SocratiQ~\cite{al2023socratic}{,}
                                \\
                                multi-turn Socratic advice~\cite{ang2023socratic}{,}
                                \highlight{SocraticLLM~\cite{ding2024boosting}{,}
                                MATHDIAL~\cite{macina2023mathdial}}
                                , leaf, text width=44.6em
                            ]
                        ]
                    ]
                    [
                        \highlight{Multi-modal} \\ \highlight{Methods} ~(\S\ref{Multi-modal Methods})
                        [
                            \highlight{Multi-modal} \\ \highlight{Reasoning} 
                            [
                                \highlight{
                                GPT-4V~\cite{openai2023gpt4}{,}
                                Qwen2-VL~\cite{Qwen2-VL}{,}
                                Gemini~\cite{team2023gemini}{,}
                                GPT-4o~\cite{hurst2024gpt}{,}
                                GLM-4V~\cite{glm2024chatglm}{,}
                                LLaVA-o1~\cite{xu2024llava}{,}
                                AtomThink~\cite{xiang2024atomthink}{,}}\\
                                \highlight{
                                Math-LLaVA~\cite{shihu2024mathllava}{,}
                                M-STAR~\cite{anonymous2024diving}{,}
                                UnAC~\cite{wang2024understanding}{,}
                                }
                                , leaf, text width=44.6em
                            ]
                        ]
                    ]
                ]
            ]
        \end{forest}
    }
    \vspace{-4mm}
    \caption{Taxonomy of language models for mathematic.}
    \label{fig:taxonomy}
    \vspace{-3mm}
\end{figure*}

%% file: secs/04LLMs.tex
\section{LLMs-based Methods}
\label{LLMs}
Large-scale language models (LLMs) are designed for processing and generating text akin to human communication \cite{touvron2023llama,openai2023gpt4}.
Mathematics is also a form of language, that communicates complex concepts and relationships through a structured system of symbols and notation, akin to the rules of a spoken language. Thus, a language model that grasps these mathematical rules can "speak" the language of mathematics, proving to be a valuable asset for mathematicians.
In numerous ways, a sophisticated language model like GPT-4 \cite{openai2023gpt4} becomes an invaluable tool in the field of mathematics \cite{azerbayev8llemma,luo2023wizardmath}.
We classify the existing studies into four parts: instruction learning, tool-based methods, fundamental CoT methods, and advanced CoT methods (Figure \ref{fig:taxonomy}).

\subsection{Instruction Learning}
\label{Instruction Learning}
Numerous approaches have been proposed to enhance the mathematical performance of models by instruction learning. These approaches are categorically delineated as instruction building, instruction tuning, and in-context learning, based on their distinctive characteristics.

\paragraph{Instruction Building.}
\label{Instruction Building}
\highlight{Instruction Building refers to the process of designing and generating high-quality instruction data that guide models in improving specific capabilities, such as mathematical reasoning. After pre-training, these instructions are crucial for directing models towards more sophisticated understanding and application of mathematical concepts.} A semi-supervised approach, as presented in Auto-explanation \cite{nakamoto2023enhancing}, leverages LLMs to create datasets for automating the scoring of mathematical self-explanations in the context of mathematics education. \highlight{Similarly, RefAug \cite{zhang2024learn} leverages GPT-4o to construct reflective augmentation data for training model, thereby enhancing their performance on mathematical reasoning tasks.} A challenging benchmark ProofNet \cite{azerbayev2023proofnet} is introduced to build a system of automated theorem proving. Meanwhile, prompt retrieval and distilled back translation methods are introduced for statement autoformalization. 
WizardLM \cite{xu_wizardlm_2023} presents a groundbreaking method called Evol-Instruct, designed to autonomously generate high-quality instructions by LLMs \highlight{themselves. 
Furthermore, Wizardmath} \cite{luo2023wizardmath} proposes a reinforced Evol-instruct \highlight{Feedback} method (RLEIF) to build more complexity of instruct dataset. This method effectively enhances the mathematical reasoning capabilities of Llama-2 by supervised fine-tuning and \highlight{‌Proximal Policy Optimization (PPO)} training \cite{schulman2017proximal}. \highlight{ControlMath \cite{chen2024controlmath} introduces a novel approach that first generates diverse equations, which are then transformed into math word problems by a Problem Crafter LLM-agent. EURUS \cite{yuan2024advancing} generates a large number of reasoning trajectories by leveraging CoT \cite{wei2022chain} and tool creation \cite{qian2023creator} methods. These trajectories within multi-turn interactions \cite{wang2023mint} are organized into a preference tree structure. Then extensive instruction and preference data are collected from this tree to enhance the model's mathematical reasoning capabilities by Direct Preference Optimization (DPO) \cite{rafailov2024direct}, Kahneman-Tversky Optimization (KTO) \cite{ethayarajh2024kto}, and Noise Contrastive Alignment (NCA) \cite{chen2024noise}.}




\paragraph{Instruction Tuning.}
\label{Instruct Tuning}
Instruction tuning stands as a potent method to elevate the prowess of large models, aiding in steering the model towards generating outputs that align with human intent \cite{zhang_instruction_2023}. Compared to the pre-training phase of LLMs, instruction tuning demands \highlight{notably less data and computational resources}, making it a prevalent strategy for enhancing a model's domain-specific capabilities. For instance, MathGLM \cite{yang2023gpt} showcases that even an LLM with \highlight{2B parameters can adeptly} execute multi-digit arithmetic operations with limited training data.
Goat \cite{liu_goat_2023} discerns patterns in the tokenization of different numbers and proposes a consistent approach \highlight{for their tokenization based on the LLaMA model.}  
Similarly, Calculon \cite{muffo_evaluating_nodate} refines the model's arithmetic skills by employing a digit decomposition technique to construct a fine-tuning dataset. 




In PaLM 2-L-Math \cite{liu_improving_2023}, three fine-tuning strategies were explored, revealing key insights: 1) The quality and style of step-by-step solutions significantly influence model performance; 2) Combining solution re-ranking and majority voting yields a more substantial performance enhancement compared to using them individually. Additionally, multi-task fine-tuning, separating solution generation and evaluation tasks, surpasses the baseline achieved by solution fine-tuning alone.
Moving beyond the enhancement of mathematical capabilities through fine-tuning, LLEMMA \cite{azerbayev8llemma} introduces the Proof-Pile-2 dataset, an amalgamation of mathematical texts and code. Continual pre-training with Code Llama empowers the model to leverage Python interpreters and formal theorem provers, demonstrating impressive performance on the MATH benchmark.
\highlight{To improve the effectiveness of instruction tuning, Luong et al. \cite{luong2024reft} introduce an approach that incorporates online reinforcement learning. This method enhances the model's generalization by exploring multiple CoTs and using real-world answers as reward signals.}

\highlight{Recently, more math-specific LLMs \cite{yang2024qwen2, ying2024internlmmath} have been proposed, with enhanced instruction tuning tailored for the mathematical domain.
Qwen2.5-MATH \cite{yang2024qwen2} begins by pretraining on large-scale, high-quality mathematical data. In the post-training phase, it develops a reward model (RM), which is then utilized to the iterative evolution of data in supervised fine-tuning (SFT). 
Thereafter, both the SFT model and the RM undergo iterative training. During the inference stage, the RM guides the sampling process to optimize the model's performance.
In contrast, InternLM-Math \cite{ying2024internlmmath} undergoes secondary pretraining on a base model using a mathematical corpus. In the post-training phase, it is fine-tuned with instructions through unifying CoT and code interpretation, reward modeling, data augmentation under an unified seq2seq format.}

\paragraph{In-context Learning.}
\label{In-context Learning}
In-context Learning (ICL) \cite{brown2020language,kaplan2020scaling} empowers LLMs to execute target tasks by presenting specific task examples as conditions during inference, without updating model parameters. 
Inspired by this, ScratchpadGPT \cite{nye_show_2021} enhances its proficiency in multi-step computations by mandating the model to output intermediate computation steps into a designated ``scratchpad."
Codex-math \cite{drori2022neural} refines Codex by generating programs through code and integrates few-shot learning to automatically create programs for solving \highlight{MWPs}. This method has substantially improved the previous state-of-the-art accuracy in automatic solutions by 73.1\% across various benchmarks like MIT's mathematics courses, Columbia University’s Computational Linear Algebra, and the MATH benchmark \cite{hendrycks2021measuring}. Notably, few-shot learning contributed to a 10\% accuracy boost.
Recent studies also highlight the variability in in-context learning performance across different chosen examples \cite{liu2022makes}. Fu et al. \cite{fu2022complexity} and Zhang et al. \cite{zhang2023automatic} selected intricate and diverse examples to enhance reasoning performance. \highlight{LAMBADA \cite{kazemi2022lambada} introduces a backward reasoning method for mathematical inference, starting from conclusions and working back to axioms. It uses few-shot to guide LLMs in reverse problem-solving, achieving promising results. LogicSolver \cite{yang2022logicsolver} first constructed a dataset of mathematical word problems, InterMWP, where each problem includes interpretable logical formulas based on algebraic knowledge. The relevant logical formulas in InterMWP is used as In-context to enhance both interpretability and accuracy.}




\subsection{Tool-based Methods}
\label{Tool-based Methods}
LLMs are designed to use tools, such as codes and calculators, to enhance their problem-solving abilities \cite{schick2023toolformer,parisi2022talm}. 

\paragraph{Single-tool Methods.}
\label{Single-tool} 
To improve the performance of mathematical reasoning, math-specific tools such as symbolic solvers and programs are utilized for LLMs \cite{he2023solving,chen2022program}.
For example, SymbLLM \cite{he2023solving} \highlight{solved MWPs by combining LMs} with symbolic solvers. They focus on automatically generating high-quality, step-by-step solutions to mathematical word problems, especially those encountered in mathematical applications. \highlight{LeanReasoner \cite{jiang2024leanreasoner} and Lean Copilot \cite{song2024towards} integrates LLMs with the Lean theorem proving framework. Leveraging Lean's symbolic solver reduces the risk of logical inconsistencies and enhances the LLMs' capability to handle complex reasoning tasks through the extensive theorem libraries available in Lean. Similarly, LINC \cite{olausson2023linc} integrates LLMs with a first-order logic theorem prover. The LLM translates natural language premises and conclusions into First-Order Logic expressions. These expressions are then sent to an external theorem prover, such as Prover9, for symbolic deductive reasoning to determine the truth value of the conclusion.}

Furthermore, previous studies have explored the process of converting \highlight{MWPs} into code \cite{chen2022program,drori2022neural}. PoT \cite{chen2022program} proposes a fusion of CoT with programs, while Drori et al. \cite{drori2022neural} showcased the effectiveness \highlight{of pre-training on text and fine-tuning on code.} This approach successfully solves, explains and generates math problems at the university level. MathPrompter \cite{imani2023mathprompter} employs a \highlight{zero-shot CoT prompting technique} to generate multiple algebraic expressions or Python functions in varied ways to \highlight{solve the same math problem.} PAL \cite{gao2023pal} introduces an innovative approach to bolster the performance of \highlight{LLMs} in mathematical problem-solving. This involves utilizing LLMs to comprehend natural language problems and generate programs as intermediate reasoning steps. \highlight{To enhance the ability of LLMs to solve MWPs using code, MuMath-Code \cite{yin2024mumath} constructed an augmented dataset that uses multi-perspective data augmentation methods and synthesizes code-nested solutions.}

In addition to solving math problems, programs can also play a role in tutoring math. For instance, Upadhyay explores the role of \highlight{LLMs} in tutoring systems \cite{upadhyay-etal-2023-improving}, introducing a ``code scratchpad" alongside the traditional ``language scratchpad" to enhance the model's performance in tutoring steps, particularly using a grade school mathematics dataset.

\paragraph{Multi-tool Methods.}
\label{Multi-tool}
To facilitate the seamless integration of LLMs with diverse tools, several multi-tool approaches have been proposed to enable LLMs to learn how to use multiple tools simultaneously. Toolformer \cite{schick2023toolformer} adopts a self-supervised training approach to enable the utilization of different tools such as search engines, calculators and translation systems via simple API calls. This is accomplished by fine-tuning on a vast collection of sampled API calls and filtering based on their ability to reduce perplexity on subsequent tokens.
Chameleon \cite{lu2023chameleon}, on the other hand, enhances LLMs with plug-and-play modules designed for compositional reasoning. It creates programs by combining various tools, including LLMs, off-the-shelf vision models, web search engines, Python functions, and heuristic-based modules, to tackle complex reasoning tasks.
The Automatic Reasoning and Tool-use (ART) framework \cite{paranjape2023art} leverages frozen LLMs to automatically generate intermediate reasoning steps as a program. This seamlessly incorporates external tools to support computations that surpass the core capabilities of LLMs.
Meanwhile, ToolkenGPT \cite{hao2023toolkengpt} adopts a strategy of learning ``toolken" embeddings to represent each tool as a token. This empowers LLMs to effortlessly utilize tools similar to generating word tokens. It caters to a broader range of tools and utilizes extensive demonstration data to learn toolkit embeddings.

Furthermore, CRITIC \cite{gou2023critic} significantly enhances the outputs of LLMs by enabling them to verify and self-correct through interactions with external tools. Inspired by human cognition and critical thinking, CRITIC continuously refines text generated by LLMs through iterative interactions with tools like search engines and code interpreters. Tool Augmented Language Models (TALM) \cite{parisi2022talm} seamlessly integrate language models with non-differentiable tools. They employ an iterative ``self-play" method to bootstrap performance based on a limited number of tool demonstrations. In contrast, Tool-Documentation \cite{hsieh2023tool} opted for tools based on documentation rather than relying on demonstrations. Current practices involve teaching LLMs through a few-shot demonstration of tool usage, a process prone to bias and overfitting. The alternative proposed by Tool-Documentation, utilizing tool documentation that provides descriptions of individual tool usage, is argued to be a more effective approach. ToRA \cite{gou2023tora} employs a set of Tool-integrated Reasoning Agents, which integrate computational libraries and symbolic solvers, among other mathematical tools, to effectively solve complex \highlight{MWPs}.



\subsection{Fundamental CoT Methods}
\label{Fundamental CoT Methods}

A multitude of fundamental Chain-of-Thought (CoT) methods integrated with LLMs have been proposed to enhance the mathematical reasoning abilities of LLMs.

\paragraph{\highlight{Foundation of CoT}}
\label{Foundation of CoT}

In the initial stages of research, a limited body of work has leveraged the principles of CoT to enhance the mathematical capabilities of language models. Notably, Ling et al \cite{ling2017program} propose a methodology involving the generation of a series of concise steps, termed "Answer Rationales," to guide the resolution of algebraic word problems. MathQA \cite{amini2019mathqa} suggests decomposing Math Word Problems into multiple steps corresponding to programmatic operations for resolution. 

Utilizing the in-context learning of LLMs, CoT \cite{wei2022chain} explicitly introduces the concept of the CoT for the first time, and substantiates its efficacy in enhancing reasoning abilities. 
This validation is also corroborated in Complexity-based CoT \cite{fu2022complexity}. Consequently, several methods \cite{chen2022program,yuan2023scaling,yu2023thought,wang2023making,sel2023algorithm,yue2023mammoth} leverage CoT methodologies to enhance the mathematical performance of models.



RFT \cite{yuan2023scaling} proposes the application of rejection sampling finetuning (RFT) to gather more optimal reasoning paths, thereby enhancing mathematical reasoning performance. 
Thought Propagation \cite{yu2023thought} considers gaining insights from analogous problems to assist in addressing the current issue. 
Wang et al. \cite{wang2023making} proposed a Fine-Tuning (AFT) paradigm that aligns the model to prioritize the generation of CoT with superior performance. 
\highlight{SymbCoT \cite{xu2024faithful} integrates symbolic expressions and logical rules with CoT prompting,  enhancing the logical rigor of CoT.}
Additionally, PoT \cite{chen2022program} introduces a Program of Thoughts, combining CoT and programming, and AOT \cite{sel2023algorithm} enhances reasoning abilities through algorithmic-style demonstrations of CoT. Furthermore, MAmmoTH \cite{yue2023mammoth} integrates CoT and PoT \cite{chen2022program} rationales to instruct large language models in utilizing code tools for solving mathematical problems.

\paragraph{\highlight{Construction of CoT}}
\label{Construction of CoT}
To streamline the process of CoT creation, various approaches \cite{lu2022dynamic,zhang2023automatic,fu2022complexity,wang2023making,kojima_large_nodate,shum_automatic_2023,pitis_boosted_2023,sel2023algorithm,yu2023thought} have been introduced for its automatic generation.

Zero-shot CoT \cite{kojima_large_nodate} introduces a simple CoT example, ``Let’s think step by step," which effectively enhances the model's reasoning capabilities. Auto-CoT \cite{zhang2023automatic} selects representative questions through clustering and answers them using ``Let’s think step by step," concatenating the question-answer pairs one by one to automatically generate CoT.
In the case of Complexity-based CoT \cite{fu2022complexity}, the demonstration is chosen by simply selecting the reasoning chain with the highest number of inference steps sampled from the model.
\highlight{Several methods aim to enhance the quality of CoT generated by models. CoT-Influx \cite{huang2024fewer} trains a pruner specifically optimized for enhancing the mathematical capabilities of CoT generation, and validates the effectiveness of the pruner in improving the mathematical performance of LLMs on the GSM8K \cite{cobbe2021training}.}
 Using reinforcement learning, PromptPG-CoT \cite{lu2022dynamic} trains a model with policy gradient to assist GPT-3 in selecting suitable CoT demonstrations. 
Similarly, AutoMate CoT \cite{shum_automatic_2023} employs a variance-reduced policy gradient strategy to estimate the significance of each example in a black box language model, thereby selecting more effective examples. BoostedPrompt \cite{pitis_boosted_2023} presents an iterative prompt ensemble method that enhances prompts when the current demonstration faces challenges in handling specific problems.

\subsection{Advanced CoT Methods}
\label{Advanced CoT Methods}

To further enhance the capability of CoT in LLMs, advanced CoT methods have been proposed, including Verify-based Methods, Ensemble-based Methods, Planning-based Methods, and Socratic Teaching Methods.

\paragraph{Verify-based Methods.} 
\label{Verify-based Methods}

LLMs often produce incorrect reasoning steps, which can lead to a series of cascading errors in their solutions. Implementing verification and refining the reasoning process based on feedback can significantly reduce these errors. This approach is akin to human reflection and involves critically evaluating each step of the reasoning process. Various verify-based methods \cite{zhou2023solving,ling2023deductive,li2023making,zhao2023verify,he_rethinking_2022,shridhar2023screws} have been proposed to address these issues. 

Zhou et al. \cite{zhou2023solving} propose a code-based self-verification approach, utilizing a zero-shot prompt to encourage the GPT-4 Code Interpreter to employ code for self-verifying its responses, enhancing its mathematical reasoning capabilities. To mitigate the challenge of validating the entire deductive reasoning process, VerifyCoT \cite{ling2023deductive} introduces a deductive reasoning form, ensuring that each reasoning step strictly relies on the preceding steps. Furthermore, DIVERSE \cite{li2023making} independently verifies each reasoning step and a voting mechanism to eliminate incorrect answers. Both Verify-and-Edit \cite{zhao2023verify} and Retrieval-CoT \cite{he_rethinking_2022} utilizes external retrieval tools to support the model in validating reasoning rationales. The key difference is the former edits rationales using retrieved information, while the latter helps the model self-rethink to improve performance on complex reasoning tasks. Both methods effectively reduce factual mistakes during the reasoning process. 
Shridhar et al. \cite{shridhar2023screws} summarize the use of the revisions approach in the SCREWS framework, which includes a selection module to choose between the original and modified reasoning steps.
\highlight{SSC-CoT \cite{zhao2024stepwise} found that prompting the model from critical intermediate steps can enhance the success rate of solving mathematical problems. Therefore, it emphasizes verifying these key steps during the inference process.}


\paragraph{Ensemble-based Methods.} 
\label{Ensemble-based Methods}
Due to the inherent stochastic nature of LLMs, which output probability distributions over predicted words, they may randomly generate incorrect reasoning steps and outcomes. To tackle this challenge, some methods \cite{wang2022self,naik2023diversity,fu2022complexity,li2023making,miao2023selfcheck,yoran2023answering,cobbe2021training,khalifa2023discriminator} leverage the concept of ensemble learning, employing techniques such as voting and ranking to eliminate uncertainties in the reasoning process.

Self-Consistency \cite{wang2022self} employs multiple reasoning paths and selects the final response through a simple majority vote. Similarly, Diversity-of-Thought \cite{naik2023diversity} generates diverse reasoning paths by altering prompts and aggregates the responses via the majority vote. Complexity-based CoT\cite{fu2022complexity} favors answers derived from more intricate reasoning paths. DIVERSE \cite{li2023making} uses a weighted voting mechanism to filter out incorrect answers. Nevertheless, these voting-based methods often overlook the potentially useful information within unsuccessful CoT reasoning and lack an efficient integration of multiple reasoning chains to improve performance. Self-check \cite{miao2023selfcheck} addresses this by incorporating reasoning steps into the voting mechanism, ensuring both consistent answers and reliable reasoning. MCR \cite{yoran2023answering} takes a step further by consolidating information across various reasoning chains, acquiring the most relevant facts, facilitating a more comprehensive analysis to make successful reasoning. 

Additionally, there are a few methods based on ranking. Rank-verifier \cite{cobbe2021training} proposes using a ranking system to judge the correctness of model completions. Furthermore, GRACE \cite{khalifa2023discriminator} leverages a discriminator trained via contrastive learning to rank each reasoning step.

\paragraph{Planing-based Methods.} 
\label{Planing-based Methods}

The original structure of the Chain of Thought (CoT) is sequential, facing limitations when handling highly complex problems and lacking the ability for retrospective correction. 
To make the CoT chain structure more systematic and intricate, certain planning-based methods \cite{yao2023tree,mo2023tree,long2023large,yao2023beyond,besta_graph_2023,lei2023boosting,jiang2023resprompt,hao2023reasoning,zhou2023language,liu2023llm+,dagan2023dynamic,madaan2023self,sun2023adaplanner,zhou2023isr,shinn2023reflexion} have been proposed. These methods achieve this by altering the organizational structure of reasoning steps or incorporating mechanisms for refinement and reflection. 

The Tree-of-Thought (ToT) \cite{yao2023tree} proposes organizing reasoning paths into a tree structure, where each node represents a reasoning step, and edges denote dependencies between nodes. During the inference process, ToT can use self-assessment to determine future actions. This structure facilitates both forward and backward exploration, employing either deep-first or breadth-first search techniques. TouT \cite{mo2023tree} effectively employs Monte Carlo Dropout to assess uncertainty scores associated with diverse local responses of language models at intermediate steps, enhancing response precision through integration with global search algorithms. 
Long et al. \cite{long2023large} introduced a multi-module ToT system, where the state of the ToT thought chain process can be stored and retraced through its storage module. 
Furthermore, Yao et al. \cite{yao2023beyond}, Besta et al. \cite{besta_graph_2023}, and Lei et al. \cite{lei2023boosting} introduced a novel thought structure called the Graph of Thought (GoT). 
Similar to ToT, each graph node represents a distinct thought step or state, with edges indicating dependencies between nodes. These methods can effectively integrate diverse reasoning thoughts, fostering collaborative outcomes and improving reasoning by incorporating feedback loops. This contributes to an overall enhancement in the inference capacity of the thought network. 
\highlight{CR \cite{zhang2023cumulative} proposes a method where LLMs assume three roles—Proposer, Verifier, and Reporter—to collaboratively solve problems. This approach differs from CoT and ToT in its ability to dynamically store and utilize all historically validated inference results for execution and combination, a concept akin to GoT.}
Additionally, RESPROMPT \cite{jiang2023resprompt} introduces Residual Connection Prompting, which converts the input prompt into complex reasoning graphs.

Inspired by Monte Carlo tree search \highlight{(MCTS) \cite{chaslot2008monte}}, RAP \cite{hao2023reasoning} utilizes the \highlight{MCTS} to traverse tree structures during the reasoning process. This approach effectively balances global and local searches to obtain a high-reward reasoning path. 
On the other hand, LATS \cite{zhou2023language} employs LLMs as agents, value functions, and optimizers, repurposing their inherent strengths to enhance decision-making. \highlight{Step-Plan~\cite{zhang2023interpretable} propose a step-by-step planning approach for intermediate solution generation, which strategically plans the generation of the next solution step.}
Liu et al. \cite{liu2023llm+} and Dagan et al. \cite{dagan2023dynamic} introduced LLM+P and LLM+DP, incorporating the generation of Planning Domain Definition Language (PDDL) to break down complex problems and execute plans using specialized models. 

Furthermore, Self-Refine, as proposed by Madaan et al. \cite{madaan2023self} and Sun et al. \cite{sun2023adaplanner}, focuses on error correction and summarizing past experiences. Specifically, Self-Refine performs pre-execution validation, providing feedback to the model for refinement in case of errors. Zhou et al. \cite{zhou2023isr} then took a step further by merging the iterative self-refine with PDDL in ISR-LLM. 
Additionally, Shinn et al. \cite{shinn2023reflexion} introduced re-flexion to rectify errors stemming from previous actions.

\paragraph{\highlight{Self-correct Methods.}} 
\label{Self-correct Methods}
\highlight{
Recently, a lot of Self-correct Methods have been proposed, which introduce a self-correction mechanism on top of CoT. Combined with techniques like MCTS and RL, these methods enable the model to iteratively enhance itself. The success of OpenAI’s o1  \cite{openai_o1_system_card} marks a significant advancement in high-level reasoning by employing long CoT and self-correction mechanisms. Building on this, QwQ~\cite{qwq-32b-preview}, Marco-o1~\cite{zhao2024marco} and o3 have also been introduced, further validating the effectiveness of long CoT and self-correction.}
\highlight{STaR~\cite{zelikman2022star} leverages existing LLM reasoning to iteratively create a high-quality dataset, refining the model through self-generated inferences. 
Similar to STaR, REFT~\cite{trung2024reft} does not employ SFT, and instead refines the model using the PPO in an online reinforcement learning setting.
V-STaR~\cite{hosseini2024v} builds upon STaR by retaining error trajectories during the iterative generation process. 
Quiet-STaR \cite{zelikman2024quiet} enables LLMs to generate a ``rationale" before predicting the next token, explaining the anticipated content of the future text. 
CoRe~\cite{zhu2022solving} mimics the human brain's dual-system model, separating reasoning into generation and verification, and it uses MCTS for feedback during inference. 
Additionally, SCoRe~\cite{kumar2024training} introduces a two-stage reinforcement strategy to mitigate distribution shifts and behavior collapse, facilitating self-correction.
MCTSr \cite{zhang2024accessing} proposes a method based on Monte Carlo Tree Self-Refine that enhances LLM's performance on the Mathematical Olympiad, matching GPT-4. 
}

\highlight{
Flow-DPO~\cite{deng2024flow} introduces a method where two LLM agents collaborate to generate step-level preference data, which is used to train the LLM using the DPO algorithm. 
Lightman et al. \cite{DBLP:conf/iclr/LightmanKBEBLLS24} presents a Process Reward Model (PRM) and compares it with the Outcome Reward Model. 
Step-DPO~\cite{lai2024step} also introduces a method in which each step of the reasoning process is optimized based on preferences, rather than evaluating the entire answer as a whole. 
OmegaPRM~\cite{luo2024improve} proposes an automated process supervision method with a divide-and-conquer style MCTS algorithm, which significantly improves Gemini's proficiency in mathematical reasoning.
Moreover, HGS-PRM \cite{ma2023let} proposes combining Heuristic Greedy Search with PRM to optimize the reasoning paths.
}

\paragraph{Socratic Teaching Methods.}
\label{Socratic Teaching Methods}
The Socratic teaching method \cite{nelson1980socratic}, derived from the philosophies of ancient Greek thinker Socrates, provides a unique perspective when considering the integration of LMs in mathematics education \cite{davies2021advancing}. This method, emphasizing questioning and dialogue, aims to guide learners \highlight{toward profound thinking \cite{davis1997dialogue,brickhouse2009socratic,ding2024boosting}.} In realm of mathematics education, employing this method can aid students in grasping concepts with clarity, fostering more rigorous logical application, and promoting reflection and self-discovery.

For LLMs, research shows that the Socratic teaching method can also improve their abilities to a certain extent. For example, the framework of Socratic Models (SMs) \cite{zeng2022socratic} significantly broadens the scope of interactive capabilities. SMs facilitate the amalgamation of multiple LLMs to tackle novel multi-modal tasks using structured Socratic dialogues. 
In Chang et al. \cite{chang2023prompting}, techniques including definition, dialectics, cross-examination, induction, and counterfactual reasoning are utilized. These techniques contribute to enhancing the LLMs' comprehension of user intentions, thereby improving model's effectiveness in addressing user queries. 
The inductive, deductive, and analogical reasoning techniques in the Socratic teaching method establish a comprehensive analytical framework for solving mathematical problems \cite{chang2023prompting,alhossami2023language}, ensuring a strong and concise ability to address the complexities of mathematical problem-solving.
\highlight{MATHDIAL \cite{macina2023mathdial} enables LLMs to engage in multi-round dialogues as both teacher and student, creating a conversational  dataset that incorporates rich educational socratic attributes and background in mathematical reasoning problems.}
Moreover, contributions from Ang et al. \cite{ang2023socratic} and Al et al. \cite{al2023socratic} provided invaluable datasets and evaluation methodologies for Socratic questioning. These contributions underscore the practical applications and potential impact of Socratic methods in counseling and coaching, showcasing their diverse applicability.

Overall, the integration of Socratic teaching techniques in LLMs offers novel perspectives and tools for mathematics education. Unlike CoT \cite{wei2022chain}, Socratic questioning explicitly guides the thinking process, stimulating effective recursive thinking and displaying greater resilience to errors in the thinking process \cite{qi2023art}. Ranging from basic conversational guidance to intricate problem-solving, these methods demonstrate the effective fusion of modern technology with classical ideas to bolster the learning and comprehension abilities of LLMs.

\subsection{\highlight{Multi-modal Methods}}
\label{Multi-modal Methods}
\highlight{Building on LMs, some multimodal approaches incorporate visual encoders, enabling the models to handle mathematics problems that rely on visual elements, such as geometry, topology, tabular statistics, etc. GPT-4V \cite{openai2023gpt4} is the first successfully proposed Multimodal LLMs (MLLMs) that demonstrate strong visual reasoning capabilities to handle certain image-based mathematical problems. Subsequently, a lot of other high-performing MLLMs have been introduced, including Gemini \cite{team2023gemini}, Qwen2-VL \cite{Qwen2-VL}, CogVLM \cite{wang2023cogvlm}, GPT-4o \cite{hurst2024gpt}, GLM-4V~\cite{glm2024chatglm} and InternVL \cite{chen2024internvl}, etc. 
}

\highlight{UnAC~\cite{wang2024understanding} introduces a multimodal prompting approach that decomposes the reasoning process into three stages: understanding the image, abstracting image information, and systematically verifying the reasoning steps. Math-LLaVA \cite{shihu2024mathllava} introduces a high-quality, diverse multimodal mathematical dataset called MathV360K, which significantly improves the model's multimodal mathematical reasoning capabilities. LLaVA-o1 \cite{xu2024llava} and AtomThink \cite{xiang2024atomthink} introduce long CoT into the domain of MLLMs. 
M-STAR \cite{anonymous2024diving} investigates the application of self-evolving training in multimodal reasoning by combining continuous self-evolving training, reward model re-ranking, and adaptive exploration strategies.}


%% file: secs/05Datasets.tex
\input{tabs/tab_dataset}

\section{Datasets}
\label{Datasets}
To train and evaluate the arithmetic and math ability of language models, various math word problems (MWPs) datasets \cite{saxton2018analysing,hendrycks2021measuring,cobbe2021training,kushman-etal-2014-learning} are designed.
In this paper, we organize the frequently used datasets for mathematical language models by dividing them into training, benchmark and data augmentation datasets (Table \ref{table:statistis_datasets}).

\subsection{Training Datasets}
\label{Training Datasets}
\subsubsection{Mathematical Calculation}
Several studies have introduced datasets aiming to identify numerical information within text \cite{spithourakis2016numerically,spithourakis2018numeracy,elazar2019large}, specifically targeting the prediction of attribute values.
The Clinical Data \cite{spithourakis2016numerically,spithourakis2018numeracy} consists of clinical records sourced from the London Chest Hospital. Each patient record includes a text report and structured KB tuples detailing 20 potential numeric attributes like age and gender.
Scientific Data \cite{spithourakis2018numeracy} encompasses paragraphs extracted from Cornell’s ARXIV repository, featuring over half a million converted papers across 37 scientific sub-fields.
Additionally, Elazar et al. \cite{elazar2019large} proposed the Distributions over Quantities (DoQ) dataset, comprising empirical counts of scalar attribute values linked to more than 350K nouns, adjectives, and verbs across 10 different attributes.

Furthermore, certain studies consider the relational knowledge between various numbers. For instance, \highlight{VERBPHYSICS \cite{forbes2017verb} aggregates crowdsourced data on actions and objects, including knowledge of grounded object pairs and actions related to them.
More intricately, DROP \cite{dua2019drop} dataset necessitates Discrete Reasoning Over the content of Paragraphs. In this benchmark of 55,000 adversarial questions, a system resolves references within a question across multiple positions and performs operations like addition, counting, or sorting.}

\subsubsection{Math Word Problems}
A large number of datasets are proposed for MWPs.
\highlight{Hosseini et al. \cite{hosseini-etal-2014-learning} curated AddSub,} primarily focusing on addition and subtraction problems. 
This dataset identifies relevant variables and their respective values to translate sentences into problem statements, represented as equations.
\highlight{SingleOp \cite{roy2015reasoning}, encompasses a wider range} of mathematical operations including multiplication and division. Its purpose is to facilitate reasoning about quantities articulated in natural language.
DRAW \cite{upadhyay2015draw} comprises 1000 algebra word problems semi-automatically annotated for evaluating automatic solvers. It features gold coefficient alignments crucial for uniquely identifying equation system derivations.
The Alg514 \cite{kushman-etal-2014-learning} dataset comprises 514 linear algebra problems structured around 28 equation templates.
Meanwhile, MultiArith \cite{roy2015solving} is designed to handle arithmetic problems involving multiple steps and operations, without predefined templates. 
MATHPILE \cite{wang2023generative} is a diverse and high-quality math-centric corpus comprising about 9.5 billion tokens. OpenWebMath \cite{paster2023openwebmath} is an open dataset inspired by
these works containing 14.7B tokens of mathematical webpages from Common Crawl.

Several studies have introduced methods focused on predicting equations through semantic parsing.
\highlight{SingleEq \cite{koncel2015parsing} revolves around} grade-school algebra word problems. It comprises samples that can be correlated to a single equation involving multiple mathematical operations on one variable, structured as a parsing tree. 
In contrast, \highlight{Dolphin1878 \cite{shi2015automatically} contains 1,878 word problems in mathematics}, wherein each sample might encompass multiple equations and variables depicted through multiple trees.
AllArith \cite{roy2017unit} offers a concise representation of the dependencies among number units mentioned in a given problem, referred to as Unit Dependency Graphs (UDGs).
To augment dataset size, Math23k \cite{wang2017deep} gathers 23,161 problems labeled with structured equations and corresponding answers.
GSM8K \cite{cobbe2021training} \highlight{comprises 8.5K high-quality grade school math problems crafted by human problem writers, where the problems typically involve 2 to 8 steps for resolution.}

HMWP \cite{qin2020semantically} aims to enhance dataset diversity by extracting MWPs from a Chinese K12 problem bank. 
This effort sought to validate the universality of math word problem solvers and expand research in MWPs to better align \highlight{with real-world scenarios. 
In total, it comprises 5,491 MWPs,} including 2,955 single-variable linear MWPs, 1,636 two-variable linear MWPs, and 900 single-variable nonlinear MWPs.
Additionally, MathQA \cite{amini2019mathqa} presents a diverse collection of 37,000 English multiple-choice MWPs spanning various mathematical domains. 



Existing research endeavors to enhance the comprehensibility and precision of intricate reasoning by providing fine-grained annotations of reasoning processes \cite{upadhyay2017annotating,ling2017program,chen2021finqa,DBLP:conf/iclr/LightmanKBEBLLS24,zhao2022multihiertt}.
DRAW-1K \cite{upadhyay2017annotating} introduces a novel dataset featuring 1,000 general algebra word problems, each annotated with derivations.
AQUA \cite{ling2017program} structures each question into four components: problem description, multiple-choice answer options, rationale, and correct option label. 
FinQA \cite{chen2021finqa} presents an expert-annotated dataset comprising 8,281 financial QA pairs, meticulously annotated with numerical reasoning processes to ensure comprehensive explication.
PRM800K \cite{DBLP:conf/iclr/LightmanKBEBLLS24} assigns labels (positive, negative, or neutral) to each step in solving MATH problems sampled by a large-scale generator. The training set encompasses 800K step-level labels across 75K solutions to 12K problems.


Additionally, Patel et al. \cite{patel2021nlp} introduced the SVAMP challenge set to establish a more robust evaluation framework for methods designed to tackle elementary-level MWPs.
Alghamdi et al. \cite{alghamdi2022armath} contributed the inaugural large-scale dataset, ArMATH, for Arabic MWPs. This dataset comprises 6,000 samples of primary-school math problems written in Modern Standard Arabic.
Kalyan et al. \cite{kalyan2021much} proposed two datasets, REALFP and SYNTHFP, for a novel task \highlight{called Fermi Problems (FPs). 
This task necessitates the amalgamation of various reasoning abilities, including suitable abstractions, commonsense knowledge, and creative synthesis.}

Recently, mathematical reasoning has incorporated multiple input modalities such as \highlight{textual, visual, and tabular} data \cite{lu2022dynamic,zhu2021tat,zhao2022multihiertt}.
For instance, TabMWP \cite{lu2022dynamic} presents a new dataset comprising 38,431 open-domain grade-level problems that necessitate mathematical reasoning across both textual and tabular data.
Zhu et al. \cite{zhu2021tat} introduced a large-scale QA dataset, TAT-QA, encompassing both tabular and textual data. This dataset often requires numerical reasoning—such as addition, subtraction, multiplication, division, counting, comparison/sorting, and their combinations—based on real financial reports to infer answers.
Additionally, MultiHiertt \cite{zhao2022multihiertt} is a proposed dataset consisting of QA pairs derived from Multi Hierarchical Tabular and \highlight{Textual data, where each document} in this dataset contains multiple tables and longer unstructured texts.


\subsubsection{Theorem Proving}
Current research delves into exploring mathematical theorem datasets (e.g., INT \cite{wu2020int}, Feit-Thompson \cite{huang2018gamepad}, and IsarStep \cite{li2020isarstep}) to develop novel machine learning-based theorem-proving strategies. 
Huang et al. \cite{huang2018gamepad} introduced the Feit-Thompson dataset, encompassing 1,602 lemmas, expanding into 83,478 proof states for the Feit-Thompson theorem \cite{gonthier2013machine}.
IsarStep \cite{li2020isarstep} constitutes a non-synthetic dataset extracted from the largest repository of proofs handwritten by human experts in a theorem prover.
Furthermore, NaturalProofs \cite{welleck2021naturalproofs} comprises 32,000 theorem statements and proofs, 14,000 definitions, and 2,000 other types of pages (e.g., axioms, corollaries). This dataset draws from three domains: broad-coverage data sourced from ProofWiki, \highlight{deep-coverage data from the Stacks, and} low-resource data extracted from mathematics textbooks.


Several formal mathematical libraries \cite{megill2019metamath,rudnicki1992overview,grabowski2015four} encompass a range of formal languages tailored for theorem proving (TP). Examples include MML \cite{grabowski2015four}, Coq \cite{bertot2013interactive}, Lean \cite{han2021proof}, Isabelle \cite{wenzel2008isabelle}, and Hol light \cite{slind2008brief}.
For instance, MML \cite{grabowski2015four}, also known as the Mizar Mathematical Library, was established to systematically construct a centralized, reusable knowledge base reflecting the standard foundations of mathematics, namely classical first-order logic and set theory.
Yang et al. \cite{yang2019learning} compiled the CoqGym dataset, comprising 71,000 human-written proofs from 123 projects developed using the Coq proof assistant.
LeanStep \cite{han2021proof} serves as the tactic proof dataset for the Lean theorem prover, featuring high-level human-written tactics alongside kernel-level proof terms.
LISA \cite{jiang2021lisa}, one of the largest proof corpora for interactive theorem provers, contains 183,000 theorems and 2.16 million proof steps. This dataset is extracted from the Archive of Formal Proofs and the standard library of Isabelle.
Kaliszyk et al. \cite{kaliszyk2016holstep} introduced the HolStep dataset rooted in Higher-Order Logic (HOL) proofs.
HOList \cite{bansal2019holist} includes almost 30,000 theorems and proofs across three corpora: core, complex, and flyspeck. The core corpus contains fundamental theorems necessary for defining tactics, while the complex corpus consists of theorems related to complex calculus. Flyspeck contains the majority of the lemmas and theorems related to the Kepler conjecture \cite{hales2017formal}.

\subsection{Benchmark Datasets}
\label{Benchmark Datasets}
Various benchmark datasets have been introduced to assess the performance of different mathematical algorithms, such as MAWPS \cite{koncel2016mawps}.
Huang et al. \cite{huang2016well} \highlight{developed Dolphin18K, a diverse} collection containing over 18,000 annotated MWPs.
\highlight{Mathematics \cite{saxton2018analysing} encompasses} problems spanning arithmetic, algebra, probability, and calculus. These problems involve sequential questions and answers presented in a free-form textual input/output format.
For assessing arithmetic understanding, Mishra et al. \cite{mishra2022numglue} introduced NumGLUE, a multi-task benchmark that evaluates AI systems across eight distinct tasks.
miniF2F \cite{zheng2021minif2f} is a dataset designed to create a unified cross-system benchmark for neural theorem proving. It comprises 488 formal Olympiad-level mathematics problem statements, encompassing Metamath, Lean, Isabelle, and HOL Light.
\highlight{FOLIO \cite{DBLP:conf/emnlp/HanS0QRZCPQBSWS24} is a human-annotated, logically complex and diverse dataset for reasoning in natural language (NL), equipped with 1,430 first-order logic (FOL) annotations.}

Several datasets consider varying difficulty levels across educational stages, ranging from \highlight{elementary to high school \cite{miao2020diverse,shi2022language,hendrycks2021measuring}.}
Miao et al. \cite{miao2020diverse} \highlight{introduced ASDiv, comprising 2,305 MWPs} categorized by problem type and elementary school grade level.
The Multilingual Grade School Math Benchmark (MGSM) \cite{shi2022language} leverages problems from GSM8K, translating them into 10 languages with the assistance of human annotators.
The MATH dataset \cite{hendrycks2021measuring} contains 12,500 problems sourced from high school math \highlight{competitions, where each problem within MATH is equipped with a step-by-step solution.}

Furthermore, several datasets serve to evaluate foundation models \cite{zhong2023agieval,yuan2023well}.
AGIEval \cite{zhong2023agieval} functions as a human-centric benchmark tailored to assess foundation models' general abilities in tasks related to human cognition and problem-solving. 
Yuan et al. \cite{yuan2023well} introduced the arithmetic dataset MATH 401, designed to \highlight{test the latest large language models. }
\highlight{With the development of multimodal LLMs (MLLMs), benchmarks for multimodal mathematical reasoning are proposed \cite{yue2024mmmumassivemultidisciplinemultimodal,qiao2024wemathdoeslargemultimodal,chernyshev2024umathuniversitylevelbenchmarkevaluating,chen-etal-2024-m3cot,zhang2025mathverse,wang2024measuring}. 
For example, MathVista \cite{lu2023mathvista} is a popular English mathematical benchmark for MLLMs. 
Liu et al. \cite{liu2024cmm} released a Chinese multimodal mathematical reasoning dataset to enhance and measure the performance of Multimodal LLMs.
Moreover, MMMU-Math \cite{yue2024mmmumassivemultidisciplinemultimodal} and U-Math \cite{chernyshev2024umathuniversitylevelbenchmarkevaluating} collect university-level problems to further evaluate the performance.
}

\subsection{Augmented Datasets}
\label{Augmented Datasets}
Augmented datasets serve to enhance existing datasets by incorporating additional samples or information.
Roy et al. \cite{roy2018mapping} introduced Aggregate by augmenting the AllArith dataset with 661 word problems from Perturb. Perturb contains problems that were manually pruned either for not yielding the desired solution $a$ or being too dissimilar from the input problem $p$.
Large-scale language models like ChatGPT and GPT-4 have been utilized to expand datasets. For instance, MetaMathQA \cite{yu2023metamath} rephrases questions from multiple perspectives without introducing extra knowledge to bootstrap mathematical questions.
The Math5K dataset \cite{li2023camel} comprises 50,000 problem-solution pairs generated using GPT-4. These datasets leverage advanced language models to enrich the existing collection of problems and solutions.

Several studies have expanded existing datasets by including supplementary information, such as code programs \cite{austin2021program,mishra2022lila} and explanations \cite{kim2022ept}.
MathQA-Python \cite{austin2021program} is a Python adaptation of the MathQA benchmark, comprising 23,914 problems. This dataset aims to evaluate models' capabilities in synthesizing code from intricate textual descriptions.
Lila \cite{mishra2022lila} is a comprehensive mathematical reasoning benchmark composed of 23 diverse tasks. It extends 20 existing datasets by gathering task instructions and solutions as Python programs, offering not only correct answers but also explainable solutions.
PEN \cite{kim2022ept} focuses on providing plausible and accurate explanations for solving algebraic word problems present in three benchmark datasets: ALG514, DRAW-1K, and MAWPS.
Moreover, some studies have contributed statements and proofs based on existing datasets for theorem proving \cite{jiang2022draft,welleck2022naturalprover}.
For example, miniF2F+informal \cite{jiang2022draft} is a dataset consisting of manually curated informal statements and proofs aligned with formal statements from the miniF2F dataset.
NaturalProofs-Gen \cite{welleck2022naturalprover} adapts data from NATURALPROOFS, including theorem statements, proofs, definitions, and additional pages sourced from ProofWiki. Each proof follows a multi-step structure and references a variable number of sources.

%% file: tabs/tab_dataset.tex
\begin{table*}[htp]
\caption{{\scriptsize The statistics information of mathematical datasets.
\textbf{Level}: \encircle[fill=MyGreenv2, text=white]{E} = \underline{E}lementary, \encircle[fill=myBlue, text=white]{M} = \underline{M}iddle School, \encircle[fill=myYellowv2, text=white]{H} = \underline{H}igh School, \encircle[fill=myOrangev2, text=white]{U} = \underline{U}niversity, \encircle[fill=myViolet, text=white]{C} = \underline{C}ompetition, \encircle[fill=MyRedv2, text=white]{H} = \underline{H}ybrid, \encircle[fill=myGreen, text=white]{O} = \underline{O}thers.
\textbf{Modality}: \encircle[fill=DarkYellow, text=white]{T} = \underline{T}ext, \encircle[fill=DarkRed, text=white]{M} = \underline{M}ultimodal}}
\label{table:statistis_datasets}
\vspace{-3mm}
\centering
\scriptsize
\setlength{\tabcolsep}{2.0pt}{ 
\begin{tabular}{lcccccccccc}
\hlineB{4}
Dataset & \#Train & \#Val & \#Test & \#Total & Language & Task  & Type & Solution & Level & Modality\\ \hline
VERBPHYSICS \cite{forbes2017verb} & 733 & 1,096 & 1,828 & 3,657 & EN   &  Calculation & Training    &  Formula  & - & \encircle[fill=DarkYellow, text=white]{T} \\
Clinical \cite{spithourakis2016numerically,spithourakis2018numeracy} & 11,170 & 1,625 & 3,220 & 16,015 & EN   &  Calculation & Training    &  Formula  & \encircle[fill=myGreen, text=white]{O} & \encircle[fill=DarkYellow, text=white]{T}\\
Scientific \cite{spithourakis2018numeracy} & 14,694 & 2,037 & 4,231 &  20,962 & EN   &  Calculation & Training    &  Formula  & \encircle[fill=myGreen, text=white]{O} & \encircle[fill=DarkYellow, text=white]{T}\\
DoQ \cite{elazar2019large} &   587      &   5,418    &     6,007   &   12,012  &     EN   &  Calculation & Training    &  Formula & - & \encircle[fill=DarkYellow, text=white]{T}\\
DROP \cite{dua2019drop} &    77,409     &   9,536    &     9,622   &   96,567  &     EN   &  Calculation & Training    &  Text & \encircle[fill=MyGreenv2, text=white]{E} & \encircle[fill=DarkYellow, text=white]{T}\\
AddSub \cite{hosseini-etal-2014-learning} &     -    &    -   &   -     &    395  &   EN    &  MWP  & Training    & Formula   &  \encircle[fill=MyGreenv2, text=white]{E} & \encircle[fill=DarkYellow, text=white]{T}\\
SingleOp \cite{roy2015reasoning} &     265    &    107   &     159   &    531  &    EN    & MWP &   Training   & Formula  & \encircle[fill=MyGreenv2, text=white]{E} & \encircle[fill=DarkYellow, text=white]{T}\\
SingleEq \cite{koncel2015parsing} &     -    &     -  &      -  &   508   &   EN     &  MWP &     Training & Formula & \encircle[fill=MyGreenv2, text=white]{E} & \encircle[fill=DarkYellow, text=white]{T}\\
MultiArith \cite{roy2015solving} &    420     &     -  &    180    &   600   &    EN    & MWP  &    Training  &  Formula & - & \encircle[fill=DarkYellow, text=white]{T}\\
Alg514 \cite{kushman-etal-2014-learning}  &   -      &    -   &   -     &    514 &     EN   & MWP  &   Training   &  Formula & - & \encircle[fill=DarkYellow, text=white]{T}\\
Math23k \cite{wang2017deep} &    22,162     &    -   &   1000    &    23,162  &   CH     & MWP &     Training  & Formula & \encircle[fill=MyGreenv2, text=white]{E} & \encircle[fill=DarkYellow, text=white]{T}\\
AQuA \cite{ling2017program} &    97,467     &   254    &    254    &    97,975  &    EN    &  MWP &      Training  & Text & \encircle[fill=myOrangev2, text=white]{U} & \encircle[fill=DarkYellow, text=white]{T}\\
GSM8K \cite{cobbe2021training} &    7,473     &       &  1,319      &   8,792 &     EN   &  MWP & Training  & Text & \encircle[fill=MyGreenv2, text=white]{E}  & \encircle[fill=DarkYellow, text=white]{T}\\
SVAMP \cite{patel2021nlp} &    700     &       &   300     &   1,000  &    EN    &  MWP &     Training &  Formula & \encircle[fill=MyGreenv2, text=white]{E} & \encircle[fill=DarkYellow, text=white]{T}\\
DRAW \cite{upadhyay2015draw}  &   -      &    -   &   -     &   1,000  &    EN    & MWP &   Training   &   Formula & \encircle[fill=myGreen, text=white]{O} & \encircle[fill=DarkYellow, text=white]{T}\\
Dolphin1878 \cite{shi2015automatically}  &      -   &    374   &    1,504    &   1,878  &     EN   & MWP &  Training  & Formula & - & \encircle[fill=DarkYellow, text=white]{T}\\
HMWP \cite{qin2020semantically}  &   -      &  -     &     -   &  5,470   &     CH   & MWP &  Training  &  Formula & - & \encircle[fill=DarkYellow, text=white]{T}\\
ArMATH \cite{alghamdi2022armath}  &   -      &   -    &     -   & 6,000    &    Arabic    & MWP &   Training  &  Formula & \encircle[fill=MyGreenv2, text=white]{E}  & \encircle[fill=DarkYellow, text=white]{T}\\
TabMWP \cite{lu2022dynamic}  &     -    &    -   &    -    &    38,431 &     EN   & MWP &  Training   &  Text &  \encircle[fill=MyGreenv2, text=white]{E} \hspace{1pt} \encircle[fill=myBlue, text=white]{M} & \encircle[fill=DarkYellow, text=white]{T}\\
TAL-SCQ5K \footnote{https://github.com/math-eval/TAL-SCQ5K}  &    3,000     &  -   &     2,000     &   5,000  &   CH/EN     & MWP &   Training      &   Text & \encircle[fill=myViolet, text=white]{C} & \encircle[fill=DarkYellow, text=white]{T}\\
REALFP \cite{kalyan2021much}  &    185     &    185   &     558   &  928  &   EN     & MWP  &   Training      &    Formula  & \encircle[fill=myViolet, text=white]{C} & \encircle[fill=DarkYellow, text=white]{T}\\
SYNTHFP \cite{kalyan2021much}  &    10,000     &   1,000    &   1,000     &  12,000  &   EN     &  MWP &      Training   &    Formula & \encircle[fill=myGreen, text=white]{O}  & \encircle[fill=DarkYellow, text=white]{T}\\
MultiHiertt \cite{zhao2022multihiertt}  &     7,830     &    1,044   &     1,566    &  10,440  &    EN    &  MWP &    Training    &  Formula & \encircle[fill=myGreen, text=white]{O} & \encircle[fill=DarkRed, text=white]{M}\\ 
MATHPILE \cite{wang2023generative} &     -     &    -   &     -    &  903,180  &    EN    &  MWP &    Training    &  Text & \encircle[fill=MyRedv2, text=white]{H} & \encircle[fill=DarkYellow, text=white]{T}\\ 
OpenWebMath \cite{paster2023openwebmath}  &     -    &     -  &   -     &  -   &   EN     & MWP  &    Training   &   Formula  &  \encircle[fill=MyRedv2, text=white]{H} & \encircle[fill=DarkYellow, text=white]{T}\\
MathQA \cite{amini2019mathqa} &    29,837     &    4,475   &   28,985     &  37,297   &     EN   & \highlight{MQA} &    Training  & Formula & - & \encircle[fill=DarkYellow, text=white]{T}\\
FinQA \cite{chen2021finqa}  &    6,251     &   883    &    1,147    &  8,281  &    EN    &  \highlight{MQA}   &    Training    &   Formula  & \encircle[fill=myGreen, text=white]{O} & \encircle[fill=DarkYellow, text=white]{T}\\
TAT-QA \cite{zhu2021tat}  &     -    &   -    &    -    &  16,552  &      EN  &  \highlight{MQA} &     Training     &   Text & \encircle[fill=myGreen, text=white]{O}& \encircle[fill=DarkRed, text=white]{M}\\
MML \cite{grabowski2015four}  &     -    &     -  &   -     &  57,882   &   EN     & TP  &    Training   &   Formula &  -  & \encircle[fill=DarkYellow, text=white]{T}\\
HolStep \cite{kaliszyk2016holstep}  &    2,013,046     &    -   &   196,030     &  2,209,076   &    EN    & TP  &     Training & Formula   &   -  & \encircle[fill=DarkYellow, text=white]{T}\\
CoqGym \cite{yang2019learning}  &      -   & -    &  -      &  71,000   &     EN   & TP  &   Training    &    Formula  & - & \encircle[fill=DarkYellow, text=white]{T}\\
HOList \cite{bansal2019holist}  &    -     &    -   &    -    &  29,462   &     EN   & TP  &    Training    &   Formula  & - & \encircle[fill=DarkYellow, text=white]{T}\\
IsarStep \cite{li2020isarstep}  &    820,000     &   5,000    &  5,000      &  830,000   &   EN     & TP  &   Training   &   Formula    & \encircle[fill=myOrangev2, text=white]{U} & \encircle[fill=DarkYellow, text=white]{T}\\
LISA \cite{jiang2021lisa}  &     -    &   -    &   -     &  183,000   &    EN    & TP  &    Training    &   Formula & - & \encircle[fill=DarkYellow, text=white]{T}\\
NaturalProofs \cite{welleck2021naturalproofs}  &         &       &        &  32,000   &     EN   & TP  &    Training   &   Text   & \encircle[fill=myGreen, text=white]{O} & \encircle[fill=DarkYellow, text=white]{T}\\
LeanStep \cite{han2021proof}  &    -     &     -  &     -   &  21,606,000   &    EN    & TP  &      Training    &   Formula & - & \encircle[fill=DarkYellow, text=white]{T}\\
\hline
NumGLUE \cite{mishra2022numglue}  &   -    &    -   &    -    &  101,835   &     EN   &  Calculation &   Benchmark   &  Text  & \encircle[fill=MyRedv2, text=white]{H} & \encircle[fill=DarkYellow, text=white]{T}\\
Dophin18k \cite{huang2016well} &    -     &   -    &     -   &   18,460   &     EN   &   MWP &   Benchmark  & Text & \encircle[fill=MyGreenv2, text=white]{E} & \encircle[fill=DarkYellow, text=white]{T}\\
MAWPS \cite{koncel2016mawps} &    -     &    -   &      -  &    3,320  &    EN    &  MWP &     Benchmark  &  Formula & \encircle[fill=MyRedv2, text=white]{H} & \encircle[fill=DarkYellow, text=white]{T}\\
ASDiv \cite{miao2020diverse} &    -     &   -    &    -    &   2,305  &   EN     &  MWP &      Benchmark & Formula & \encircle[fill=MyGreenv2, text=white]{E} & \encircle[fill=DarkYellow, text=white]{T}\\
MATH \cite{hendrycks2021measuring} &     7,500    &       &  5,000      &   12,500  & EN       & MWP &     Benchmark  &  Text & \encircle[fill=myYellowv2, text=white]{H} \hspace{1pt} \encircle[fill=myGreen, text=white]{O} & \encircle[fill=DarkYellow, text=white]{T}\\
MGSM \cite{shi2022language}  &    -     &  -     &   -     &  -   &     Multilingual   & MWP &   Benchmark   &  Text & \encircle[fill=MyGreenv2, text=white]{E} & \encircle[fill=DarkYellow, text=white]{T}\\
Mathematics \cite{saxton2018analysing}  &   2,000,000      &     &     100,000     &   2,100,000   &    EN    & MWP &    Benchmark  &    Formula & - & \encircle[fill=DarkYellow, text=white]{T}\\
MMLU-Math \cite{hendryckstest2021}  &      -   &    -   &    -    &   906  &    EN    &  WMP &     Benchmark  & Formula & \encircle[fill=MyGreenv2, text=white]{E} \hspace{1pt} \encircle[fill=myYellowv2, text=white]{H} \hspace{1pt} \encircle[fill=myViolet, text=white]{C} & \encircle[fill=DarkYellow, text=white]{T}\\
AGIEval \cite{zhong2023agieval}  &    -     &    -   &    -    &  469/220   &    CH/EN    & MWP  &  Benchmark     &    Formula & \encircle[fill=myYellowv2, text=white]{H} \hspace{1pt} \encircle[fill=myViolet, text=white]{C} \hspace{1pt} \encircle[fill=myGreen, text=white]{O}  & \encircle[fill=DarkYellow, text=white]{T}\\
\highlight{AIME\footnote{https://www.kaggle.com/datasets/hemishveeraboina/aime-problem-set-1983-2024}} &    -     &    -   &     -   &  \highlight{933}   &    \highlight{EN}    & \highlight{MWP}  &     \highlight{Benchmark}    &   \highlight{Formula} & \encircle[fill=myViolet, text=white]{C} & \encircle[fill=DarkYellow, text=white]{T}\\
\highlight{MATHTRAP \cite{zhao2024exploring}} & - & - & - & \highlight{105/155} & \highlight{EN} & \highlight{MWP} & \highlight{Benchmark} & \highlight{Formula} & \encircle[fill=MyRedv2, text=white]{H} & \encircle[fill=DarkYellow, text=white]{T}\\
\highlight{MathVista \cite{lu2023mathvista}} & - & - & - & \highlight{6,141} & \highlight{EN} & \highlight{MWP} & \highlight{Benchmark} & \highlight{Formula} & \encircle[fill=MyGreenv2, text=white]{E} \hspace{1pt} \encircle[fill=myBlue, text=white]{M} \hspace{1pt} \encircle[fill=myYellowv2, text=white]{H} \hspace{1pt} \encircle[fill=myOrangev2, text=white]{U} & \encircle[fill=DarkRed, text=white]{M} \\
\highlight{Math-V \cite{wang2024measuring}} & - & - & - & \highlight{3,040} & \highlight{EN} & \highlight{MWP} & \highlight{Benchmark} & \highlight{Formula} & \encircle[fill=MyGreenv2, text=white]{E} \hspace{1pt} \encircle[fill=myYellowv2, text=white]{H}  \hspace{1pt} \encircle[fill=myGreen, text=white]{O} & \encircle[fill=DarkRed, text=white]{M} \\
\highlight{CMM-Math \cite{liu2024cmm} } &     \highlight{22,248}    &   -    &     \highlight{5,821}   &  \highlight{28,069}  &   \highlight{CH}    &  \highlight{MWP}   &  \highlight{Benchmark}    &    \highlight{Text} & \encircle[fill=MyGreenv2, text=white]{E} \hspace{1pt} \encircle[fill=myBlue, text=white]{M} \hspace{1pt} \encircle[fill=myYellowv2, text=white]{H} & \encircle[fill=DarkRed, text=white]{M} \\
\highlight{MathVerse \cite{zhang2025mathverse}} & - & - & - & \highlight{2,612} & \highlight{EN} & \highlight{MWP} & \highlight{Benchmark} & \highlight{Formula} & \encircle[fill=MyRedv2, text=white]{H} & \encircle[fill=DarkRed, text=white]{M} \\
\highlight{GeoQA \cite{chen2021geoqa}} & \highlight{3,499}  & \highlight{745}  & \highlight{754} & \highlight{4,998} & \highlight{EN} & \highlight{MWP} & \highlight{Benchmark} & \highlight{Formula} & \encircle[fill=myBlue, text=white]{M} & \encircle[fill=DarkRed, text=white]{M} \\
\highlight{M3CoT \cite{chen-etal-2024-m3cot}} & - & - & - & \highlight{11,459} & \highlight{EN} & \highlight{MWP} & \highlight{Benchmark} & \highlight{Formula} & \encircle[fill=myGreen, text=white]{O} & \encircle[fill=DarkRed, text=white]{M} \\
\highlight{MR-MATH \cite{xia2024evaluating}} & - & - & - & - & \highlight{EN} & \highlight{MWP} & \highlight{Benchmark} & \highlight{Formula} & \encircle[fill=MyGreenv2, text=white]{E} & \encircle[fill=DarkYellow, text=white]{T}\\
\highlight{MMMU-Math \cite{yue2024mmmumassivemultidisciplinemultimodal}} & - & - & - & \highlight{505} & \highlight{EN} & \highlight{MWP} & \highlight{Benchmark} & \highlight{Text} & \encircle[fill=myOrangev2, text=white]{U} & \encircle[fill=DarkRed, text=white]{M} \\
\highlight{We-Math \cite{qiao2024wemathdoeslargemultimodal}} & - & - & - & \highlight{1700} & \highlight{EN} & \highlight{MWP} & \highlight{Benchmark} & \highlight{Formula} & \encircle[fill=MyGreenv2, text=white]{E} \hspace{1pt} \encircle[fill=myYellowv2, text=white]{H}  \hspace{1pt} 
 \encircle[fill=myOrangev2, text=white]{U} & \encircle[fill=DarkRed, text=white]{M} \\
\highlight{U-MATH \cite{chernyshev2024umathuniversitylevelbenchmarkevaluating}}  &     -    &   -    &   -     &  \highlight{1,080}   &     \highlight{EN}   & \highlight{MWP}  &    \highlight{Benchmark}      &   \highlight{Formula} & \encircle[fill=myOrangev2, text=white]{U} & \encircle[fill=DarkRed, text=white]{M} \\
INT \cite{wu2020int}  &     -    &  -     &     -   &  -   &   EN     & TP  &     Benchmark     &  Formula & - & \encircle[fill=DarkYellow, text=white]{T}\\
miniF2F \cite{zheng2021minif2f}  &     -    &   244    &   244     &  488   &     EN   & TP  &    Benchmark      &   Formula & \encircle[fill=MyGreenv2, text=white]{E} \hspace{1pt} \encircle[fill=myYellowv2, text=white]{H}  \hspace{1pt} \encircle[fill=myGreen, text=white]{O} & \encircle[fill=DarkYellow, text=white]{T}\\
\highlight{FOLIO \cite{DBLP:conf/emnlp/HanS0QRZCPQBSWS24}}  &    -     &     -  &     -   &  \highlight{1,430}   &    \highlight{EN}    & \highlight{TP}  &   \highlight{Benchmark}    & \highlight{Formula} & \encircle[fill=MyRedv2, text=white]{H} & \encircle[fill=DarkYellow, text=white]{T}\\
\hline
Aggregate \cite{roy2018mapping}  &     -    &    -   &   -     &   1,492  &    EN    & MWP &    Augmented   &  Formula & \encircle[fill=MyRedv2, text=white]{H} & \encircle[fill=DarkYellow, text=white]{T}\\
MathQA-Python \cite{austin2021program}  &   19,209      &   2,822    &    2,822    &  23,914   &    EN    &  MWP &   Augmented   &  Code & - & \encircle[fill=DarkYellow, text=white]{T}\\
Math50k \cite{li2023camel}  &     -    &    -   &   -     &   50,000  &    EN    &  WMP &  Augmented    &   Text &  -& \encircle[fill=DarkYellow, text=white]{T}\\
PRM800K \cite{DBLP:conf/iclr/LightmanKBEBLLS24}  &      -   &    -   &   -     &   2,868  &    EN    &   WMP &    Augmented  &   Text & \encircle[fill=myYellowv2, text=white]{H} \hspace{1pt} \encircle[fill=myGreen, text=white]{O} & \encircle[fill=DarkYellow, text=white]{T}\\
MetaMathQA \cite{yu2023metamath}  &   -      &    -   &    -    &    395,000  &     EN   &  MWP &  Augmented   &   Text & \encircle[fill=MyRedv2, text=white]{H}  & \encircle[fill=DarkYellow, text=white]{T}\\
Lila \cite{mishra2022lila}  &      -   &    -   &    -    &   134,000  &     EN   &  MWP &  Augmented   & Code  & \encircle[fill=MyRedv2, text=white]{H} & \encircle[fill=DarkYellow, text=white]{T}\\
PEN \cite{kim2022ept}  &     -    &     -  & -       &   3,581   &     EN   & MWP &   Augmented   &  Formula & \encircle[fill=MyRedv2, text=white]{H} & \encircle[fill=DarkYellow, text=white]{T}\\
miniF2F+informal \cite{jiang2022draft}  &    -     &    244   &     244   &  488   &    EN    & TP  &     Augmented    &   Formula & \encircle[fill=MyGreenv2, text=white]{E} \hspace{1pt} \encircle[fill=myYellowv2, text=white]{H}  \hspace{1pt} \encircle[fill=myGreen, text=white]{O} & \encircle[fill=DarkYellow, text=white]{T}\\
NaturalProofs-Gen \cite{welleck2022naturalprover}  &     12,500    &  1,000     &   1,000     &  14,500   &    EN    & TP  &      Augmented &    Text  & \encircle[fill=myGreen, text=white]{O} & \encircle[fill=DarkYellow, text=white]{T}\\
\hlineB{4}
\end{tabular}}
\end{table*}

%% file: secs/06Disscusion.tex
\input{tabs/method_compare}

\input{tabs/method_results}

\section{Analysis and Discussion}
\label{Analysis and Discussion}
\highlight{To gain a deeper understanding of the advantages of typical mathematical LMs, we conduct a comprehensive comparison of their characteristics and performance (Table \ref{table: comparison of mathematical LMs} and Table \ref{tab: performance}).
}

\highlight{
Table \ref{table: comparison of mathematical LMs} provides a detailed overview of mathematical LLMs, their architectures, training strategies, reasoning capabilities, and other advanced features. 
From the table, we observe the following findings.
First, CoT has become a critical enabler for mathematical reasoning.
Additionally, with the introduction of OpenAI's o1 model, an increasing number of studies are focusing on Long CoT.
Second, only a few models (e.g., LLaVA, M-STAR) support multimodal input (text + image). This indicates that most mathematical LLMs focus solely on text-based reasoning. As tasks increasingly require interpreting diagrams and visual data, multimodal models could play a more prominent role in future developments.
Third, models like Toolformer and LeanReasoner integrate external tools (e.g., calculators or symbolic solvers). These tools complement the LLMs' reasoning by handling computations or symbolic operations that the models alone might struggle with.
Fourth, reinforcement learning (RL) is primarily used to refine reasoning strategies and align model outputs with user expectations or task requirements. However, only a small subset of models uses RL techniques, such as Direct Preference Optimization (DPO) \cite{rafailov2024direct} or Proximal Policy Optimization (PPO) \cite{schulman2017proximal}, due to the instability of RL.
Fifth, many recent models (e.g., CoT, BoostedPrompt, PoT) rely on prompt engineering to tackle complex problems without requiring heavy computational resources for fine-tuning. Fine-tuning is essential for most advanced mathematical models (e.g., WizardMath, EURUS) to enhance mathematical reasoning capabilities. 
}

\highlight{
Table \ref{tab: performance} compares the performance of various mathematical LMs across different datasets. The methods can be classified into two categories: Text-based Methods and Multi-modal Methods. Such benchmarks are widely recognized and serve to assess mathematical capabilities in diverse contexts. Here are some conclusions based on the data:
First, o1 stands out as the most accurate model across most datasets, achieving the highest scores in MATH (94.8\%), MMLU (92.3\%), AIME (83.3\%), MMMU (78.1\%) and MathVista (73.9\%). 
It indicates that slow thinking with long CoT improves the performance of mathematical reasoning effectively.
Second, methods like EURIUS and PoT show dataset-specific strengths, suggesting that tailored approaches can yield better results for certain problem types (e.g., formal proofs, complex reasoning). For example, CR achieves the best performance on the FOLIO Dataset with 98.0\%, highlighting its capability for handling formal logic problems.
Third, some Multi-modal benchmarks, such as CMM-Math \cite{liu2024cmm} and Math-V \cite{wang2024measuring}, show relatively low accuracy across all methods, indicating a need for further research and improvement in these areas.  
}

%% file: tabs/method_compare.tex
\begin{table*}[]
\scriptsize
\caption{Comparison of typical mathematical LMs. Pre-T: Pre-Training, FT: Finetuning, MM: Multimodal, Symbolic-S: Symbolic solvers, F-CoT: Foundational CoT, A-CoT: Advanced CoT, RL: Reinforce Learning}
\label{table: comparison of mathematical LMs}
\begin{tabular}{lccccccc}
\hlineB{4}
   Methods     & Base Model & Training & Tool & CoT & RL & MM  & Access \\
\hline
GPT-$f$~\cite{polu2020generative}  & - &  Pre-T  & Symbolic-S  & \textcolor{red}{\xmark} & \textcolor{red}{\xmark}  & \textcolor{red}{\xmark} & \textcolor{red}{\xmark} \\
LISA~\cite{jiang2021lisa}  & -    &   Pre-T   &  Symbolic-S  &  \textcolor{red}{\xmark} & \textcolor{red}{\xmark} &   \textcolor{red}{\xmark}    & \textcolor{red}{\xmark} \\
MATH-PLM~\cite{hendrycks2021measuring} &  GPT-2, GPT-3 &  Pre-T &   \textcolor{red}{\xmark} &  F-CoT &  \textcolor{red}{\xmark}  &  \textcolor{red}{\xmark}  & \textcolor{red}{\xmark} \\
Minerva~\cite{lewkowycz2022solving} & PaLM  &  Pre-T  & \textcolor{red}{\xmark} &  \textcolor{red}{\xmark}  &  \textcolor{red}{\xmark} & \textcolor{red}{\xmark}  & \textcolor{red}{\xmark} \\
MWP-BERT~\cite{liangMWPBERTNumeracyaugmentedPretraining2022} & BERT &  Pre-T  &  \textcolor{red}{\xmark}  &  \textcolor{red}{\xmark}  &  \textcolor{red}{\xmark} & \textcolor{red}{\xmark}  & \textcolor{red}{\xmark} \\
ControlMath~\cite{chen2024controlmath} &  LLaMA-2, Mistral  &  FT   & \textcolor{red}{\xmark} &  F-CoT  &  \textcolor{red}{\xmark}  &  \textcolor{red}{\xmark}  &  \textcolor{red}{\xmark} \\
EURUS~\cite{yuan2024advancing} &  Mistral, CodeLLaMA   &  FT   & \textcolor{red}{\xmark} &  F-CoT   &  DPO, KTO, NCA   &  \textcolor{red}{\xmark}  &  \textcolor{green}{\cmark} \\
MathGLM \cite{yang2023gpt} & GLM  & FT &   \textcolor{red}{\xmark} & \textcolor{red}{\xmark} &  \textcolor{red}{\xmark}  &  \textcolor{red}{\xmark}  & \textcolor{green}{\cmark} \\
WizardLM~\cite{xu_wizardlm_2023}& LLaMA  & FT &    \textcolor{red}{\xmark} & \textcolor{red}{\xmark} & \textcolor{red}{\xmark}  &  \textcolor{red}{\xmark} & \textcolor{green}{\cmark} \\
Wizardmath~\cite{luo2023wizardmath} & LLaMA-2  &  FT  & \textcolor{red}{\xmark}  & \textcolor{red}{\xmark} & PPO & \textcolor{red}{\xmark} & \textcolor{green}{\cmark} \\
PaLM 2-L-Math~\cite{liu_improving_2023} & PaLM-2-L &  FT & \textcolor{red}{\xmark} &  F-CoT &  \textcolor{red}{\xmark}  & \textcolor{red}{\xmark} & \textcolor{red}{\xmark} \\
Codex-math~\cite{drori2022neural} & Codex &  Prompt & Program & \textcolor{red}{\xmark} &  \textcolor{red}{\xmark}  & \textcolor{red}{\xmark}  & \textcolor{red}{\xmark} \\
Toolformer~\cite{schick2023toolformer} & GPT-J  &  FT &  Calculator  & \textcolor{red}{\xmark} &  \textcolor{red}{\xmark}  &  \textcolor{red}{\xmark}  & \textcolor{red}{\xmark} \\
LeanReasoner~\cite{jiang2024leanreasoner} & CodeLLaMA &  FT  & Symbolic-S & \textcolor{red}{\xmark} &  \textcolor{red}{\xmark}  & \textcolor{red}{\xmark} & \textcolor{green}{\cmark} \\
LINC~\cite{olausson2023linc} & StarCoder+  &  Prompt  & Symbolic-S  &  F-CoT  & \textcolor{red}{\xmark} & \textcolor{red}{\xmark} & \textcolor{green}{\cmark} \\
MuMath-Code~\cite{yin2024mumath} & Llama-2 &  FT & Program & \textcolor{red}{\xmark} & \textcolor{red}{\xmark} & \textcolor{red}{\xmark} &  \textcolor{green}{\cmark} \\
MAmmoTH~\cite{yue2023mammoth} & - & FT & Program & F-CoT & \textcolor{red}{\xmark} & \textcolor{red}{\xmark} & \textcolor{red}{\xmark} \\
InternLM-Math \cite{ying2024internlmmath} & InternLM2 & Pre-T, FT & Program &  F-CoT & PPO & \textcolor{red}{\xmark} & \textcolor{green}{\cmark} \\
Qwen-Math \cite{yang2024qwen2} & - & Pre-T, FT & Program &  F-CoT & PPO & \textcolor{red}{\xmark} & \textcolor{green}{\cmark} \\
PAL~\cite{gao2023pal} & -  & Prompt & Program & F-CoT & \textcolor{red}{\xmark} & \textcolor{red}{\xmark} & \textcolor{green}{\cmark} \\
CoT~\cite{wei2022chain} & -  & Prompt & \textcolor{red}{\xmark} & F-CoT & \textcolor{red}{\xmark} & \textcolor{red}{\xmark} & \textcolor{green}{\cmark} \\
PromptPG-CoT~\cite{lu2022dynamic} & - & Prompt & \textcolor{red}{\xmark} & \textcolor{red}{\xmark} & \textcolor{red}{\xmark} & \textcolor{red}{\xmark} & \textcolor{green}{\cmark} \\
BoostedPrompt~\cite{pitis_boosted_2023} & - & Prompt & \textcolor{red}{\xmark} & F-CoT & \textcolor{red}{\xmark} & \textcolor{red}{\xmark} & \textcolor{red}{\xmark}\\
ToT~\cite{yao2023tree} & - & Prompt & \textcolor{red}{\xmark} & A-CoT & \textcolor{red}{\xmark} & \textcolor{red}{\xmark} &  \textcolor{green}{\cmark} \\
PoT~\cite{chen2022program} & - & Prompt & Program &  F-CoT & \textcolor{red}{\xmark}  & \textcolor{red}{\xmark} & \textcolor{green}{\cmark} \\
Self-check~\cite{miao2023selfcheck} & - & Prompt & \textcolor{red}{\xmark} & A-CoT & \textcolor{red}{\xmark} & \textcolor{red}{\xmark} & \textcolor{green}{\cmark} \\
Self-Consistency~\cite{wang2022self}& - & Prompt & \textcolor{red}{\xmark} & A-CoT & \textcolor{red}{\xmark} & \textcolor{red}{\xmark} & \textcolor{green}{\cmark} \\
Diversity-of-Thought~\cite{naik2023diversity}& - & Prompt & \textcolor{red}{\xmark} & A-CoT & \textcolor{red}{\xmark} & \textcolor{red}{\xmark} & \textcolor{red}{\xmark} \\
RAP~\cite{hao2023reasoning} &  - & Prompt & \textcolor{red}{\xmark} & A-CoT & \textcolor{red}{\xmark} & \textcolor{red}{\xmark} & \textcolor{green}{\cmark} \\
LATS~\cite{zhou2023language} &  - & Prompt & \textcolor{red}{\xmark} & A-CoT & \textcolor{red}{\xmark} & \textcolor{red}{\xmark} & \textcolor{red}{\xmark} \\
LLM+P~\cite{liu2023llm+} & - & Prompt &  Symbolic-S & A-CoT & \textcolor{red}{\xmark} & \textcolor{red}{\xmark} & \textcolor{green}{\cmark} \\
LLM+DP~\cite{dagan2023dynamic} & - & Prompt &  Symbolic-S & A-CoT & \textcolor{red}{\xmark} & \textcolor{red}{\xmark} & \textcolor{green}{\cmark} \\
ISR-LLM~\cite{zhou2023isr} &  - & Prompt & \textcolor{red}{\xmark} & A-CoT & \textcolor{red}{\xmark} & \textcolor{red}{\xmark} & \textcolor{green}{\cmark} \\
MCR~\cite{yoran2023answering} & - & Prompt & \textcolor{red}{\xmark} & A-CoT & \textcolor{red}{\xmark} & \textcolor{red}{\xmark} & \textcolor{green}{\cmark} \\
Rank-verifier~\cite{cobbe2021training} &  - & Prompt & \textcolor{red}{\xmark} & A-CoT & \textcolor{red}{\xmark} & \textcolor{red}{\xmark} & \textcolor{red}{\xmark} \\
GRACE~\cite{khalifa2023discriminator} & - & Prompt & \textcolor{red}{\xmark} & A-CoT & \textcolor{red}{\xmark} & \textcolor{red}{\xmark} & \textcolor{green}{\cmark} \\
Reflexion~\cite{shinn2023reflexion} & - & Prompt & \textcolor{red}{\xmark} & A-CoT & \textcolor{red}{\xmark} & \textcolor{red}{\xmark} & \textcolor{green}{\cmark} \\
CR~\cite{zhang2023cumulative} & - & Prompt & \textcolor{red}{\xmark} & A-CoT & \textcolor{red}{\xmark} & \textcolor{red}{\xmark} & \textcolor{green}{\cmark} \\
Step-Plan~\cite{zhang2023interpretable} & - & Prompt & \textcolor{red}{\xmark} & A-CoT & \textcolor{red}{\xmark} & \textcolor{red}{\xmark} & \textcolor{green}{\cmark} \\
STaR~\cite{zelikman2022star}& GPT-J & FT & \textcolor{red}{\xmark} & A-CoT & \textcolor{red}{\xmark} & \textcolor{red}{\xmark} & \textcolor{green}{\cmark} \\
V-STaR~\cite{hosseini2024v}& LLaMA2, CodeLLaMA & FT & \textcolor{red}{\xmark} & A-CoT & DPO &  \textcolor{red}{\xmark} & \textcolor{red}{\xmark} \\
Quiet-STaR~\cite{zelikman2024quiet}& Mistral & Pre-T & \textcolor{red}{\xmark} & A-CoT & \textcolor{red}{\xmark} &  \textcolor{red}{\xmark} & \textcolor{red}{\xmark} \\
REFT~\cite{luong2024reft} &  Galactica, CodeLLaMA  & FT &  \textcolor{red}{\xmark}  & A-CoT &  PPO  & \textcolor{red}{\xmark} &  \textcolor{green}{\cmark} \\
SCoRe~\cite{kumar2024training}& Gemini & FT & \textcolor{red}{\xmark} & A-CoT  & SCoRe &  \textcolor{red}{\xmark} & \textcolor{red}{\xmark} \\
HGS-PRM~\cite{ma2023let}& LLaMA2, WizardMath & FT & \textcolor{red}{\xmark} & A-CoT & PPO &  \textcolor{red}{\xmark} & \textcolor{red}{\xmark} \\
MCTSr~\cite{zhang2024accessing}& LLaMa-3 & FT & \textcolor{red}{\xmark} & A-CoT & \textcolor{red}{\xmark} &  \textcolor{red}{\xmark} & \textcolor{green}{\cmark} \\
CoRe~\cite{zhu2022solving}& GPT-J, DeBERTa-large & FT  &  \textcolor{red}{\xmark} & A-CoT & \textcolor{red}{\xmark} &  \textcolor{red}{\xmark} & \textcolor{green}{\cmark} \\
Marco-o1~\cite{zhao2024marco}& Qwen2 &  FT & \textcolor{red}{\xmark} & A-CoT  & \textcolor{red}{\xmark} &  \textcolor{red}{\xmark} & \textcolor{green}{\cmark} \\
Step-DPO~\cite{lai2024step}& Qwen2 &  FT  & \textcolor{red}{\xmark} & A-CoT & DPO &  \textcolor{red}{\xmark} &  \textcolor{green}{\cmark} \\
Flow-DPO~\cite{deng2024flow}& LLaMA3, Phi-3  & FT &  \textcolor{red}{\xmark} & A-CoT & DPO &  \textcolor{red}{\xmark} & \textcolor{red}{\xmark} \\
OmegaPRM~\cite{luo2024improve}&  Gemini Pro, Gemma2 & FT  & \textcolor{red}{\xmark} & A-CoT & PRM &  \textcolor{red}{\xmark} & \textcolor{red}{\xmark} \\
SocraticLLM~\cite{ding2024boosting}& Qwen2 &  FT  & \textcolor{red}{\xmark} & A-CoT & \textcolor{red}{\xmark} &  \textcolor{red}{\xmark} &  \textcolor{green}{\cmark}\\
MATHDIAL~\cite{macina2023mathdial}& T5 &  FT & \textcolor{red}{\xmark} &  A-CoT & \textcolor{red}{\xmark} &  \textcolor{red}{\xmark} & \textcolor{green}{\cmark} \\
QwQ \cite{qwq-32b-preview}& Qwen2.5 &  Pre-T  & \textcolor{red}{\xmark} & Long CoT & \textcolor{green}{\cmark} &  \textcolor{red}{\xmark} &  \textcolor{green}{\cmark} \\
o1 \cite{openai_o1_system_card}  &  - &   Pre-T   &   \textcolor{red}{\xmark} &  Long CoT   &  \textcolor{green}{\cmark}  & \textcolor{green}{\cmark} & \textcolor{red}{\xmark} \\
GPT-4o~\cite{hurst2024gpt}& - & Pre-T  & \textcolor{green}{\cmark} & A-CoT & \textcolor{green}{\cmark} &  \textcolor{green}{\cmark} & \textcolor{red}{\xmark} \\
GPT-4V~\cite{openai2023gpt4}& - & Pre-T  & \textcolor{green}{\cmark}  & A-CoT & \textcolor{green}{\cmark} &  \textcolor{green}{\cmark} &  \textcolor{red}{\xmark}\\
Qwen2-VL~\cite{Qwen2-VL}& Qwen2-VL & Pre-T  & \textcolor{red}{\xmark} & F-CoT & \textcolor{red}{\xmark} &  \textcolor{green}{\cmark} & \textcolor{green}{\cmark} \\
LLaVA-o1~\cite{xu2024llava}&  LLaVA&  FT & \textcolor{red}{\xmark} & Long CoT & \textcolor{red}{\xmark}  &  \textcolor{green}{\cmark} & \textcolor{green}{\cmark} \\
Gemini~\cite{team2023gemini}& Gemini &  Pre-T & \textcolor{green}{\cmark} &  A-CoT & \textcolor{green}{\cmark} &  \textcolor{green}{\cmark} & \textcolor{red}{\xmark} \\
GLM-4V~\cite{glm2024chatglm}& ChatGLM &  Pre-T  & \textcolor{green}{\cmark} & A-CoT & \textcolor{green}{\cmark} &  \textcolor{green}{\cmark} & \textcolor{red}{\xmark} \\
AtomThink~\cite{xiang2024atomthink}& LLaVA, EMOVA & FT  & \textcolor{red}{\xmark} & Long CoT  & PRM  &  \textcolor{green}{\cmark} & \textcolor{green}{\cmark} \\
Math-LLaVA~\cite{shihu2024mathllava}& LLaVA & FT  & \textcolor{red}{\xmark} & F-CoT  &  \textcolor{red}{\xmark} &  \textcolor{green}{\cmark} &  \textcolor{green}{\cmark} \\
M-STAR~\cite{anonymous2024diving}& MiniCPM-V & FT  & \textcolor{red}{\xmark} & A-CoT & PRM &  \textcolor{green}{\cmark} & \textcolor{red}{\xmark} \\
UnAC~\cite{wang2024understanding}& GPT-4V, Gemini, LLaVA &  Prompt & \textcolor{red}{\xmark} & F-CoT & \textcolor{red}{\xmark}  &  \textcolor{green}{\textcolor{green}{\cmark}} & \textcolor{red}{\xmark}  \\
\hlineB{4}
\end{tabular}
\end{table*}

%% file: tabs/method_results.tex
\begin{table*}[]
\scriptsize
\caption{Performance of various mathematical LMs in terms of accuracy.}
\label{tab: performance}
\begin{tabular}{lccccccccc}
\hlineB{4}
Text-based Methods  & MATH & GSM8K & MMLU & ASDiv & AIME & MathQA & SVAMP & FOLIO & AQUA \\
\hline
MATH-PLM~\cite{hendrycks2021measuring} &  6.9 &  - &  -   & - & -  & - & -  & - & - \\
Minerva~\cite{lewkowycz2022solving} & 50.3  & 78.5 & 75.0   & - & - & - & - & - & - \\
MWP-BERT~\cite{liangMWPBERTNumeracyaugmentedPretraining2022} &  -  &  - & -  & - & - & 76.6 & - & - & - \\ 
ControlMath~\cite{chen2024controlmath} & 9.5 & 49.3 &  -  &  - & - & -  & 51.6 & -  & - \\
EURUS~\cite{yuan2024advancing} &  41.7   & 62.8 &  -  &  \textbf{93.0} & - & -  & \textbf{90.4} & -  & - \\
Wizardmath~\cite{luo2023wizardmath} &  22.7  & 81.6  & - & - & - & - & -  & - \\
Codex-math~\cite{drori2022neural} &  81.8 & - & - &  -  & -  & - & -  & - & - \\
Toolformer~\cite{schick2023toolformer} &  - &  -  & - & 40.4 & - & -  & 29.4 & - & - \\
LeanReasoner~\cite{jiang2024leanreasoner} &  -  & - & - &  -  & - & - & - & 82.6 & -\\
LINC~\cite{olausson2023linc} &  -  & -  &  -  & - & - & - & -  & 72.5 & - \\
MuMath-Code~\cite{yin2024mumath} & 55.1 & 90.7 & - & - & - &  - & -  & - & - \\
MAmmoTH~\cite{yue2023mammoth} & 44.2 & 76.7 & - & - & - & - & -  & - & 61.4 \\
InternLM-Math \cite{ying2024internlmmath} & 37.7 &  82.6 &  - & - & - & - & - & - & - \\
Qwen-Math \cite{yang2024qwen2} & 66.8 & 91.6 & 82.8 & - & 63.3 &  86.3 & -  & - & - \\ 
PAL~\cite{gao2023pal} & - & 72.0 & - &  79.6 & - & - & 79.4 & - & - \\
BoostedPrompt~\cite{pitis_boosted_2023} & - & 87.1 & - & - & - & - & - & - & 63.8 \\
PoT~\cite{chen2022program} & - & 80.0 &  - & -  & - & \textbf{89.1} & -  & - & 58.6 \\ 
Self-check~\cite{miao2023selfcheck} & 51.3 & 88.1 & - & - & \textbf{81.2} & - & -  & - & - \\
Self-Consistency~\cite{wang2022self}& - & 78.0 & - & 87.8 & - & - & 86.8  & - & 52.0 \\
Diversity-of-Thought~\cite{naik2023diversity}& - & 96.3 & - & - & - & - & -  & - & \textbf{81.7} \\
GRACE~\cite{khalifa2023discriminator} & - & 36.3 & - & - & - & 84.4 & 68.6  & - & - \\
CR~\cite{zhang2023cumulative} & 72.2 & - & - & - & - & - & -  & \textbf{98.0} & - \\ 
REFT~\cite{luong2024reft} & - &  75.3 & - &  -  & - &  71.8 & 79.2 & - & - \\
SCoRe~\cite{kumar2024training}& 64.4 & - & -  & - &  - & - & -  & - & - \\
HGS-PRM~\cite{ma2023let}&  13.7 & 65.4 & - & - &  - & - & -  & - & - \\
MCTSr~\cite{zhang2024accessing}& 58.2 & \textbf{96.6} & - & - &  11.8 & - & -  & - & - \\
CoRe~\cite{zhu2022solving}& - & 63.2 & - & 90.5 & - & - & - & - & - \\
Step-DPO~\cite{lai2024step}& 70.8 & 94.0 & - & - & - &  - &  - & - & - \\
Flow-DPO~\cite{deng2024flow}&  38.6 &  71.3 & - & - & - &  - & - & - & - \\
OmegapRM~\cite{luo2024improve}& 69.4 & 93.6 & - & - & - &  - & - & - & - \\
QwQ \cite{qwq-32b-preview}& 90.6  & - & - & - & 50.0 &  - & - & - & - \\
GLM-4-9B-Chat \cite{glm2024chatglm}& 50.6 & 79.6 & 72.4 & - & - &  - & - & - & - \\
\hlineB{4}
Multi-modal Methods  & MATH & GSM8K & MMLU & ASDiv & AIME & MMMU & MathVista & CMM-Math & Math-V \\
\hline
o1 \cite{openai_o1_system_card}  & \textbf{94.8}    &   - &  \textbf{92.3}   &  -  & \textbf{83.3} & \textbf{78.1} & \textbf{73.9} & -  & - \\
GPT-4o~\cite{hurst2024gpt} & 60.3  & - & 88.0 & - &  13.4 & 69.2 & 63.8 & 29.02  & 30.4 \\
GPT-4V~\cite{openai2023gpt4} & -  & -  & - & - &  - & 56.8 & 49.9 & -  & 22.8 \\
Qwen2-VL~\cite{Qwen2-VL} &  -  & - & 46.2 & - &  - & 64.5 & 70.5 & \textbf{43.0}  & 25.9 \\
LLaVA-o1~\cite{xu2024llava} &  64.0 & - & - & -  &  - & - & 54.8 & -  & - \\
Gemini~\cite{team2023gemini} &   67.7 & - &  85.9 & - &  - & 62.2 & 63.9 & 41.9  & 17.7 \\
GLM-4V~\cite{glm2024chatglm} &   -  & - & - & - &  - & 47.2 & - & -  & - \\
AtomThink~\cite{xiang2024atomthink} & -  & - & -  & -  &  - & - & 53.3 & -  &  \textbf{40.5} \\
Math-LLaVA~\cite{shihu2024mathllava} &  -  & - & -  &  - &  38.3 &  38.3 & 46.6 & -  & 15.7 \\
M-STAR~\cite{anonymous2024diving} & -  & - & - & - &  - & - & 59.5 & -  & - \\
UnAC~\cite{wang2024understanding} & - & - & - & -  &  - & 59.7 & 56.6 & -  & - \\
\hlineB{4}
\end{tabular}
\end{table*}

%% file: secs/07Challenges.tex
\section{Challenges and Further Directions}
\label{Challenges and Further Directions}
\paragraph{Faithfulness.} 
Mathematical LLMs reside in the phenomena of hallucination and faithfulness, where the generated output may lack factual accuracy or logical grounding \cite{rawte2023survey,ji2023survey}. 
This phenomenon leads to the production of erroneous or misleading mathematical results, undermining the reliability of the model's outputs. 
Some studies explored to address this problem by integrating extra knowledge \cite{he_rethinking_2022}, reinforcement learning from human feedback \cite{DBLP:conf/iclr/LightmanKBEBLLS24}, tools \cite{schick2023toolformer,parisi2022talm}, and verify-based methods \cite{zhou2023solving,ling2023deductive,li2023making,zhao2023verify,he_rethinking_2022,shridhar2023screws}. 
However, the improvement of the hallucination phenomena is limited, which influences the trustworthiness and utility of mathematical language models in practical applications and scholarly pursuits.

\paragraph{Multi-modal.} 
In particular, math problems (e.g., geometry problems \cite{gelernter1960empirical}) involve not only textual information but also various modalities such as diagrams, graphs, or mathematical symbols \cite{lu2021iconqa}. 
While existing models excel in processing textual information, interpreting and reasoning across multiple modalities concurrently remains a formidable task. 
Few multi-modal mathematical datasets and methods are proposed for this task \cite{zhao2022multihiertt,lu2021iconqa}.
Bridging the gap between textual representations and visual/mathematical elements necessitates robust mechanisms that can effectively capture and synthesize information from disparate sources, ultimately leading to comprehensive and accurate problem-solving capabilities. 
In fact, the multi-modal information in mathematics is much more complex than general multi-modal tasks like vision question answering \cite{antol2015vqa} and image captioning \cite{hossain2019comprehensive}.
Achieving proficiency in handling multi-modal mathematical problems stands as a pivotal yet intricate objective in advancing the competency of mathematical LLMs.

\paragraph{Uncertainty.}
The uncertainty of LLMs \cite{gawlikowski2023survey,duan2023shifting} leads to the ambiguity and variability problem of mathematical problems. While these models excel in deterministic tasks, their handling of uncertainty, such as probabilistic reasoning or dealing with incomplete or vague information, poses a significant challenge. 
Mathematical problems often entail nuanced interpretations, fuzzy constraints, or scenarios where a single precise solution might not exist. 
Several studies investigated this problem via controlled generation technologies \cite{DBLP:conf/icml/ZhouJWCS23}.
However, ensuring that LLMs can navigate and appropriately account for these uncertainties while providing accurate and contextually relevant solutions remains a complex task.

\paragraph{Evaluation.} 
It is still a challenge to evaluate mathematical LMs with robust and comprehensive evaluation metrics that adequately capture the models' performance across various mathematical tasks. 
Traditional evaluation metrics in natural language processing might not effectively capture the intricacies of mathematical reasoning and problem-solving. 
Designing evaluation benchmarks \cite{saxton2018analysing,hendrycks2021measuring,cobbe2021training,kushman-etal-2014-learning} and metrics \cite{wang2021gpt,chung2022scaling,thoppilan2022lamda,yuan2023well} that encompass a wide spectrum of mathematical tasks, spanning from basic arithmetic to complex theorem proving, while accounting for linguistic fluency and mathematical accuracy, remains a significant challenge. 
Addressing these challenges is crucial to ascertain the reliability, efficacy, and generalizability of mathematical LMs, fostering advancements in this burgeoning field.

\paragraph{Creation.} 
While previous models exhibit remarkable capabilities in understanding and manipulating existing mathematical concepts, their ability to autonomously devise and rigorously prove entirely new theorems presents a formidable hurdle. 
The development of novel mathematical theorems demands not only profound mathematical reasoning but also creative and insightful problem-solving abilities, aspects that necessitate a deeper understanding of abstract mathematical concepts beyond what the models have been trained on. 
Davies et al. \cite{davies2021advancing} applied machine learning techniques to discover potential patterns and relationships among mathematical entities. 
Recently, Bernardino et al. \cite{bernardino2023mathematical} proposed FunSearch to find the first discoveries made for established open problems using LLMs. 
Bridging the gap between the models' proficiency in assimilating existing mathematical knowledge and their capability to generate novel and impactful theorems represents a significant frontier in leveraging LLMs for the advancement of mathematical discovery.

\paragraph{Application.} 
While mathematical LMs exhibit promising potential in autonomously solving math problems, their deployment in educational settings as teaching aids or tutors necessitates addressing several pivotal challenges. 
Tailoring these models to serve as effective educational tools demands not only mathematical proficiency but also adeptness in pedagogy and instructional methodologies. 
Few studies applied Socratic questioning for mathematical teaching \cite{zeng2022socratic,shridhar2022automatic}.
Customizing LLMs to cater to diverse learning styles, adapting explanations to suit different proficiency levels, and fostering an interactive and engaging learning environment is interesting. 

\paragraph{Data scarcity.} 
The training data significantly influences the performance of language models, particularly in LLMs \cite{ding_enhancing_2023}. High-quality and diverse training data can assist the model in enhancing its mathematical reasoning capabilities \cite{luo2023wizardmath}. As discussed in Section \ref{Instruction Building}, while there have been limited studies on constructing instruction data through LLMs, these efforts have only considered building from a small set of mathematical reasoning datasets, such as GSM8k \cite{cobbe2021training} and MATH \cite{hendrycks2021measuring}. High-quality and diverse mathematical instruction data remains scarce. It is necessary to explore additional forms and construction methods for mathematical train data. Additionally, the generation of mathematical training datasets in a multimodal context is also a promising direction.

%% file: main.bbl

\begin{thebibliography}{286}


\ifx \showCODEN    \undefined \def \showCODEN     #1{\unskip}     \fi
\ifx \showDOI      \undefined \def \showDOI       #1{#1}\fi
\ifx \showISBNx    \undefined \def \showISBNx     #1{\unskip}     \fi
\ifx \showISBNxiii \undefined \def \showISBNxiii  #1{\unskip}     \fi
\ifx \showISSN     \undefined \def \showISSN      #1{\unskip}     \fi
\ifx \showLCCN     \undefined \def \showLCCN      #1{\unskip}     \fi
\ifx \shownote     \undefined \def \shownote      #1{#1}          \fi
\ifx \showarticletitle \undefined \def \showarticletitle #1{#1}   \fi
\ifx \showURL      \undefined \def \showURL       {\relax}        \fi
\providecommand\bibfield[2]{#2}
\providecommand\bibinfo[2]{#2}
\providecommand\natexlab[1]{#1}
\providecommand\showeprint[2][]{arXiv:#2}

\bibitem[Al-Hossami et~al\mbox{.}(2023a)]%
        {alhossami2023language}
\bibfield{author}{\bibinfo{person}{Erfan Al-Hossami}, \bibinfo{person}{Razvan Bunescu}, \bibinfo{person}{Justin Smith}, {and} \bibinfo{person}{Ryan Teehan}.} \bibinfo{year}{2023}\natexlab{a}.
\newblock \bibinfo{title}{Can Language Models Employ the Socratic Method? Experiments with Code Debugging}.
\newblock
\newblock
\showeprint[arxiv]{2310.03210}~[cs.CL]


\bibitem[Al-Hossami et~al\mbox{.}(2023b)]%
        {al2023socratic}
\bibfield{author}{\bibinfo{person}{Erfan Al-Hossami}, \bibinfo{person}{Razvan Bunescu}, \bibinfo{person}{Ryan Teehan}, \bibinfo{person}{Laurel Powell}, \bibinfo{person}{Khyati Mahajan}, {and} \bibinfo{person}{Mohsen Dorodchi}.} \bibinfo{year}{2023}\natexlab{b}.
\newblock \showarticletitle{Socratic questioning of novice debuggers: A benchmark dataset and preliminary evaluations}. In \bibinfo{booktitle}{\emph{BEA}}. \bibinfo{pages}{709--726}.
\newblock


\bibitem[Alghamdi et~al\mbox{.}(2022)]%
        {alghamdi2022armath}
\bibfield{author}{\bibinfo{person}{Reem Alghamdi}, \bibinfo{person}{Zhenwen Liang}, {and} \bibinfo{person}{Xiangliang Zhang}.} \bibinfo{year}{2022}\natexlab{}.
\newblock \showarticletitle{ArMATH: a Dataset for Solving Arabic Math Word Problems}. In \bibinfo{booktitle}{\emph{LREC}}. \bibinfo{pages}{351--362}.
\newblock


\bibitem[Amini et~al\mbox{.}(2019)]%
        {amini2019mathqa}
\bibfield{author}{\bibinfo{person}{Aida Amini}, \bibinfo{person}{Saadia Gabriel}, \bibinfo{person}{Shanchuan Lin}, \bibinfo{person}{Rik Koncel-Kedziorski}, \bibinfo{person}{Yejin Choi}, {and} \bibinfo{person}{Hannaneh Hajishirzi}.} \bibinfo{year}{2019}\natexlab{}.
\newblock \showarticletitle{MathQA: Towards Interpretable Math Word Problem Solving with Operation-Based Formalisms}. In \bibinfo{booktitle}{\emph{NAACL}}. \bibinfo{pages}{2357--2367}.
\newblock


\bibitem[Ang et~al\mbox{.}(2023)]%
        {ang2023socratic}
\bibfield{author}{\bibinfo{person}{Beng~Heng Ang}, \bibinfo{person}{Sujatha~Das Gollapalli}, {and} \bibinfo{person}{See~Kiong Ng}.} \bibinfo{year}{2023}\natexlab{}.
\newblock \showarticletitle{Socratic Question Generation: A Novel Dataset, Models, and Evaluation}. In \bibinfo{booktitle}{\emph{EACL}}. \bibinfo{pages}{147--165}.
\newblock


\bibitem[Anonymous(2024)]%
        {anonymous2024diving}
\bibfield{author}{\bibinfo{person}{Anonymous}.} \bibinfo{year}{2024}\natexlab{}.
\newblock \showarticletitle{Diving into Self-Evolve Training for Multimodal Reasoning}. In \bibinfo{booktitle}{\emph{Submitted to The Thirteenth International Conference on Learning Representations}}.
\newblock


\bibitem[Antol et~al\mbox{.}(2015)]%
        {antol2015vqa}
\bibfield{author}{\bibinfo{person}{Stanislaw Antol}, \bibinfo{person}{Aishwarya Agrawal}, \bibinfo{person}{Jiasen Lu}, \bibinfo{person}{Margaret Mitchell}, \bibinfo{person}{Dhruv Batra}, \bibinfo{person}{C~Lawrence Zitnick}, {and} \bibinfo{person}{Devi Parikh}.} \bibinfo{year}{2015}\natexlab{}.
\newblock \showarticletitle{Vqa: Visual question answering}. In \bibinfo{booktitle}{\emph{ICCV}}. \bibinfo{pages}{2425--2433}.
\newblock


\bibitem[Austin et~al\mbox{.}(2021)]%
        {austin2021program}
\bibfield{author}{\bibinfo{person}{Jacob Austin}, \bibinfo{person}{Augustus Odena}, \bibinfo{person}{Maxwell Nye}, \bibinfo{person}{Maarten Bosma}, \bibinfo{person}{Henryk Michalewski}, \bibinfo{person}{David Dohan}, \bibinfo{person}{Ellen Jiang}, \bibinfo{person}{Carrie Cai}, \bibinfo{person}{Michael Terry}, \bibinfo{person}{Quoc Le}, {et~al\mbox{.}}} \bibinfo{year}{2021}\natexlab{}.
\newblock \showarticletitle{Program synthesis with large language models}.
\newblock \bibinfo{journal}{\emph{arXiv}} (\bibinfo{year}{2021}).
\newblock


\bibitem[Azerbayev et~al\mbox{.}(2023)]%
        {azerbayev2023proofnet}
\bibfield{author}{\bibinfo{person}{Zhangir Azerbayev}, \bibinfo{person}{Bartosz Piotrowski}, \bibinfo{person}{Hailey Schoelkopf}, \bibinfo{person}{Edward~W Ayers}, \bibinfo{person}{Dragomir Radev}, {and} \bibinfo{person}{Jeremy Avigad}.} \bibinfo{year}{2023}\natexlab{}.
\newblock \showarticletitle{Proofnet: Autoformalizing and formally proving undergraduate-level mathematics}.
\newblock \bibinfo{journal}{\emph{arXiv}} (\bibinfo{year}{2023}).
\newblock


\bibitem[Azerbayev et~al\mbox{.}({[n.\,d.]})]%
        {azerbayev8llemma}
\bibfield{author}{\bibinfo{person}{Zhangir Azerbayev}, \bibinfo{person}{Hailey Schoelkopf}, \bibinfo{person}{Keiran Paster}, \bibinfo{person}{Marco Dos~Santos}, \bibinfo{person}{Stephen McAleer}, \bibinfo{person}{Albert~Q Jiang}, \bibinfo{person}{Jia Deng}, \bibinfo{person}{Stella Biderman}, {and} \bibinfo{person}{Sean Welleck}.} \bibinfo{year}{[n.\,d.]}\natexlab{}.
\newblock \showarticletitle{LLEMMA: AN OPEN LANGUAGE MODEL FOR MATHEMATICS}.
\newblock \bibinfo{journal}{\emph{Minerva}}  \bibinfo{volume}{8} (\bibinfo{year}{[n.\,d.]}), \bibinfo{pages}{164B}.
\newblock


\bibitem[Bansal et~al\mbox{.}(2019)]%
        {bansal2019holist}
\bibfield{author}{\bibinfo{person}{Kshitij Bansal}, \bibinfo{person}{Sarah Loos}, \bibinfo{person}{Markus Rabe}, \bibinfo{person}{Christian Szegedy}, {and} \bibinfo{person}{Stewart Wilcox}.} \bibinfo{year}{2019}\natexlab{}.
\newblock \showarticletitle{Holist: An environment for machine learning of higher order logic theorem proving}. In \bibinfo{booktitle}{\emph{ICML}}. \bibinfo{pages}{454--463}.
\newblock


\bibitem[Berg-Kirkpatrick and Spokoyny(2020)]%
        {berg-kirkpatrickEmpiricalInvestigationContextualized2020}
\bibfield{author}{\bibinfo{person}{Taylor Berg-Kirkpatrick} {and} \bibinfo{person}{Daniel Spokoyny}.} \bibinfo{year}{2020}\natexlab{}.
\newblock \showarticletitle{An empirical investigation of contextualized number prediction}. In \bibinfo{booktitle}{\emph{EMNLP}}. \bibinfo{pages}{4754–4764}.
\newblock


\bibitem[Bernardino Romera-Paredes et~al\mbox{.}(2023)]%
        {bernardino2023mathematical}
\bibfield{author}{\bibinfo{person}{Alexander~Novikov Bernardino Romera-Paredes, Mohammadamin~Barekatain} {et~al\mbox{.}}} \bibinfo{year}{2023}\natexlab{}.
\newblock \showarticletitle{Mathematical discoveries from program search with large language models}.
\newblock \bibinfo{journal}{\emph{Nature}} (\bibinfo{year}{2023}).
\newblock


\bibitem[Bertot and Cast{\'e}ran(2013)]%
        {bertot2013interactive}
\bibfield{author}{\bibinfo{person}{Yves Bertot} {and} \bibinfo{person}{Pierre Cast{\'e}ran}.} \bibinfo{year}{2013}\natexlab{}.
\newblock \bibinfo{booktitle}{\emph{Interactive theorem proving and program development: Coq’Art: the calculus of inductive constructions}}.
\newblock


\bibitem[Besta et~al\mbox{.}(2023)]%
        {besta_graph_2023}
\bibfield{author}{\bibinfo{person}{Maciej Besta}, \bibinfo{person}{Nils Blach}, \bibinfo{person}{Ales Kubicek}, \bibinfo{person}{Robert Gerstenberger}, \bibinfo{person}{Lukas Gianinazzi}, \bibinfo{person}{Joanna Gajda}, \bibinfo{person}{Tomasz Lehmann}, \bibinfo{person}{Michal Podstawski}, \bibinfo{person}{Hubert Niewiadomski}, \bibinfo{person}{Piotr Nyczyk}, {and} \bibinfo{person}{Torsten Hoefler}.} \bibinfo{year}{2023}\natexlab{}.
\newblock \bibinfo{title}{Graph of {Thoughts}: {Solving} {Elaborate} {Problems} with {Large} {Language} {Models}}.
\newblock
\newblock


\bibitem[Bobrow et~al\mbox{.}(1964)]%
        {bobrow1964natural}
\bibfield{author}{\bibinfo{person}{Daniel Bobrow} {et~al\mbox{.}}} \bibinfo{year}{1964}\natexlab{}.
\newblock \showarticletitle{Natural language input for a computer problem solving system}.
\newblock  (\bibinfo{year}{1964}).
\newblock


\bibitem[Briars and Larkin(1984)]%
        {briars1984integrated}
\bibfield{author}{\bibinfo{person}{Diane~J Briars} {and} \bibinfo{person}{Jill~H Larkin}.} \bibinfo{year}{1984}\natexlab{}.
\newblock \showarticletitle{An integrated model of skill in solving elementary word problems}.
\newblock \bibinfo{journal}{\emph{Cognition and instruction}} \bibinfo{volume}{1}, \bibinfo{number}{3} (\bibinfo{year}{1984}), \bibinfo{pages}{245--296}.
\newblock


\bibitem[Brickhouse and Smith(2009)]%
        {brickhouse2009socratic}
\bibfield{author}{\bibinfo{person}{Thomas~C Brickhouse} {and} \bibinfo{person}{Nicholas~D Smith}.} \bibinfo{year}{2009}\natexlab{}.
\newblock \showarticletitle{Socratic teaching and Socratic method}.
\newblock  (\bibinfo{year}{2009}).
\newblock


\bibitem[Brown et~al\mbox{.}(2020)]%
        {brown2020language}
\bibfield{author}{\bibinfo{person}{Tom Brown}, \bibinfo{person}{Benjamin Mann}, \bibinfo{person}{Nick Ryder}, \bibinfo{person}{Melanie Subbiah}, \bibinfo{person}{Jared~D Kaplan}, \bibinfo{person}{Prafulla Dhariwal}, \bibinfo{person}{Arvind Neelakantan}, \bibinfo{person}{Pranav Shyam}, \bibinfo{person}{Girish Sastry}, \bibinfo{person}{Amanda Askell}, {et~al\mbox{.}}} \bibinfo{year}{2020}\natexlab{}.
\newblock \showarticletitle{Language models are few-shot learners}.
\newblock \bibinfo{journal}{\emph{NeurIPS}}  \bibinfo{volume}{33} (\bibinfo{year}{2020}), \bibinfo{pages}{1877--1901}.
\newblock


\bibitem[Chang(2023)]%
        {chang2023prompting}
\bibfield{author}{\bibinfo{person}{Edward~Y Chang}.} \bibinfo{year}{2023}\natexlab{}.
\newblock \showarticletitle{Prompting large language models with the socratic method}. In \bibinfo{booktitle}{\emph{CCWC}}. \bibinfo{pages}{0351--0360}.
\newblock


\bibitem[Chaslot et~al\mbox{.}(2008)]%
        {chaslot2008monte}
\bibfield{author}{\bibinfo{person}{Guillaume Chaslot}, \bibinfo{person}{Sander Bakkes}, \bibinfo{person}{Istvan Szita}, {and} \bibinfo{person}{Pieter Spronck}.} \bibinfo{year}{2008}\natexlab{}.
\newblock \showarticletitle{Monte-carlo tree search: A new framework for game ai}. In \bibinfo{booktitle}{\emph{AAAI}}, Vol.~\bibinfo{volume}{4}. \bibinfo{pages}{216--217}.
\newblock


\bibitem[Chen et~al\mbox{.}(2024a)]%
        {chen2024noise}
\bibfield{author}{\bibinfo{person}{Huayu Chen}, \bibinfo{person}{Guande He}, \bibinfo{person}{Lifan Yuan}, \bibinfo{person}{Ganqu Cui}, \bibinfo{person}{Hang Su}, {and} \bibinfo{person}{Jun Zhu}.} \bibinfo{year}{2024}\natexlab{a}.
\newblock \showarticletitle{Noise contrastive alignment of language models with explicit rewards}.
\newblock \bibinfo{journal}{\emph{arXiv}} (\bibinfo{year}{2024}).
\newblock


\bibitem[Chen et~al\mbox{.}(2021b)]%
        {chen2021geoqa}
\bibfield{author}{\bibinfo{person}{Jiaqi Chen}, \bibinfo{person}{Jianheng Tang}, \bibinfo{person}{Jinghui Qin}, \bibinfo{person}{Xiaodan Liang}, \bibinfo{person}{Lingbo Liu}, \bibinfo{person}{Eric~P Xing}, {and} \bibinfo{person}{Liang Lin}.} \bibinfo{year}{2021}\natexlab{b}.
\newblock \showarticletitle{GeoQA: A geometric question answering benchmark towards multimodal numerical reasoning}.
\newblock \bibinfo{journal}{\emph{arXiv}} (\bibinfo{year}{2021}).
\newblock


\bibitem[Chen et~al\mbox{.}(2024c)]%
        {chen2024controlmath}
\bibfield{author}{\bibinfo{person}{Nuo Chen}, \bibinfo{person}{Ning Wu}, \bibinfo{person}{Jianhui Chang}, {and} \bibinfo{person}{Jia Li}.} \bibinfo{year}{2024}\natexlab{c}.
\newblock \showarticletitle{ControlMath: Controllable Data Generation Promotes Math Generalist Models}.
\newblock \bibinfo{journal}{\emph{arXiv}} (\bibinfo{year}{2024}).
\newblock


\bibitem[Chen et~al\mbox{.}(2024b)]%
        {chen-etal-2024-m3cot}
\bibfield{author}{\bibinfo{person}{Qiguang Chen}, \bibinfo{person}{Libo Qin}, \bibinfo{person}{Jin Zhang}, \bibinfo{person}{Zhi Chen}, \bibinfo{person}{Xiao Xu}, {and} \bibinfo{person}{Wanxiang Che}.} \bibinfo{year}{2024}\natexlab{b}.
\newblock \showarticletitle{{M}$^3${C}o{T}: A Novel Benchmark for Multi-Domain Multi-step Multi-modal Chain-of-Thought}. In \bibinfo{booktitle}{\emph{ACL}}. \bibinfo{pages}{8199--8221}.
\newblock


\bibitem[Chen et~al\mbox{.}(2022a)]%
        {chen2022program}
\bibfield{author}{\bibinfo{person}{Wenhu Chen}, \bibinfo{person}{Xueguang Ma}, \bibinfo{person}{Xinyi Wang}, {and} \bibinfo{person}{William~W Cohen}.} \bibinfo{year}{2022}\natexlab{a}.
\newblock \showarticletitle{Program of thoughts prompting: Disentangling computation from reasoning for numerical reasoning tasks}.
\newblock \bibinfo{journal}{\emph{arXiv}} (\bibinfo{year}{2022}).
\newblock


\bibitem[Chen et~al\mbox{.}(2022b)]%
        {chen2022meta}
\bibfield{author}{\bibinfo{person}{Yanda Chen}, \bibinfo{person}{Ruiqi Zhong}, \bibinfo{person}{Sheng Zha}, \bibinfo{person}{George Karypis}, {and} \bibinfo{person}{He He}.} \bibinfo{year}{2022}\natexlab{b}.
\newblock \showarticletitle{Meta-learning via Language Model In-context Tuning}. In \bibinfo{booktitle}{\emph{ACL}}. \bibinfo{pages}{719--730}.
\newblock


\bibitem[Chen et~al\mbox{.}(2021a)]%
        {chen2021finqa}
\bibfield{author}{\bibinfo{person}{Zhiyu Chen}, \bibinfo{person}{Wenhu Chen}, \bibinfo{person}{Charese Smiley}, \bibinfo{person}{Sameena Shah}, \bibinfo{person}{Iana Borova}, \bibinfo{person}{Dylan Langdon}, \bibinfo{person}{Reema Moussa}, \bibinfo{person}{Matt Beane}, \bibinfo{person}{Ting-Hao Huang}, \bibinfo{person}{Bryan~R Routledge}, {et~al\mbox{.}}} \bibinfo{year}{2021}\natexlab{a}.
\newblock \showarticletitle{FinQA: A Dataset of Numerical Reasoning over Financial Data}. In \bibinfo{booktitle}{\emph{EMNLP}}. \bibinfo{pages}{3697--3711}.
\newblock


\bibitem[Chen et~al\mbox{.}(2024d)]%
        {chen2024internvl}
\bibfield{author}{\bibinfo{person}{Zhe Chen}, \bibinfo{person}{Jiannan Wu}, \bibinfo{person}{Wenhai Wang}, \bibinfo{person}{Weijie Su}, \bibinfo{person}{Guo Chen}, \bibinfo{person}{Sen Xing}, \bibinfo{person}{Muyan Zhong}, \bibinfo{person}{Qinglong Zhang}, \bibinfo{person}{Xizhou Zhu}, \bibinfo{person}{Lewei Lu}, {et~al\mbox{.}}} \bibinfo{year}{2024}\natexlab{d}.
\newblock \showarticletitle{Internvl: Scaling up vision foundation models and aligning for generic visual-linguistic tasks}. In \bibinfo{booktitle}{\emph{CVPR}}. \bibinfo{pages}{24185--24198}.
\newblock


\bibitem[Chernyshev et~al\mbox{.}(2024)]%
        {chernyshev2024umathuniversitylevelbenchmarkevaluating}
\bibfield{author}{\bibinfo{person}{Konstantin Chernyshev}, \bibinfo{person}{Vitaliy Polshkov}, \bibinfo{person}{Ekaterina Artemova}, \bibinfo{person}{Alex Myasnikov}, \bibinfo{person}{Vlad Stepanov}, \bibinfo{person}{Alexei Miasnikov}, {and} \bibinfo{person}{Sergei Tilga}.} \bibinfo{year}{2024}\natexlab{}.
\newblock \bibinfo{title}{U-MATH: A University-Level Benchmark for Evaluating Mathematical Skills in LLMs}.
\newblock
\newblock
\showeprint[arxiv]{2412.03205}


\bibitem[Chowdhery et~al\mbox{.}(2022)]%
        {chowdhery2022palm}
\bibfield{author}{\bibinfo{person}{Aakanksha Chowdhery}, \bibinfo{person}{Sharan Narang}, \bibinfo{person}{Jacob Devlin}, \bibinfo{person}{Maarten Bosma}, \bibinfo{person}{Gaurav Mishra}, \bibinfo{person}{Adam Roberts}, \bibinfo{person}{Paul Barham}, \bibinfo{person}{Hyung~Won Chung}, \bibinfo{person}{Charles Sutton}, \bibinfo{person}{Sebastian Gehrmann}, {et~al\mbox{.}}} \bibinfo{year}{2022}\natexlab{}.
\newblock \showarticletitle{Palm: Scaling language modeling with pathways}.
\newblock \bibinfo{journal}{\emph{arXiv}} (\bibinfo{year}{2022}).
\newblock


\bibitem[Chu et~al\mbox{.}(2023)]%
        {chuCoTReasoningSurvey2023}
\bibfield{author}{\bibinfo{person}{Zheng Chu}, \bibinfo{person}{Jingchang Chen}, \bibinfo{person}{Qianglong Chen}, \bibinfo{person}{Weijiang Yu}, \bibinfo{person}{Tao He}, \bibinfo{person}{Haotian Wang}, \bibinfo{person}{Weihua Peng}, \bibinfo{person}{Ming Liu}, \bibinfo{person}{Bing Qin}, {and} \bibinfo{person}{Ting Liu}.} \bibinfo{year}{2023}\natexlab{}.
\newblock \bibinfo{title}{A Survey of Chain of Thought Reasoning: Advances, Frontiers and Future}.
\newblock
\newblock
\showeprint[arxiv]{2309.15402}~[cs.CL]


\bibitem[Chung et~al\mbox{.}(2022)]%
        {chung2022scaling}
\bibfield{author}{\bibinfo{person}{Hyung~Won Chung}, \bibinfo{person}{Le Hou}, \bibinfo{person}{Shayne Longpre}, \bibinfo{person}{Barret Zoph}, \bibinfo{person}{Yi Tay}, \bibinfo{person}{William Fedus}, \bibinfo{person}{Yunxuan Li}, \bibinfo{person}{Xuezhi Wang}, \bibinfo{person}{Mostafa Dehghani}, \bibinfo{person}{Siddhartha Brahma}, {et~al\mbox{.}}} \bibinfo{year}{2022}\natexlab{}.
\newblock \showarticletitle{Scaling instruction-finetuned language models}.
\newblock \bibinfo{journal}{\emph{arXiv}} (\bibinfo{year}{2022}).
\newblock


\bibitem[Clark et~al\mbox{.}(2021)]%
        {clarkRegentsScienceExams2021}
\bibfield{author}{\bibinfo{person}{Peter Clark}, \bibinfo{person}{Oren Etzioni}, \bibinfo{person}{Daniel Khashabi}, \bibinfo{person}{Tushar Khot}, \bibinfo{person}{Bhavana~Dalvi Mishra}, \bibinfo{person}{Kyle Richardson}, \bibinfo{person}{Ashish Sabharwal}, \bibinfo{person}{Carissa Schoenick}, \bibinfo{person}{Oyvind Tafjord}, \bibinfo{person}{Niket Tandon}, \bibinfo{person}{Sumithra Bhakthavatsalam}, \bibinfo{person}{Dirk Groeneveld}, \bibinfo{person}{Michal Guerquin}, {and} \bibinfo{person}{Michael Schmitz}.} \bibinfo{year}{2021}\natexlab{}.
\newblock \bibinfo{title}{From '{{F}}' to 'a' on the {{N}}.{{Y}}. Regents Science Exams: {{An}} Overview of the Aristo Project}.
\newblock
\newblock
\showeprint{1909.01958}


\bibitem[Cobbe et~al\mbox{.}(2021)]%
        {cobbe2021training}
\bibfield{author}{\bibinfo{person}{Karl Cobbe}, \bibinfo{person}{Vineet Kosaraju}, \bibinfo{person}{Mohammad Bavarian}, \bibinfo{person}{Mark Chen}, \bibinfo{person}{Heewoo Jun}, \bibinfo{person}{Lukasz Kaiser}, \bibinfo{person}{Matthias Plappert}, \bibinfo{person}{Jerry Tworek}, \bibinfo{person}{Jacob Hilton}, \bibinfo{person}{Reiichiro Nakano}, \bibinfo{person}{Christopher Hesse}, {and} \bibinfo{person}{John Schulman}.} \bibinfo{year}{2021}\natexlab{}.
\newblock \bibinfo{title}{Training Verifiers to Solve Math Word Problems}.
\newblock
\newblock
\showeprint[arxiv]{2110.14168}


\bibitem[Couperus(2023)]%
        {couperus2023large}
\bibfield{author}{\bibinfo{person}{Jelle Couperus}.} \bibinfo{year}{2023}\natexlab{}.
\newblock \emph{\bibinfo{title}{Large Language Models and Mathematical Understanding}}.
\newblock \bibinfo{thesistype}{Master's\ thesis}.
\newblock


\bibitem[Dagan et~al\mbox{.}(2023)]%
        {dagan2023dynamic}
\bibfield{author}{\bibinfo{person}{Gautier Dagan}, \bibinfo{person}{Frank Keller}, {and} \bibinfo{person}{Alex Lascarides}.} \bibinfo{year}{2023}\natexlab{}.
\newblock \showarticletitle{Dynamic Planning with a LLM}.
\newblock \bibinfo{journal}{\emph{arXiv}} (\bibinfo{year}{2023}).
\newblock


\bibitem[Davies et~al\mbox{.}(2021)]%
        {davies2021advancing}
\bibfield{author}{\bibinfo{person}{Alex Davies}, \bibinfo{person}{Petar Veli{\v{c}}kovi{\'c}}, \bibinfo{person}{Lars Buesing}, \bibinfo{person}{Sam Blackwell}, \bibinfo{person}{Daniel Zheng}, \bibinfo{person}{Nenad Toma{\v{s}}ev}, \bibinfo{person}{Richard Tanburn}, \bibinfo{person}{Peter Battaglia}, \bibinfo{person}{Charles Blundell}, \bibinfo{person}{Andr{\'a}s Juh{\'a}sz}, {et~al\mbox{.}}} \bibinfo{year}{2021}\natexlab{}.
\newblock \showarticletitle{Advancing mathematics by guiding human intuition with AI}.
\newblock \bibinfo{journal}{\emph{Nature}} \bibinfo{volume}{600}, \bibinfo{number}{7887} (\bibinfo{year}{2021}), \bibinfo{pages}{70--74}.
\newblock


\bibitem[Davis and Steinglass(1997)]%
        {davis1997dialogue}
\bibfield{author}{\bibinfo{person}{Peggy~Cooper Davis} {and} \bibinfo{person}{Elizabeth~Ehrenfest Steinglass}.} \bibinfo{year}{1997}\natexlab{}.
\newblock \showarticletitle{A dialogue about Socratic teaching}.
\newblock \bibinfo{journal}{\emph{NYU Rev. L. \& Soc. Change}}  \bibinfo{volume}{23} (\bibinfo{year}{1997}), \bibinfo{pages}{249}.
\newblock


\bibitem[Deng and Mineiro(2024)]%
        {deng2024flow}
\bibfield{author}{\bibinfo{person}{Yihe Deng} {and} \bibinfo{person}{Paul Mineiro}.} \bibinfo{year}{2024}\natexlab{}.
\newblock \showarticletitle{Flow-DPO: Improving LLM Mathematical Reasoning through Online Multi-Agent Learning}.
\newblock \bibinfo{journal}{\emph{arXiv}} (\bibinfo{year}{2024}).
\newblock


\bibitem[Devlin et~al\mbox{.}(2019)]%
        {devlinBERTPretrainingDeep2019}
\bibfield{author}{\bibinfo{person}{Jacob Devlin}, \bibinfo{person}{Ming-Wei Chang}, \bibinfo{person}{Kenton Lee}, {and} \bibinfo{person}{Kristina Toutanova}.} \bibinfo{year}{2019}\natexlab{}.
\newblock \bibinfo{title}{{{BERT}}: {{Pre-training}} of {{Deep Bidirectional Transformers}} for {{Language Understanding}}}.
\newblock
\newblock
\showeprint{1810.04805}


\bibitem[Ding et~al\mbox{.}(2023)]%
        {ding_enhancing_2023}
\bibfield{author}{\bibinfo{person}{Ning Ding}, \bibinfo{person}{Yulin Chen}, \bibinfo{person}{Bokai Xu}, \bibinfo{person}{Yujia Qin}, \bibinfo{person}{Shengding Hu}, \bibinfo{person}{Zhiyuan Liu}, \bibinfo{person}{Maosong Sun}, {and} \bibinfo{person}{Bowen Zhou}.} \bibinfo{year}{2023}\natexlab{}.
\newblock \showarticletitle{Enhancing Chat Language Models by Scaling High-quality Instructional Conversations}. In \bibinfo{booktitle}{\emph{EMNLP}}. \bibinfo{pages}{3029--3051}.
\newblock


\bibitem[Ding et~al\mbox{.}(2024)]%
        {ding2024boosting}
\bibfield{author}{\bibinfo{person}{Yuyang Ding}, \bibinfo{person}{Hanglei Hu}, \bibinfo{person}{Jie Zhou}, \bibinfo{person}{Qin Chen}, \bibinfo{person}{Bo Jiang}, {and} \bibinfo{person}{Liang He}.} \bibinfo{year}{2024}\natexlab{}.
\newblock \showarticletitle{Boosting Large Language Models with Socratic Method for Conversational Mathematics Teaching}. In \bibinfo{booktitle}{\emph{CIKM}}. \bibinfo{pages}{3730--3735}.
\newblock


\bibitem[Drori et~al\mbox{.}(2022)]%
        {drori2022neural}
\bibfield{author}{\bibinfo{person}{Iddo Drori}, \bibinfo{person}{Sarah Zhang}, \bibinfo{person}{Reece Shuttleworth}, \bibinfo{person}{Leonard Tang}, \bibinfo{person}{Albert Lu}, \bibinfo{person}{Elizabeth Ke}, \bibinfo{person}{Kevin Liu}, \bibinfo{person}{Linda Chen}, \bibinfo{person}{Sunny Tran}, \bibinfo{person}{Newman Cheng}, {et~al\mbox{.}}} \bibinfo{year}{2022}\natexlab{}.
\newblock \showarticletitle{A neural network solves, explains, and generates university math problems by program synthesis and few-shot learning at human level}.
\newblock \bibinfo{journal}{\emph{PNAS}} \bibinfo{volume}{119}, \bibinfo{number}{32} (\bibinfo{year}{2022}), \bibinfo{pages}{e2123433119}.
\newblock


\bibitem[Dua et~al\mbox{.}(2019)]%
        {dua2019drop}
\bibfield{author}{\bibinfo{person}{Dheeru Dua}, \bibinfo{person}{Yizhong Wang}, \bibinfo{person}{Pradeep Dasigi}, \bibinfo{person}{Gabriel Stanovsky}, \bibinfo{person}{Sameer Singh}, {and} \bibinfo{person}{Matt Gardner}.} \bibinfo{year}{2019}\natexlab{}.
\newblock \showarticletitle{DROP: A Reading Comprehension Benchmark Requiring Discrete Reasoning Over Paragraphs}. In \bibinfo{booktitle}{\emph{NAACL}}. \bibinfo{pages}{2368--2378}.
\newblock


\bibitem[Duan et~al\mbox{.}(2023)]%
        {duan2023shifting}
\bibfield{author}{\bibinfo{person}{Jinhao Duan}, \bibinfo{person}{Hao Cheng}, \bibinfo{person}{Shiqi Wang}, \bibinfo{person}{Chenan Wang}, \bibinfo{person}{Alex Zavalny}, \bibinfo{person}{Renjing Xu}, \bibinfo{person}{Bhavya Kailkhura}, {and} \bibinfo{person}{Kaidi Xu}.} \bibinfo{year}{2023}\natexlab{}.
\newblock \showarticletitle{Shifting attention to relevance: Towards the uncertainty estimation of large language models}.
\newblock \bibinfo{journal}{\emph{arXiv}} (\bibinfo{year}{2023}).
\newblock


\bibitem[Elazar et~al\mbox{.}(2019)]%
        {elazar2019large}
\bibfield{author}{\bibinfo{person}{Yanai Elazar}, \bibinfo{person}{Abhijit Mahabal}, \bibinfo{person}{Deepak Ramachandran}, \bibinfo{person}{Tania Bedrax-Weiss}, {and} \bibinfo{person}{Dan Roth}.} \bibinfo{year}{2019}\natexlab{}.
\newblock \showarticletitle{How Large Are Lions? Inducing Distributions over Quantitative Attributes}. In \bibinfo{booktitle}{\emph{ACL}}. \bibinfo{pages}{3973--3983}.
\newblock


\bibitem[Ethayarajh et~al\mbox{.}(2024)]%
        {ethayarajh2024kto}
\bibfield{author}{\bibinfo{person}{Kawin Ethayarajh}, \bibinfo{person}{Winnie Xu}, \bibinfo{person}{Niklas Muennighoff}, \bibinfo{person}{Dan Jurafsky}, {and} \bibinfo{person}{Douwe Kiela}.} \bibinfo{year}{2024}\natexlab{}.
\newblock \showarticletitle{Kto: Model alignment as prospect theoretic optimization}.
\newblock \bibinfo{journal}{\emph{arXiv}} (\bibinfo{year}{2024}).
\newblock


\bibitem[Feigenbaum et~al\mbox{.}(1963)]%
        {feigenbaum1963computers}
\bibfield{author}{\bibinfo{person}{Edward~A Feigenbaum}, \bibinfo{person}{Julian Feldman}, {et~al\mbox{.}}} \bibinfo{year}{1963}\natexlab{}.
\newblock \bibinfo{booktitle}{\emph{Computers and thought}}. Vol.~\bibinfo{volume}{7}.
\newblock


\bibitem[Feng et~al\mbox{.}(2022)]%
        {fengInjectingNumericalReasoning2022}
\bibfield{author}{\bibinfo{person}{Yu Feng}, \bibinfo{person}{Jing Zhang}, \bibinfo{person}{Xiaokang Zhang}, \bibinfo{person}{Lemao Liu}, \bibinfo{person}{Cuiping Li}, {and} \bibinfo{person}{Hong Chen}.} \bibinfo{year}{2022}\natexlab{}.
\newblock \bibinfo{title}{Injecting Numerical Reasoning Skills into Knowledge Base Question Answering Models}.
\newblock
\newblock
\showeprint{2112.06109}


\bibitem[First et~al\mbox{.}(2023)]%
        {first2023baldur}
\bibfield{author}{\bibinfo{person}{Emily First}, \bibinfo{person}{Markus~N Rabe}, \bibinfo{person}{Talia Ringer}, {and} \bibinfo{person}{Yuriy Brun}.} \bibinfo{year}{2023}\natexlab{}.
\newblock \showarticletitle{Baldur: whole-proof generation and repair with large language models}.
\newblock \bibinfo{journal}{\emph{arXiv}} (\bibinfo{year}{2023}).
\newblock


\bibitem[Fletcher(1985)]%
        {fletcher1985understanding}
\bibfield{author}{\bibinfo{person}{Charles~R Fletcher}.} \bibinfo{year}{1985}\natexlab{}.
\newblock \showarticletitle{Understanding and solving arithmetic word problems: A computer simulation}.
\newblock \bibinfo{journal}{\emph{Behavior Research Methods, Instruments, \& Computers}} \bibinfo{volume}{17}, \bibinfo{number}{5} (\bibinfo{year}{1985}), \bibinfo{pages}{565--571}.
\newblock


\bibitem[Forbes and Choi(2017)]%
        {forbes2017verb}
\bibfield{author}{\bibinfo{person}{Maxwell Forbes} {and} \bibinfo{person}{Yejin Choi}.} \bibinfo{year}{2017}\natexlab{}.
\newblock \showarticletitle{Verb Physics: Relative Physical Knowledge of Actions and Objects}. In \bibinfo{booktitle}{\emph{ACL}}. \bibinfo{pages}{266--276}.
\newblock


\bibitem[Fu et~al\mbox{.}(2023)]%
        {fu2022complexity}
\bibfield{author}{\bibinfo{person}{Yao Fu}, \bibinfo{person}{Hao Peng}, \bibinfo{person}{Ashish Sabharwal}, \bibinfo{person}{Peter Clark}, {and} \bibinfo{person}{Tushar Khot}.} \bibinfo{year}{2023}\natexlab{}.
\newblock \showarticletitle{Complexity-Based Prompting for Multi-step Reasoning}. In \bibinfo{booktitle}{\emph{ICLR}}.
\newblock


\bibitem[Gao et~al\mbox{.}(2023)]%
        {gao2023pal}
\bibfield{author}{\bibinfo{person}{Luyu Gao}, \bibinfo{person}{Aman Madaan}, \bibinfo{person}{Shuyan Zhou}, \bibinfo{person}{Uri Alon}, \bibinfo{person}{Pengfei Liu}, \bibinfo{person}{Yiming Yang}, \bibinfo{person}{Jamie Callan}, {and} \bibinfo{person}{Graham Neubig}.} \bibinfo{year}{2023}\natexlab{}.
\newblock \showarticletitle{Pal: Program-aided language models}. In \bibinfo{booktitle}{\emph{ICML}}. \bibinfo{pages}{10764--10799}.
\newblock


\bibitem[Gawlikowski et~al\mbox{.}(2023)]%
        {gawlikowski2023survey}
\bibfield{author}{\bibinfo{person}{Jakob Gawlikowski}, \bibinfo{person}{Cedrique Rovile~Njieutcheu Tassi}, \bibinfo{person}{Mohsin Ali}, \bibinfo{person}{Jongseok Lee}, \bibinfo{person}{Matthias Humt}, \bibinfo{person}{Jianxiang Feng}, \bibinfo{person}{Anna Kruspe}, \bibinfo{person}{Rudolph Triebel}, \bibinfo{person}{Peter Jung}, \bibinfo{person}{Ribana Roscher}, {et~al\mbox{.}}} \bibinfo{year}{2023}\natexlab{}.
\newblock \showarticletitle{A survey of uncertainty in deep neural networks}.
\newblock \bibinfo{journal}{\emph{Artificial Intelligence Review}} \bibinfo{volume}{56}, \bibinfo{number}{Suppl 1} (\bibinfo{year}{2023}), \bibinfo{pages}{1513--1589}.
\newblock


\bibitem[Gelernter et~al\mbox{.}(1960)]%
        {gelernter1960empirical}
\bibfield{author}{\bibinfo{person}{Herbert Gelernter}, \bibinfo{person}{James~R Hansen}, {and} \bibinfo{person}{Donald~W Loveland}.} \bibinfo{year}{1960}\natexlab{}.
\newblock \showarticletitle{Empirical explorations of the geometry theorem machine}. In \bibinfo{booktitle}{\emph{western joint IRE-AIEE-ACM computer conference}}. \bibinfo{pages}{143--149}.
\newblock


\bibitem[Geva et~al\mbox{.}(2020)]%
        {gevaInjectingNumericalReasoning2020}
\bibfield{author}{\bibinfo{person}{Mor Geva}, \bibinfo{person}{Ankit Gupta}, {and} \bibinfo{person}{Jonathan Berant}.} \bibinfo{year}{2020}\natexlab{}.
\newblock \showarticletitle{Injecting Numerical Reasoning Skills into Language Models}. In \bibinfo{booktitle}{\emph{ACL}}, \bibfield{editor}{\bibinfo{person}{Dan Jurafsky}, \bibinfo{person}{Joyce Chai}, \bibinfo{person}{Natalie Schluter}, {and} \bibinfo{person}{Joel Tetreault}} (Eds.). \bibinfo{pages}{946--958}.
\newblock


\bibitem[GLM et~al\mbox{.}(2024)]%
        {glm2024chatglm}
\bibfield{author}{\bibinfo{person}{Team GLM}, \bibinfo{person}{Aohan Zeng}, \bibinfo{person}{Bin Xu}, \bibinfo{person}{Bowen Wang}, \bibinfo{person}{Chenhui Zhang}, \bibinfo{person}{Da Yin}, \bibinfo{person}{Dan Zhang}, \bibinfo{person}{Diego Rojas}, \bibinfo{person}{Guanyu Feng}, \bibinfo{person}{Hanlin Zhao}, {et~al\mbox{.}}} \bibinfo{year}{2024}\natexlab{}.
\newblock \showarticletitle{Chatglm: A family of large language models from glm-130b to glm-4 all tools}.
\newblock \bibinfo{journal}{\emph{arXiv}} (\bibinfo{year}{2024}).
\newblock


\bibitem[Gonthier et~al\mbox{.}(2013)]%
        {gonthier2013machine}
\bibfield{author}{\bibinfo{person}{Georges Gonthier}, \bibinfo{person}{Andrea Asperti}, \bibinfo{person}{Jeremy Avigad}, \bibinfo{person}{Yves Bertot}, \bibinfo{person}{Cyril Cohen}, \bibinfo{person}{Fran{\c{c}}ois Garillot}, \bibinfo{person}{St{\'e}phane Le~Roux}, \bibinfo{person}{Assia Mahboubi}, \bibinfo{person}{Russell O’Connor}, \bibinfo{person}{Sidi Ould~Biha}, {et~al\mbox{.}}} \bibinfo{year}{2013}\natexlab{}.
\newblock \showarticletitle{A machine-checked proof of the odd order theorem}. In \bibinfo{booktitle}{\emph{ITP}}. \bibinfo{pages}{163--179}.
\newblock


\bibitem[Gou et~al\mbox{.}(2023a)]%
        {gou2023critic}
\bibfield{author}{\bibinfo{person}{Zhibin Gou}, \bibinfo{person}{Zhihong Shao}, \bibinfo{person}{Yeyun Gong}, \bibinfo{person}{Yelong Shen}, \bibinfo{person}{Yujiu Yang}, \bibinfo{person}{Nan Duan}, {and} \bibinfo{person}{Weizhu Chen}.} \bibinfo{year}{2023}\natexlab{a}.
\newblock \bibinfo{title}{CRITIC: Large Language Models Can Self-Correct with Tool-Interactive Critiquing}.
\newblock
\newblock
\showeprint[arxiv]{2305.11738}


\bibitem[Gou et~al\mbox{.}(2023b)]%
        {gou2023tora}
\bibfield{author}{\bibinfo{person}{Zhibin Gou}, \bibinfo{person}{Zhihong Shao}, \bibinfo{person}{Yeyun Gong}, \bibinfo{person}{Yujiu Yang}, \bibinfo{person}{Minlie Huang}, \bibinfo{person}{Nan Duan}, \bibinfo{person}{Weizhu Chen}, {et~al\mbox{.}}} \bibinfo{year}{2023}\natexlab{b}.
\newblock \showarticletitle{Tora: A tool-integrated reasoning agent for mathematical problem solving}.
\newblock \bibinfo{journal}{\emph{arXiv}} (\bibinfo{year}{2023}).
\newblock


\bibitem[Grabowski et~al\mbox{.}(2015)]%
        {grabowski2015four}
\bibfield{author}{\bibinfo{person}{Adam Grabowski}, \bibinfo{person}{Artur Korni{\l}owicz}, {and} \bibinfo{person}{Adam Naumowicz}.} \bibinfo{year}{2015}\natexlab{}.
\newblock \showarticletitle{Four Decades of Mizar: Foreword}.
\newblock \bibinfo{journal}{\emph{Journal of Automated Reasoning}}  \bibinfo{volume}{55} (\bibinfo{year}{2015}), \bibinfo{pages}{191--198}.
\newblock


\bibitem[Hales et~al\mbox{.}(2017)]%
        {hales2017formal}
\bibfield{author}{\bibinfo{person}{Thomas Hales}, \bibinfo{person}{Mark Adams}, \bibinfo{person}{Gertrud Bauer}, \bibinfo{person}{Tat~Dat Dang}, \bibinfo{person}{John Harrison}, \bibinfo{person}{Hoang Le~Truong}, \bibinfo{person}{Cezary Kaliszyk}, \bibinfo{person}{Victor Magron}, \bibinfo{person}{Sean McLaughlin}, \bibinfo{person}{Tat~Thang Nguyen}, {et~al\mbox{.}}} \bibinfo{year}{2017}\natexlab{}.
\newblock \showarticletitle{A formal proof of the Kepler conjecture}. In \bibinfo{booktitle}{\emph{Forum of mathematics, Pi}}, Vol.~\bibinfo{volume}{5}. \bibinfo{pages}{e2}.
\newblock


\bibitem[Han et~al\mbox{.}(2022)]%
        {han2021proof}
\bibfield{author}{\bibinfo{person}{Jesse~Michael Han}, \bibinfo{person}{Jason Rute}, \bibinfo{person}{Yuhuai Wu}, \bibinfo{person}{Edward Ayers}, {and} \bibinfo{person}{Stanislas Polu}.} \bibinfo{year}{2022}\natexlab{}.
\newblock \showarticletitle{Proof Artifact Co-Training for Theorem Proving with Language Models}. In \bibinfo{booktitle}{\emph{ICLR}}.
\newblock


\bibitem[Han et~al\mbox{.}(2024)]%
        {DBLP:conf/emnlp/HanS0QRZCPQBSWS24}
\bibfield{author}{\bibinfo{person}{Simeng Han}, \bibinfo{person}{Hailey Schoelkopf}, \bibinfo{person}{Yilun Zhao}, \bibinfo{person}{Zhenting Qi}, \bibinfo{person}{Martin Riddell}, \bibinfo{person}{Wenfei Zhou}, \bibinfo{person}{James Coady}, \bibinfo{person}{David Peng}, \bibinfo{person}{Yujie Qiao}, \bibinfo{person}{Luke Benson}, \bibinfo{person}{Lucy Sun}, \bibinfo{person}{Alexander Wardle{-}Solano}, \bibinfo{person}{Hannah Szab{\'{o}}}, \bibinfo{person}{Ekaterina Zubova}, \bibinfo{person}{Matthew Burtell}, \bibinfo{person}{Jonathan Fan}, \bibinfo{person}{Yixin Liu}, \bibinfo{person}{Brian Wong}, \bibinfo{person}{Malcolm Sailor}, \bibinfo{person}{Ansong Ni}, \bibinfo{person}{Linyong Nan}, \bibinfo{person}{Jungo Kasai}, \bibinfo{person}{Tao Yu}, \bibinfo{person}{Rui Zhang}, \bibinfo{person}{Alexander~R. Fabbri}, \bibinfo{person}{Wojciech Kryscinski}, \bibinfo{person}{Semih Yavuz}, \bibinfo{person}{Ye Liu}, \bibinfo{person}{Xi~Victoria Lin}, \bibinfo{person}{Shafiq Joty}, \bibinfo{person}{Yingbo Zhou},
  \bibinfo{person}{Caiming Xiong}, \bibinfo{person}{Rex Ying}, \bibinfo{person}{Arman Cohan}, {and} \bibinfo{person}{Dragomir Radev}.} \bibinfo{year}{2024}\natexlab{}.
\newblock \showarticletitle{{FOLIO:} Natural Language Reasoning with First-Order Logic}. In \bibinfo{booktitle}{\emph{Proceedings of the 2024 Conference on Empirical Methods in Natural Language Processing, {EMNLP} 2024, Miami, FL, USA, November 12-16, 2024}}, \bibfield{editor}{\bibinfo{person}{Yaser Al{-}Onaizan}, \bibinfo{person}{Mohit Bansal}, {and} \bibinfo{person}{Yun{-}Nung Chen}} (Eds.). \bibinfo{publisher}{Association for Computational Linguistics}, \bibinfo{pages}{22017--22031}.
\newblock
\urldef\tempurl%
\url{https://aclanthology.org/2024.emnlp-main.1229}
\showURL{%
\tempurl}


\bibitem[Hao et~al\mbox{.}(2023a)]%
        {hao2023reasoning}
\bibfield{author}{\bibinfo{person}{Shibo Hao}, \bibinfo{person}{Yi Gu}, \bibinfo{person}{Haodi Ma}, \bibinfo{person}{Joshua~Jiahua Hong}, \bibinfo{person}{Zhen Wang}, \bibinfo{person}{Daisy~Zhe Wang}, {and} \bibinfo{person}{Zhiting Hu}.} \bibinfo{year}{2023}\natexlab{a}.
\newblock \showarticletitle{Reasoning with language model is planning with world model}.
\newblock \bibinfo{journal}{\emph{arXiv}} (\bibinfo{year}{2023}).
\newblock


\bibitem[Hao et~al\mbox{.}(2023b)]%
        {hao2023toolkengpt}
\bibfield{author}{\bibinfo{person}{Shibo Hao}, \bibinfo{person}{Tianyang Liu}, \bibinfo{person}{Zhen Wang}, {and} \bibinfo{person}{Zhiting Hu}.} \bibinfo{year}{2023}\natexlab{b}.
\newblock \bibinfo{title}{ToolkenGPT: Augmenting Frozen Language Models with Massive Tools via Tool Embeddings}.
\newblock
\newblock
\showeprint[arxiv]{2305.11554}


\bibitem[Hao et~al\mbox{.}(2022)]%
        {Hao2022PGDP5KAD}
\bibfield{author}{\bibinfo{person}{Yihan Hao}, \bibinfo{person}{Mingliang Zhang}, \bibinfo{person}{Fei Yin}, {and} \bibinfo{person}{Linlin Huang}.} \bibinfo{year}{2022}\natexlab{}.
\newblock \showarticletitle{PGDP5K: A Diagram Parsing Dataset for Plane Geometry Problems}.
\newblock \bibinfo{journal}{\emph{ICPR}} (\bibinfo{year}{2022}), \bibinfo{pages}{1763--1769}.
\newblock


\bibitem[He et~al\mbox{.}(2022)]%
        {he_rethinking_2022}
\bibfield{author}{\bibinfo{person}{Hangfeng He}, \bibinfo{person}{Hongming Zhang}, {and} \bibinfo{person}{Dan Roth}.} \bibinfo{year}{2022}\natexlab{}.
\newblock \showarticletitle{Rethinking with retrieval: Faithful large language model inference}.
\newblock \bibinfo{journal}{\emph{arXiv}} (\bibinfo{year}{2022}).
\newblock


\bibitem[He-Yueya et~al\mbox{.}(2023)]%
        {he2023solving}
\bibfield{author}{\bibinfo{person}{Joy He-Yueya}, \bibinfo{person}{Gabriel Poesia}, \bibinfo{person}{Rose~E Wang}, {and} \bibinfo{person}{Noah~D Goodman}.} \bibinfo{year}{2023}\natexlab{}.
\newblock \showarticletitle{Solving math word problems by combining language models with symbolic solvers}.
\newblock \bibinfo{journal}{\emph{arXiv}} (\bibinfo{year}{2023}).
\newblock


\bibitem[Hendrycks et~al\mbox{.}(2021a)]%
        {hendryckstest2021}
\bibfield{author}{\bibinfo{person}{Dan Hendrycks}, \bibinfo{person}{Collin Burns}, \bibinfo{person}{Steven Basart}, \bibinfo{person}{Andy Zou}, \bibinfo{person}{Mantas Mazeika}, \bibinfo{person}{Dawn Song}, {and} \bibinfo{person}{Jacob Steinhardt}.} \bibinfo{year}{2021}\natexlab{a}.
\newblock \showarticletitle{Measuring Massive Multitask Language Understanding}.
\newblock \bibinfo{journal}{\emph{ICLR}} (\bibinfo{year}{2021}).
\newblock


\bibitem[Hendrycks et~al\mbox{.}(2021b)]%
        {hendrycks2021measuring}
\bibfield{author}{\bibinfo{person}{Dan Hendrycks}, \bibinfo{person}{Collin Burns}, \bibinfo{person}{Saurav Kadavath}, \bibinfo{person}{Akul Arora}, \bibinfo{person}{Steven Basart}, \bibinfo{person}{Eric Tang}, \bibinfo{person}{Dawn Song}, {and} \bibinfo{person}{Jacob Steinhardt}.} \bibinfo{year}{2021}\natexlab{b}.
\newblock \showarticletitle{Measuring Mathematical Problem Solving With the MATH Dataset}. In \bibinfo{booktitle}{\emph{NeurIPS Datasets and Benchmarks Track (Round 2)}}.
\newblock


\bibitem[Hossain et~al\mbox{.}(2019)]%
        {hossain2019comprehensive}
\bibfield{author}{\bibinfo{person}{MD~Zakir Hossain}, \bibinfo{person}{Ferdous Sohel}, \bibinfo{person}{Mohd~Fairuz Shiratuddin}, {and} \bibinfo{person}{Hamid Laga}.} \bibinfo{year}{2019}\natexlab{}.
\newblock \showarticletitle{A comprehensive survey of deep learning for image captioning}.
\newblock \bibinfo{journal}{\emph{ACM Computing Surveys (CsUR)}} \bibinfo{volume}{51}, \bibinfo{number}{6} (\bibinfo{year}{2019}), \bibinfo{pages}{1--36}.
\newblock


\bibitem[Hosseini et~al\mbox{.}(2024)]%
        {hosseini2024v}
\bibfield{author}{\bibinfo{person}{Arian Hosseini}, \bibinfo{person}{Xingdi Yuan}, \bibinfo{person}{Nikolay Malkin}, \bibinfo{person}{Aaron Courville}, \bibinfo{person}{Alessandro Sordoni}, {and} \bibinfo{person}{Rishabh Agarwal}.} \bibinfo{year}{2024}\natexlab{}.
\newblock \showarticletitle{V-star: Training verifiers for self-taught reasoners}.
\newblock \bibinfo{journal}{\emph{arXiv}} (\bibinfo{year}{2024}).
\newblock


\bibitem[Hosseini et~al\mbox{.}(2014)]%
        {hosseini-etal-2014-learning}
\bibfield{author}{\bibinfo{person}{Mohammad~Javad Hosseini}, \bibinfo{person}{Hannaneh Hajishirzi}, \bibinfo{person}{Oren Etzioni}, {and} \bibinfo{person}{Nate Kushman}.} \bibinfo{year}{2014}\natexlab{}.
\newblock \showarticletitle{Learning to Solve Arithmetic Word Problems with Verb Categorization}. In \bibinfo{booktitle}{\emph{EMNLP}}. \bibinfo{pages}{523--533}.
\newblock


\bibitem[Hsieh et~al\mbox{.}(2023)]%
        {hsieh2023tool}
\bibfield{author}{\bibinfo{person}{Cheng-Yu Hsieh}, \bibinfo{person}{Si-An Chen}, \bibinfo{person}{Chun-Liang Li}, \bibinfo{person}{Yasuhisa Fujii}, \bibinfo{person}{Alexander Ratner}, \bibinfo{person}{Chen-Yu Lee}, \bibinfo{person}{Ranjay Krishna}, {and} \bibinfo{person}{Tomas Pfister}.} \bibinfo{year}{2023}\natexlab{}.
\newblock \bibinfo{title}{Tool Documentation Enables Zero-Shot Tool-Usage with Large Language Models}.
\newblock
\newblock
\showeprint[arxiv]{2308.00675}


\bibitem[Huang et~al\mbox{.}(2019)]%
        {huang2018gamepad}
\bibfield{author}{\bibinfo{person}{Daniel Huang}, \bibinfo{person}{Prafulla Dhariwal}, \bibinfo{person}{Dawn Song}, {and} \bibinfo{person}{Ilya Sutskever}.} \bibinfo{year}{2019}\natexlab{}.
\newblock \showarticletitle{GamePad: A Learning Environment for Theorem Proving}. In \bibinfo{booktitle}{\emph{ICLR}}.
\newblock


\bibitem[Huang et~al\mbox{.}(2016)]%
        {huang2016well}
\bibfield{author}{\bibinfo{person}{Danqing Huang}, \bibinfo{person}{Shuming Shi}, \bibinfo{person}{Chin-Yew Lin}, \bibinfo{person}{Jian Yin}, {and} \bibinfo{person}{Wei-Ying Ma}.} \bibinfo{year}{2016}\natexlab{}.
\newblock \showarticletitle{How well do computers solve math word problems? large-scale dataset construction and evaluation}. In \bibinfo{booktitle}{\emph{ACL}}. \bibinfo{pages}{887--896}.
\newblock


\bibitem[Huang et~al\mbox{.}(2024)]%
        {huang2024fewer}
\bibfield{author}{\bibinfo{person}{Xijie Huang}, \bibinfo{person}{Li~Lyna Zhang}, \bibinfo{person}{Kwang-Ting Cheng}, \bibinfo{person}{Fan Yang}, {and} \bibinfo{person}{Mao Yang}.} \bibinfo{year}{2024}\natexlab{}.
\newblock \showarticletitle{Fewer is More: Boosting Math Reasoning with Reinforced Context Pruning}. In \bibinfo{booktitle}{\emph{EMNLP}}. \bibinfo{pages}{13674--13695}.
\newblock


\bibitem[Hurst et~al\mbox{.}(2024)]%
        {hurst2024gpt}
\bibfield{author}{\bibinfo{person}{Aaron Hurst}, \bibinfo{person}{Adam Lerer}, \bibinfo{person}{Adam~P Goucher}, \bibinfo{person}{Adam Perelman}, \bibinfo{person}{Aditya Ramesh}, \bibinfo{person}{Aidan Clark}, \bibinfo{person}{AJ Ostrow}, \bibinfo{person}{Akila Welihinda}, \bibinfo{person}{Alan Hayes}, \bibinfo{person}{Alec Radford}, {et~al\mbox{.}}} \bibinfo{year}{2024}\natexlab{}.
\newblock \showarticletitle{Gpt-4o system card}.
\newblock \bibinfo{journal}{\emph{arXiv}} (\bibinfo{year}{2024}).
\newblock


\bibitem[Imani et~al\mbox{.}(2023)]%
        {imani2023mathprompter}
\bibfield{author}{\bibinfo{person}{Shima Imani}, \bibinfo{person}{Liang Du}, {and} \bibinfo{person}{Harsh Shrivastava}.} \bibinfo{year}{2023}\natexlab{}.
\newblock \showarticletitle{Mathprompter: Mathematical reasoning using large language models}.
\newblock \bibinfo{journal}{\emph{arXiv}} (\bibinfo{year}{2023}).
\newblock


\bibitem[Irving et~al\mbox{.}(2016)]%
        {irving2016deepmath}
\bibfield{author}{\bibinfo{person}{Geoffrey Irving}, \bibinfo{person}{Christian Szegedy}, \bibinfo{person}{Alexander~A Alemi}, \bibinfo{person}{Niklas E{\'e}n}, \bibinfo{person}{Fran{\c{c}}ois Chollet}, {and} \bibinfo{person}{Josef Urban}.} \bibinfo{year}{2016}\natexlab{}.
\newblock \showarticletitle{Deepmath-deep sequence models for premise selection}.
\newblock \bibinfo{journal}{\emph{NeurIPS}}  \bibinfo{volume}{29} (\bibinfo{year}{2016}).
\newblock


\bibitem[Jelassi et~al\mbox{.}(2023)]%
        {jelassiLengthGeneralizationArithmetic2023}
\bibfield{author}{\bibinfo{person}{Samy Jelassi}, \bibinfo{person}{St{\'e}phane d'Ascoli}, \bibinfo{person}{Carles Domingo-Enrich}, \bibinfo{person}{Yuhuai Wu}, \bibinfo{person}{Yuanzhi Li}, {and} \bibinfo{person}{Fran{\c{c}}ois Charton}.} \bibinfo{year}{2023}\natexlab{}.
\newblock \showarticletitle{Length generalization in arithmetic transformers}.
\newblock \bibinfo{journal}{\emph{arXiv}} (\bibinfo{year}{2023}).
\newblock


\bibitem[Ji et~al\mbox{.}(2023)]%
        {ji2023survey}
\bibfield{author}{\bibinfo{person}{Ziwei Ji}, \bibinfo{person}{Nayeon Lee}, \bibinfo{person}{Rita Frieske}, \bibinfo{person}{Tiezheng Yu}, \bibinfo{person}{Dan Su}, \bibinfo{person}{Yan Xu}, \bibinfo{person}{Etsuko Ishii}, \bibinfo{person}{Ye~Jin Bang}, \bibinfo{person}{Andrea Madotto}, {and} \bibinfo{person}{Pascale Fung}.} \bibinfo{year}{2023}\natexlab{}.
\newblock \showarticletitle{Survey of hallucination in natural language generation}.
\newblock \bibinfo{journal}{\emph{Comput. Surveys}} \bibinfo{volume}{55}, \bibinfo{number}{12} (\bibinfo{year}{2023}), \bibinfo{pages}{1--38}.
\newblock


\bibitem[Jiang et~al\mbox{.}(2021)]%
        {jiang2021lisa}
\bibfield{author}{\bibinfo{person}{Albert~Qiaochu Jiang}, \bibinfo{person}{Wenda Li}, \bibinfo{person}{Jesse~Michael Han}, {and} \bibinfo{person}{Yuhuai Wu}.} \bibinfo{year}{2021}\natexlab{}.
\newblock \showarticletitle{LISA: Language models of ISAbelle proofs}. In \bibinfo{booktitle}{\emph{AITP}}. \bibinfo{pages}{378--392}.
\newblock


\bibitem[Jiang et~al\mbox{.}(2022)]%
        {jiang2022thor}
\bibfield{author}{\bibinfo{person}{Albert~Q. Jiang}, \bibinfo{person}{Wenda Li}, \bibinfo{person}{Szymon Tworkowski}, \bibinfo{person}{Konrad Czechowski}, \bibinfo{person}{Tomasz Odrzygóźdź}, \bibinfo{person}{Piotr Miłoś}, \bibinfo{person}{Yuhuai Wu}, {and} \bibinfo{person}{Mateja Jamnik}.} \bibinfo{year}{2022}\natexlab{}.
\newblock \bibinfo{title}{Thor: Wielding Hammers to Integrate Language Models and Automated Theorem Provers}.
\newblock
\newblock
\showeprint[arxiv]{2205.10893}


\bibitem[Jiang et~al\mbox{.}(2023b)]%
        {jiang2022draft}
\bibfield{author}{\bibinfo{person}{Albert~Qiaochu Jiang}, \bibinfo{person}{Sean Welleck}, \bibinfo{person}{Jin~Peng Zhou}, \bibinfo{person}{Timothee Lacroix}, \bibinfo{person}{Jiacheng Liu}, \bibinfo{person}{Wenda Li}, \bibinfo{person}{Mateja Jamnik}, \bibinfo{person}{Guillaume Lample}, {and} \bibinfo{person}{Yuhuai Wu}.} \bibinfo{year}{2023}\natexlab{b}.
\newblock \showarticletitle{Draft, Sketch, and Prove: Guiding Formal Theorem Provers with Informal Proofs}. In \bibinfo{booktitle}{\emph{ICLR}}.
\newblock


\bibitem[Jiang et~al\mbox{.}(2024)]%
        {jiang2024leanreasoner}
\bibfield{author}{\bibinfo{person}{Dongwei Jiang}, \bibinfo{person}{Marcio Fonseca}, {and} \bibinfo{person}{Shay~B Cohen}.} \bibinfo{year}{2024}\natexlab{}.
\newblock \showarticletitle{LeanReasoner: Boosting Complex Logical Reasoning with Lean}.
\newblock \bibinfo{journal}{\emph{arXiv}} (\bibinfo{year}{2024}).
\newblock


\bibitem[Jiang et~al\mbox{.}(2023a)]%
        {jiang2023resprompt}
\bibfield{author}{\bibinfo{person}{Song Jiang}, \bibinfo{person}{Zahra Shakeri}, \bibinfo{person}{Aaron Chan}, \bibinfo{person}{Maziar Sanjabi}, \bibinfo{person}{Hamed Firooz}, \bibinfo{person}{Yinglong Xia}, \bibinfo{person}{Bugra Akyildiz}, \bibinfo{person}{Yizhou Sun}, \bibinfo{person}{Jinchao Li}, \bibinfo{person}{Qifan Wang}, {et~al\mbox{.}}} \bibinfo{year}{2023}\natexlab{a}.
\newblock \showarticletitle{Resprompt: Residual connection prompting advances multi-step reasoning in large language models}.
\newblock \bibinfo{journal}{\emph{arXiv}} (\bibinfo{year}{2023}).
\newblock


\bibitem[Jie et~al\mbox{.}(2022)]%
        {jieLearningReasonDeductively2022}
\bibfield{author}{\bibinfo{person}{Zhanming Jie}, \bibinfo{person}{Jierui Li}, {and} \bibinfo{person}{Wei Lu}.} \bibinfo{year}{2022}\natexlab{}.
\newblock \bibinfo{title}{Learning to Reason Deductively: {{Math}} Word Problem Solving as Complex Relation Extraction}.
\newblock
\newblock
\showeprint{2203.10316}


\bibitem[Kaliszyk et~al\mbox{.}(2017)]%
        {kaliszyk2016holstep}
\bibfield{author}{\bibinfo{person}{Cezary Kaliszyk}, \bibinfo{person}{Fran{\c{c}}ois Chollet}, {and} \bibinfo{person}{Christian Szegedy}.} \bibinfo{year}{2017}\natexlab{}.
\newblock \showarticletitle{HolStep: A Machine Learning Dataset for Higher-order Logic Theorem Proving}. In \bibinfo{booktitle}{\emph{ICLR}}.
\newblock


\bibitem[Kalyan et~al\mbox{.}(2021)]%
        {kalyan2021much}
\bibfield{author}{\bibinfo{person}{Ashwin Kalyan}, \bibinfo{person}{Abhinav Kumar}, \bibinfo{person}{Arjun Chandrasekaran}, \bibinfo{person}{Ashish Sabharwal}, {and} \bibinfo{person}{Peter Clark}.} \bibinfo{year}{2021}\natexlab{}.
\newblock \showarticletitle{How Much Coffee Was Consumed During EMNLP 2019? Fermi Problems: A New Reasoning Challenge for AI}. In \bibinfo{booktitle}{\emph{EMNLP}}. \bibinfo{pages}{7318--7328}.
\newblock


\bibitem[Kaplan et~al\mbox{.}(2020)]%
        {kaplan2020scaling}
\bibfield{author}{\bibinfo{person}{Jared Kaplan}, \bibinfo{person}{Sam McCandlish}, \bibinfo{person}{Tom Henighan}, \bibinfo{person}{Tom~B Brown}, \bibinfo{person}{Benjamin Chess}, \bibinfo{person}{Rewon Child}, \bibinfo{person}{Scott Gray}, \bibinfo{person}{Alec Radford}, \bibinfo{person}{Jeffrey Wu}, {and} \bibinfo{person}{Dario Amodei}.} \bibinfo{year}{2020}\natexlab{}.
\newblock \showarticletitle{Scaling laws for neural language models}.
\newblock \bibinfo{journal}{\emph{arXiv}} (\bibinfo{year}{2020}).
\newblock


\bibitem[Kazemi et~al\mbox{.}(2022)]%
        {kazemi2022lambada}
\bibfield{author}{\bibinfo{person}{Mehran Kazemi}, \bibinfo{person}{Najoung Kim}, \bibinfo{person}{Deepti Bhatia}, \bibinfo{person}{Xin Xu}, {and} \bibinfo{person}{Deepak Ramachandran}.} \bibinfo{year}{2022}\natexlab{}.
\newblock \showarticletitle{Lambada: Backward chaining for automated reasoning in natural language}.
\newblock \bibinfo{journal}{\emph{arXiv}} (\bibinfo{year}{2022}).
\newblock


\bibitem[Kenton and Toutanova(2019)]%
        {kenton2019bert}
\bibfield{author}{\bibinfo{person}{Jacob Devlin Ming-Wei~Chang Kenton} {and} \bibinfo{person}{Lee~Kristina Toutanova}.} \bibinfo{year}{2019}\natexlab{}.
\newblock \showarticletitle{BERT: Pre-training of Deep Bidirectional Transformers for Language Understanding}. In \bibinfo{booktitle}{\emph{NAACL}}. \bibinfo{pages}{4171--4186}.
\newblock


\bibitem[Khalifa et~al\mbox{.}(2023)]%
        {khalifa2023discriminator}
\bibfield{author}{\bibinfo{person}{Muhammad Khalifa}, \bibinfo{person}{Lajanugen Logeswaran}, \bibinfo{person}{Moontae Lee}, \bibinfo{person}{Honglak Lee}, {and} \bibinfo{person}{Lu Wang}.} \bibinfo{year}{2023}\natexlab{}.
\newblock \showarticletitle{Discriminator-Guided Multi-step Reasoning with Language Models}.
\newblock \bibinfo{journal}{\emph{arXiv}} (\bibinfo{year}{2023}).
\newblock


\bibitem[Kim et~al\mbox{.}(2020)]%
        {kim_point_2020}
\bibfield{author}{\bibinfo{person}{Bugeun Kim}, \bibinfo{person}{Kyung~Seo Ki}, \bibinfo{person}{Donggeon Lee}, {and} \bibinfo{person}{Gahgene Gweon}.} \bibinfo{year}{2020}\natexlab{}.
\newblock \showarticletitle{Point to the {Expression}: {Solving} {Algebraic} {Word} {Problems} using the {Expression}-{Pointer} {Transformer} {Model}}. In \bibinfo{booktitle}{\emph{EMNLP}}. \bibinfo{address}{Online}, \bibinfo{pages}{3768--3779}.
\newblock


\bibitem[Kim et~al\mbox{.}(2022)]%
        {kim2022ept}
\bibfield{author}{\bibinfo{person}{Bugeun Kim}, \bibinfo{person}{Kyung~Seo Ki}, \bibinfo{person}{Sangkyu Rhim}, {and} \bibinfo{person}{Gahgene Gweon}.} \bibinfo{year}{2022}\natexlab{}.
\newblock \showarticletitle{EPT-X: An Expression-Pointer Transformer model that generates eXplanations for numbers}. In \bibinfo{booktitle}{\emph{ACL}}. \bibinfo{pages}{4442--4458}.
\newblock


\bibitem[Kojima et~al\mbox{.}({[n.\,d.]})]%
        {kojima_large_nodate}
\bibfield{author}{\bibinfo{person}{Takeshi Kojima}, \bibinfo{person}{Shixiang~Shane Gu}, \bibinfo{person}{Machel Reid}, \bibinfo{person}{Yutaka Matsuo}, {and} \bibinfo{person}{Yusuke Iwasawa}.} \bibinfo{year}{[n.\,d.]}\natexlab{}.
\newblock \showarticletitle{Large {Language} {Models} are {Zero}-{Shot} {Reasoners}}.
\newblock  (\bibinfo{year}{[n.\,d.]}).
\newblock


\bibitem[Koncel-Kedziorski et~al\mbox{.}(2015)]%
        {koncel2015parsing}
\bibfield{author}{\bibinfo{person}{Rik Koncel-Kedziorski}, \bibinfo{person}{Hannaneh Hajishirzi}, \bibinfo{person}{Ashish Sabharwal}, \bibinfo{person}{Oren Etzioni}, {and} \bibinfo{person}{Siena~Dumas Ang}.} \bibinfo{year}{2015}\natexlab{}.
\newblock \showarticletitle{Parsing algebraic word problems into equations}.
\newblock \bibinfo{journal}{\emph{TACL}}  \bibinfo{volume}{3} (\bibinfo{year}{2015}), \bibinfo{pages}{585--597}.
\newblock


\bibitem[Koncel-Kedziorski et~al\mbox{.}(2016)]%
        {koncel2016mawps}
\bibfield{author}{\bibinfo{person}{Rik Koncel-Kedziorski}, \bibinfo{person}{Subhro Roy}, \bibinfo{person}{Aida Amini}, \bibinfo{person}{Nate Kushman}, {and} \bibinfo{person}{Hannaneh Hajishirzi}.} \bibinfo{year}{2016}\natexlab{}.
\newblock \showarticletitle{MAWPS: A math word problem repository}. In \bibinfo{booktitle}{\emph{NAACL}}. \bibinfo{pages}{1152--1157}.
\newblock


\bibitem[Kumar et~al\mbox{.}(2024)]%
        {kumar2024training}
\bibfield{author}{\bibinfo{person}{Aviral Kumar}, \bibinfo{person}{Vincent Zhuang}, \bibinfo{person}{Rishabh Agarwal}, \bibinfo{person}{Yi Su}, \bibinfo{person}{John~D Co-Reyes}, \bibinfo{person}{Avi Singh}, \bibinfo{person}{Kate Baumli}, \bibinfo{person}{Shariq Iqbal}, \bibinfo{person}{Colton Bishop}, \bibinfo{person}{Rebecca Roelofs}, {et~al\mbox{.}}} \bibinfo{year}{2024}\natexlab{}.
\newblock \showarticletitle{Training language models to self-correct via reinforcement learning}.
\newblock \bibinfo{journal}{\emph{arXiv}} (\bibinfo{year}{2024}).
\newblock


\bibitem[Kushman et~al\mbox{.}(2014)]%
        {kushman-etal-2014-learning}
\bibfield{author}{\bibinfo{person}{Nate Kushman}, \bibinfo{person}{Yoav Artzi}, \bibinfo{person}{Luke Zettlemoyer}, {and} \bibinfo{person}{Regina Barzilay}.} \bibinfo{year}{2014}\natexlab{}.
\newblock \showarticletitle{Learning to Automatically Solve Algebra Word Problems}. In \bibinfo{booktitle}{\emph{ACL}}. \bibinfo{pages}{271--281}.
\newblock


\bibitem[Lai et~al\mbox{.}(2024)]%
        {lai2024step}
\bibfield{author}{\bibinfo{person}{Xin Lai}, \bibinfo{person}{Zhuotao Tian}, \bibinfo{person}{Yukang Chen}, \bibinfo{person}{Senqiao Yang}, \bibinfo{person}{Xiangru Peng}, {and} \bibinfo{person}{Jiaya Jia}.} \bibinfo{year}{2024}\natexlab{}.
\newblock \showarticletitle{Step-dpo: Step-wise preference optimization for long-chain reasoning of llms}.
\newblock \bibinfo{journal}{\emph{arXiv}} (\bibinfo{year}{2024}).
\newblock


\bibitem[Lample et~al\mbox{.}({[n.\,d.]})]%
        {lample_hypertree_nodate}
\bibfield{author}{\bibinfo{person}{Guillaume Lample}, \bibinfo{person}{Marie-Anne Lachaux}, \bibinfo{person}{Thibaut Lavril}, \bibinfo{person}{Xavier Martinet}, \bibinfo{person}{Amaury Hayat}, \bibinfo{person}{Gabriel Ebner}, \bibinfo{person}{Aurélien Rodriguez}, {and} \bibinfo{person}{Timothée Lacroix}.} \bibinfo{year}{[n.\,d.]}\natexlab{}.
\newblock \showarticletitle{{HyperTree} {Proof} {Search} for {Neural} {Theorem} {Proving}}.
\newblock  (\bibinfo{year}{[n.\,d.]}).
\newblock


\bibitem[Lample et~al\mbox{.}(2022)]%
        {lample2022hypertree}
\bibfield{author}{\bibinfo{person}{Guillaume Lample}, \bibinfo{person}{Timothee Lacroix}, \bibinfo{person}{Marie-Anne Lachaux}, \bibinfo{person}{Aurelien Rodriguez}, \bibinfo{person}{Amaury Hayat}, \bibinfo{person}{Thibaut Lavril}, \bibinfo{person}{Gabriel Ebner}, {and} \bibinfo{person}{Xavier Martinet}.} \bibinfo{year}{2022}\natexlab{}.
\newblock \showarticletitle{Hypertree proof search for neural theorem proving}.
\newblock \bibinfo{journal}{\emph{NeurIPS}}  \bibinfo{volume}{35} (\bibinfo{year}{2022}), \bibinfo{pages}{26337--26349}.
\newblock


\bibitem[Lei et~al\mbox{.}(2023)]%
        {lei2023boosting}
\bibfield{author}{\bibinfo{person}{Bin Lei}, \bibinfo{person}{Chunhua Liao}, \bibinfo{person}{Caiwen Ding}, {et~al\mbox{.}}} \bibinfo{year}{2023}\natexlab{}.
\newblock \showarticletitle{Boosting logical reasoning in large language models through a new framework: The graph of thought}.
\newblock \bibinfo{journal}{\emph{arXiv}} (\bibinfo{year}{2023}).
\newblock


\bibitem[Lewis et~al\mbox{.}(2020)]%
        {lewis2020bart}
\bibfield{author}{\bibinfo{person}{Mike Lewis}, \bibinfo{person}{Yinhan Liu}, \bibinfo{person}{Naman Goyal}, \bibinfo{person}{Marjan Ghazvininejad}, \bibinfo{person}{Abdelrahman Mohamed}, \bibinfo{person}{Omer Levy}, \bibinfo{person}{Veselin Stoyanov}, {and} \bibinfo{person}{Luke Zettlemoyer}.} \bibinfo{year}{2020}\natexlab{}.
\newblock \showarticletitle{BART: Denoising Sequence-to-Sequence Pre-training for Natural Language Generation, Translation, and Comprehension}. In \bibinfo{booktitle}{\emph{ACL}}. \bibinfo{pages}{7871--7880}.
\newblock


\bibitem[Lewkowycz et~al\mbox{.}(2022)]%
        {lewkowycz2022solving}
\bibfield{author}{\bibinfo{person}{Aitor Lewkowycz}, \bibinfo{person}{Anders Andreassen}, \bibinfo{person}{David Dohan}, \bibinfo{person}{Ethan Dyer}, \bibinfo{person}{Henryk Michalewski}, \bibinfo{person}{Vinay Ramasesh}, \bibinfo{person}{Ambrose Slone}, \bibinfo{person}{Cem Anil}, \bibinfo{person}{Imanol Schlag}, \bibinfo{person}{Theo Gutman-Solo}, {et~al\mbox{.}}} \bibinfo{year}{2022}\natexlab{}.
\newblock \showarticletitle{Solving quantitative reasoning problems with language models}.
\newblock \bibinfo{journal}{\emph{NeurIPS}}  \bibinfo{volume}{35} (\bibinfo{year}{2022}), \bibinfo{pages}{3843--3857}.
\newblock


\bibitem[Li et~al\mbox{.}(2023a)]%
        {li2023camel}
\bibfield{author}{\bibinfo{person}{Guohao Li}, \bibinfo{person}{Hasan Abed Al~Kader Hammoud}, \bibinfo{person}{Hani Itani}, \bibinfo{person}{Dmitrii Khizbullin}, {and} \bibinfo{person}{Bernard Ghanem}.} \bibinfo{year}{2023}\natexlab{a}.
\newblock \showarticletitle{CAMEL: Communicative Agents for" Mind" Exploration of Large Language Model Society}. In \bibinfo{booktitle}{\emph{NeurIPS}}.
\newblock


\bibitem[Li et~al\mbox{.}(2021)]%
        {li2020isarstep}
\bibfield{author}{\bibinfo{person}{Wenda Li}, \bibinfo{person}{Lei Yu}, \bibinfo{person}{Yuhuai Wu}, {and} \bibinfo{person}{Lawrence~C Paulson}.} \bibinfo{year}{2021}\natexlab{}.
\newblock \showarticletitle{IsarStep: a Benchmark for High-level Mathematical Reasoning}. In \bibinfo{booktitle}{\emph{ICLR}}.
\newblock


\bibitem[Li et~al\mbox{.}(2023b)]%
        {li2023making}
\bibfield{author}{\bibinfo{person}{Yifei Li}, \bibinfo{person}{Zeqi Lin}, \bibinfo{person}{Shizhuo Zhang}, \bibinfo{person}{Qiang Fu}, \bibinfo{person}{Bei Chen}, \bibinfo{person}{Jian-Guang Lou}, {and} \bibinfo{person}{Weizhu Chen}.} \bibinfo{year}{2023}\natexlab{b}.
\newblock \showarticletitle{Making language models better reasoners with step-aware verifier}. In \bibinfo{booktitle}{\emph{ACL}}. \bibinfo{pages}{5315--5333}.
\newblock


\bibitem[Li et~al\mbox{.}(2022)]%
        {liSeekingPatternsNot2022}
\bibfield{author}{\bibinfo{person}{Zhongli Li}, \bibinfo{person}{Wenxuan Zhang}, \bibinfo{person}{Chao Yan}, \bibinfo{person}{Qingyu Zhou}, \bibinfo{person}{Chao Li}, \bibinfo{person}{Hongzhi Liu}, {and} \bibinfo{person}{Yunbo Cao}.} \bibinfo{year}{2022}\natexlab{}.
\newblock \bibinfo{title}{Seeking Patterns, Not Just Memorizing Procedures: {{Contrastive}} Learning for Solving Math Word Problems}.
\newblock
\newblock


\bibitem[Liang et~al\mbox{.}(2022)]%
        {liangMWPBERTNumeracyaugmentedPretraining2022}
\bibfield{author}{\bibinfo{person}{Zhenwen Liang}, \bibinfo{person}{Jipeng Zhang}, \bibinfo{person}{Lei Wang}, \bibinfo{person}{Wei Qin}, \bibinfo{person}{Yunshi Lan}, \bibinfo{person}{Jie Shao}, {and} \bibinfo{person}{Xiangliang Zhang}.} \bibinfo{year}{2022}\natexlab{}.
\newblock \bibinfo{title}{{{MWP-BERT}}: {{Numeracy-augmented}} Pre-Training for Math Word Problem Solving}.
\newblock
\newblock


\bibitem[Lightman et~al\mbox{.}(2024)]%
        {DBLP:conf/iclr/LightmanKBEBLLS24}
\bibfield{author}{\bibinfo{person}{Hunter Lightman}, \bibinfo{person}{Vineet Kosaraju}, \bibinfo{person}{Yuri Burda}, \bibinfo{person}{Harrison Edwards}, \bibinfo{person}{Bowen Baker}, \bibinfo{person}{Teddy Lee}, \bibinfo{person}{Jan Leike}, \bibinfo{person}{John Schulman}, \bibinfo{person}{Ilya Sutskever}, {and} \bibinfo{person}{Karl Cobbe}.} \bibinfo{year}{2024}\natexlab{}.
\newblock \showarticletitle{Let's Verify Step by Step}. In \bibinfo{booktitle}{\emph{ICLR}}.
\newblock


\bibitem[Ling et~al\mbox{.}(2017)]%
        {ling2017program}
\bibfield{author}{\bibinfo{person}{Wang Ling}, \bibinfo{person}{Dani Yogatama}, \bibinfo{person}{Chris Dyer}, {and} \bibinfo{person}{Phil Blunsom}.} \bibinfo{year}{2017}\natexlab{}.
\newblock \showarticletitle{Program Induction by Rationale Generation: Learning to Solve and Explain Algebraic Word Problems}. In \bibinfo{booktitle}{\emph{ACL}}. \bibinfo{pages}{158--167}.
\newblock


\bibitem[Ling et~al\mbox{.}(2023)]%
        {ling2023deductive}
\bibfield{author}{\bibinfo{person}{Zhan Ling}, \bibinfo{person}{Yunhao Fang}, \bibinfo{person}{Xuanlin Li}, \bibinfo{person}{Zhiao Huang}, \bibinfo{person}{Mingu Lee}, \bibinfo{person}{Roland Memisevic}, {and} \bibinfo{person}{Hao Su}.} \bibinfo{year}{2023}\natexlab{}.
\newblock \showarticletitle{Deductive Verification of Chain-of-Thought Reasoning}.
\newblock \bibinfo{journal}{\emph{arXiv}} (\bibinfo{year}{2023}).
\newblock


\bibitem[Liu et~al\mbox{.}(2023a)]%
        {liu2023llm+}
\bibfield{author}{\bibinfo{person}{Bo Liu}, \bibinfo{person}{Yuqian Jiang}, \bibinfo{person}{Xiaohan Zhang}, \bibinfo{person}{Qiang Liu}, \bibinfo{person}{Shiqi Zhang}, \bibinfo{person}{Joydeep Biswas}, {and} \bibinfo{person}{Peter Stone}.} \bibinfo{year}{2023}\natexlab{a}.
\newblock \showarticletitle{Llm+ p: Empowering large language models with optimal planning proficiency}.
\newblock \bibinfo{journal}{\emph{arXiv}} (\bibinfo{year}{2023}).
\newblock


\bibitem[Liu et~al\mbox{.}(2022)]%
        {liu2022makes}
\bibfield{author}{\bibinfo{person}{Jiachang Liu}, \bibinfo{person}{Dinghan Shen}, \bibinfo{person}{Yizhe Zhang}, \bibinfo{person}{Bill Dolan}, \bibinfo{person}{Lawrence Carin}, {and} \bibinfo{person}{Weizhu Chen}.} \bibinfo{year}{2022}\natexlab{}.
\newblock \showarticletitle{What Makes Good In-Context Examples for GPT-3?}. In \bibinfo{booktitle}{\emph{DeeLIO}}. \bibinfo{pages}{100--114}.
\newblock


\bibitem[Liu and Low(2023)]%
        {liu_goat_2023}
\bibfield{author}{\bibinfo{person}{Tiedong Liu} {and} \bibinfo{person}{Bryan Kian~Hsiang Low}.} \bibinfo{year}{2023}\natexlab{}.
\newblock \showarticletitle{Goat: Fine-tuned llama outperforms gpt-4 on arithmetic tasks}.
\newblock \bibinfo{journal}{\emph{arXiv}} (\bibinfo{year}{2023}).
\newblock


\bibitem[Liu et~al\mbox{.}(2024)]%
        {liu2024cmm}
\bibfield{author}{\bibinfo{person}{Wentao Liu}, \bibinfo{person}{Qianjun Pan}, \bibinfo{person}{Yi Zhang}, \bibinfo{person}{Zhuo Liu}, \bibinfo{person}{Ji Wu}, \bibinfo{person}{Jie Zhou}, \bibinfo{person}{Aimin Zhou}, \bibinfo{person}{Qin Chen}, \bibinfo{person}{Bo Jiang}, {and} \bibinfo{person}{Liang He}.} \bibinfo{year}{2024}\natexlab{}.
\newblock \showarticletitle{Cmm-math: A chinese multimodal math dataset to evaluate and enhance the mathematics reasoning of large multimodal models}.
\newblock \bibinfo{journal}{\emph{arXiv}} (\bibinfo{year}{2024}).
\newblock


\bibitem[Liu et~al\mbox{.}(2019)]%
        {liuRoBERTaRobustlyOptimized2019}
\bibfield{author}{\bibinfo{person}{Yinhan Liu}, \bibinfo{person}{Myle Ott}, \bibinfo{person}{Naman Goyal}, \bibinfo{person}{Jingfei Du}, \bibinfo{person}{Mandar Joshi}, \bibinfo{person}{Danqi Chen}, \bibinfo{person}{Omer Levy}, \bibinfo{person}{Mike Lewis}, \bibinfo{person}{Luke Zettlemoyer}, {and} \bibinfo{person}{Veselin Stoyanov}.} \bibinfo{year}{2019}\natexlab{}.
\newblock \bibinfo{title}{{{RoBERTa}}: {{A}} Robustly Optimized {{BERT}} Pretraining Approach}.
\newblock
\newblock


\bibitem[Liu et~al\mbox{.}(2023b)]%
        {liu_improving_2023}
\bibfield{author}{\bibinfo{person}{Yixin Liu}, \bibinfo{person}{Avi Singh}, \bibinfo{person}{C~Daniel Freeman}, \bibinfo{person}{John~D Co-Reyes}, {and} \bibinfo{person}{Peter~J Liu}.} \bibinfo{year}{2023}\natexlab{b}.
\newblock \showarticletitle{Improving large language model fine-tuning for solving math problems}.
\newblock \bibinfo{journal}{\emph{arXiv}} (\bibinfo{year}{2023}).
\newblock


\bibitem[Long(2023)]%
        {long2023large}
\bibfield{author}{\bibinfo{person}{Jieyi Long}.} \bibinfo{year}{2023}\natexlab{}.
\newblock \showarticletitle{Large Language Model Guided Tree-of-Thought}.
\newblock \bibinfo{journal}{\emph{arXiv}} (\bibinfo{year}{2023}).
\newblock


\bibitem[Lu et~al\mbox{.}(2023a)]%
        {lu2023mathvista}
\bibfield{author}{\bibinfo{person}{Pan Lu}, \bibinfo{person}{Hritik Bansal}, \bibinfo{person}{Tony Xia}, \bibinfo{person}{Jiacheng Liu}, \bibinfo{person}{Chunyuan Li}, \bibinfo{person}{Hannaneh Hajishirzi}, \bibinfo{person}{Hao Cheng}, \bibinfo{person}{Kai-Wei Chang}, \bibinfo{person}{Michel Galley}, {and} \bibinfo{person}{Jianfeng Gao}.} \bibinfo{year}{2023}\natexlab{a}.
\newblock \showarticletitle{Mathvista: Evaluating mathematical reasoning of foundation models in visual contexts}.
\newblock \bibinfo{journal}{\emph{arXiv}} (\bibinfo{year}{2023}).
\newblock


\bibitem[Lu et~al\mbox{.}(2021a)]%
        {lu2021inter}
\bibfield{author}{\bibinfo{person}{Pan Lu}, \bibinfo{person}{Ran Gong}, \bibinfo{person}{Shibiao Jiang}, \bibinfo{person}{Liang Qiu}, \bibinfo{person}{Siyuan Huang}, \bibinfo{person}{Xiaodan Liang}, {and} \bibinfo{person}{Song-chun Zhu}.} \bibinfo{year}{2021}\natexlab{a}.
\newblock \showarticletitle{Inter-GPS: Interpretable Geometry Problem Solving with Formal Language and Symbolic Reasoning}. In \bibinfo{booktitle}{\emph{ACL}}. \bibinfo{pages}{6774--6786}.
\newblock


\bibitem[Lu et~al\mbox{.}(2023b)]%
        {lu2023chameleon}
\bibfield{author}{\bibinfo{person}{Pan Lu}, \bibinfo{person}{Baolin Peng}, \bibinfo{person}{Hao Cheng}, \bibinfo{person}{Michel Galley}, \bibinfo{person}{Kai-Wei Chang}, \bibinfo{person}{Ying~Nian Wu}, \bibinfo{person}{Song-Chun Zhu}, {and} \bibinfo{person}{Jianfeng Gao}.} \bibinfo{year}{2023}\natexlab{b}.
\newblock \bibinfo{title}{Chameleon: Plug-and-Play Compositional Reasoning with Large Language Models}.
\newblock
\newblock
\showeprint[arxiv]{2304.09842}


\bibitem[Lu et~al\mbox{.}(2023c)]%
        {lu2022dynamic}
\bibfield{author}{\bibinfo{person}{Pan Lu}, \bibinfo{person}{Liang Qiu}, \bibinfo{person}{Kai-Wei Chang}, \bibinfo{person}{Ying~Nian Wu}, \bibinfo{person}{Song-Chun Zhu}, \bibinfo{person}{Tanmay Rajpurohit}, \bibinfo{person}{Peter Clark}, {and} \bibinfo{person}{Ashwin Kalyan}.} \bibinfo{year}{2023}\natexlab{c}.
\newblock \showarticletitle{Dynamic Prompt Learning via Policy Gradient for Semi-structured Mathematical Reasoning}. In \bibinfo{booktitle}{\emph{ICLR}}.
\newblock


\bibitem[Lu et~al\mbox{.}(2021b)]%
        {lu2021iconqa}
\bibfield{author}{\bibinfo{person}{Pan Lu}, \bibinfo{person}{Liang Qiu}, \bibinfo{person}{Jiaqi Chen}, \bibinfo{person}{Tony Xia}, \bibinfo{person}{Yizhou Zhao}, \bibinfo{person}{Wei Zhang}, \bibinfo{person}{Zhou Yu}, \bibinfo{person}{Xiaodan Liang}, {and} \bibinfo{person}{Song-Chun Zhu}.} \bibinfo{year}{2021}\natexlab{b}.
\newblock \showarticletitle{IconQA: A New Benchmark for Abstract Diagram Understanding and Visual Language Reasoning}. In \bibinfo{booktitle}{\emph{NeurIPS Datasets and Benchmarks Track}}.
\newblock


\bibitem[Lu et~al\mbox{.}(2023d)]%
        {lu-etal-2023-survey}
\bibfield{author}{\bibinfo{person}{Pan Lu}, \bibinfo{person}{Liang Qiu}, \bibinfo{person}{Wenhao Yu}, \bibinfo{person}{Sean Welleck}, {and} \bibinfo{person}{Kai-Wei Chang}.} \bibinfo{year}{2023}\natexlab{d}.
\newblock \showarticletitle{A Survey of Deep Learning for Mathematical Reasoning}. In \bibinfo{booktitle}{\emph{ACL}}, \bibfield{editor}{\bibinfo{person}{Anna Rogers}, \bibinfo{person}{Jordan Boyd-Graber}, {and} \bibinfo{person}{Naoaki Okazaki}} (Eds.). \bibinfo{pages}{14605--14631}.
\newblock


\bibitem[Luo et~al\mbox{.}(2023)]%
        {luo2023wizardmath}
\bibfield{author}{\bibinfo{person}{Haipeng Luo}, \bibinfo{person}{Qingfeng Sun}, \bibinfo{person}{Can Xu}, \bibinfo{person}{Pu Zhao}, \bibinfo{person}{Jianguang Lou}, \bibinfo{person}{Chongyang Tao}, \bibinfo{person}{Xiubo Geng}, \bibinfo{person}{Qingwei Lin}, \bibinfo{person}{Shifeng Chen}, {and} \bibinfo{person}{Dongmei Zhang}.} \bibinfo{year}{2023}\natexlab{}.
\newblock \showarticletitle{Wizardmath: Empowering mathematical reasoning for large language models via reinforced evol-instruct}.
\newblock \bibinfo{journal}{\emph{arXiv}} (\bibinfo{year}{2023}).
\newblock


\bibitem[Luo et~al\mbox{.}(2024)]%
        {luo2024improve}
\bibfield{author}{\bibinfo{person}{Liangchen Luo}, \bibinfo{person}{Yinxiao Liu}, \bibinfo{person}{Rosanne Liu}, \bibinfo{person}{Samrat Phatale}, \bibinfo{person}{Harsh Lara}, \bibinfo{person}{Yunxuan Li}, \bibinfo{person}{Lei Shu}, \bibinfo{person}{Yun Zhu}, \bibinfo{person}{Lei Meng}, \bibinfo{person}{Jiao Sun}, {et~al\mbox{.}}} \bibinfo{year}{2024}\natexlab{}.
\newblock \showarticletitle{Improve Mathematical Reasoning in Language Models by Automated Process Supervision}.
\newblock \bibinfo{journal}{\emph{arXiv}} (\bibinfo{year}{2024}).
\newblock


\bibitem[Luong et~al\mbox{.}(2024)]%
        {luong2024reft}
\bibfield{author}{\bibinfo{person}{Trung~Quoc Luong}, \bibinfo{person}{Xinbo Zhang}, \bibinfo{person}{Zhanming Jie}, \bibinfo{person}{Peng Sun}, \bibinfo{person}{Xiaoran Jin}, {and} \bibinfo{person}{Hang Li}.} \bibinfo{year}{2024}\natexlab{}.
\newblock \showarticletitle{Reft: Reasoning with reinforced fine-tuning}.
\newblock \bibinfo{journal}{\emph{arXiv}} (\bibinfo{year}{2024}).
\newblock


\bibitem[Ma et~al\mbox{.}(2023)]%
        {ma2023let}
\bibfield{author}{\bibinfo{person}{Qianli Ma}, \bibinfo{person}{Haotian Zhou}, \bibinfo{person}{Tingkai Liu}, \bibinfo{person}{Jianbo Yuan}, \bibinfo{person}{Pengfei Liu}, \bibinfo{person}{Yang You}, {and} \bibinfo{person}{Hongxia Yang}.} \bibinfo{year}{2023}\natexlab{}.
\newblock \showarticletitle{Let's reward step by step: Step-Level reward model as the Navigators for Reasoning}.
\newblock \bibinfo{journal}{\emph{arXiv}} (\bibinfo{year}{2023}).
\newblock


\bibitem[Macina et~al\mbox{.}(2023)]%
        {macina2023mathdial}
\bibfield{author}{\bibinfo{person}{Jakub Macina}, \bibinfo{person}{Nico Daheim}, \bibinfo{person}{Sankalan~Pal Chowdhury}, \bibinfo{person}{Tanmay Sinha}, \bibinfo{person}{Manu Kapur}, \bibinfo{person}{Iryna Gurevych}, {and} \bibinfo{person}{Mrinmaya Sachan}.} \bibinfo{year}{2023}\natexlab{}.
\newblock \showarticletitle{Mathdial: A dialogue tutoring dataset with rich pedagogical properties grounded in math reasoning problems}.
\newblock \bibinfo{journal}{\emph{arXiv}} (\bibinfo{year}{2023}).
\newblock


\bibitem[Madaan et~al\mbox{.}(2023)]%
        {madaan2023self}
\bibfield{author}{\bibinfo{person}{Aman Madaan}, \bibinfo{person}{Niket Tandon}, \bibinfo{person}{Prakhar Gupta}, \bibinfo{person}{Skyler Hallinan}, \bibinfo{person}{Luyu Gao}, \bibinfo{person}{Sarah Wiegreffe}, \bibinfo{person}{Uri Alon}, \bibinfo{person}{Nouha Dziri}, \bibinfo{person}{Shrimai Prabhumoye}, \bibinfo{person}{Yiming Yang}, {et~al\mbox{.}}} \bibinfo{year}{2023}\natexlab{}.
\newblock \showarticletitle{Self-refine: Iterative refinement with self-feedback}.
\newblock \bibinfo{journal}{\emph{arXiv}} (\bibinfo{year}{2023}).
\newblock


\bibitem[Matzakos et~al\mbox{.}(2023)]%
        {matzakos2023learning}
\bibfield{author}{\bibinfo{person}{Nikolaos Matzakos}, \bibinfo{person}{Spyridon Doukakis}, {and} \bibinfo{person}{Maria Moundridou}.} \bibinfo{year}{2023}\natexlab{}.
\newblock \showarticletitle{Learning Mathematics with Large Language Models: A Comparative Study with Computer Algebra Systems and Other Tools}.
\newblock \bibinfo{journal}{\emph{International Journal of Emerging Technologies in Learning (Online)}} \bibinfo{volume}{18}, \bibinfo{number}{20} (\bibinfo{year}{2023}), \bibinfo{pages}{51}.
\newblock


\bibitem[Maynez et~al\mbox{.}(2020)]%
        {maynez-etal-2020-faithfulness}
\bibfield{author}{\bibinfo{person}{Joshua Maynez}, \bibinfo{person}{Shashi Narayan}, \bibinfo{person}{Bernd Bohnet}, {and} \bibinfo{person}{Ryan McDonald}.} \bibinfo{year}{2020}\natexlab{}.
\newblock \showarticletitle{On Faithfulness and Factuality in Abstractive Summarization}. In \bibinfo{booktitle}{\emph{ACL}}. \bibinfo{pages}{1906--1919}.
\newblock


\bibitem[Megill and Wheeler(2019)]%
        {megill2019metamath}
\bibfield{author}{\bibinfo{person}{Norman Megill} {and} \bibinfo{person}{David~A Wheeler}.} \bibinfo{year}{2019}\natexlab{}.
\newblock \bibinfo{booktitle}{\emph{Metamath: a computer language for mathematical proofs}}.
\newblock \bibinfo{publisher}{Lulu. com}.
\newblock


\bibitem[Miao et~al\mbox{.}(2023)]%
        {miao2023selfcheck}
\bibfield{author}{\bibinfo{person}{Ning Miao}, \bibinfo{person}{Yee~Whye Teh}, {and} \bibinfo{person}{Tom Rainforth}.} \bibinfo{year}{2023}\natexlab{}.
\newblock \showarticletitle{Selfcheck: Using llms to zero-shot check their own step-by-step reasoning}.
\newblock \bibinfo{journal}{\emph{arXiv}} (\bibinfo{year}{2023}).
\newblock


\bibitem[Miao et~al\mbox{.}(2020)]%
        {miao2020diverse}
\bibfield{author}{\bibinfo{person}{Shen-Yun Miao}, \bibinfo{person}{Chao-Chun Liang}, {and} \bibinfo{person}{Keh-Yih Su}.} \bibinfo{year}{2020}\natexlab{}.
\newblock \showarticletitle{A Diverse Corpus for Evaluating and Developing English Math Word Problem Solvers}. In \bibinfo{booktitle}{\emph{ACL}}. \bibinfo{pages}{975--984}.
\newblock


\bibitem[Min et~al\mbox{.}(2023)]%
        {10.1145/3605943}
\bibfield{author}{\bibinfo{person}{Bonan Min}, \bibinfo{person}{Hayley Ross}, \bibinfo{person}{Elior Sulem}, \bibinfo{person}{Amir Pouran~Ben Veyseh}, \bibinfo{person}{Thien~Huu Nguyen}, \bibinfo{person}{Oscar Sainz}, \bibinfo{person}{Eneko Agirre}, \bibinfo{person}{Ilana Heintz}, {and} \bibinfo{person}{Dan Roth}.} \bibinfo{year}{2023}\natexlab{}.
\newblock \showarticletitle{Recent Advances in Natural Language Processing via Large Pre-Trained Language Models: A Survey}.
\newblock \bibinfo{journal}{\emph{ACM Comput. Surv.}} \bibinfo{volume}{56}, \bibinfo{number}{2} (\bibinfo{year}{2023}).
\newblock


\bibitem[Min et~al\mbox{.}(2022)]%
        {min2022metaicl}
\bibfield{author}{\bibinfo{person}{Sewon Min}, \bibinfo{person}{Mike Lewis}, \bibinfo{person}{Luke Zettlemoyer}, {and} \bibinfo{person}{Hannaneh Hajishirzi}.} \bibinfo{year}{2022}\natexlab{}.
\newblock \showarticletitle{MetaICL: Learning to Learn In Context}. In \bibinfo{booktitle}{\emph{NAACL}}. \bibinfo{pages}{2791--2809}.
\newblock


\bibitem[Mishra et~al\mbox{.}(2022a)]%
        {mishra2022lila}
\bibfield{author}{\bibinfo{person}{Swaroop Mishra}, \bibinfo{person}{Matthew Finlayson}, \bibinfo{person}{Pan Lu}, \bibinfo{person}{Leonard Tang}, \bibinfo{person}{Sean Welleck}, \bibinfo{person}{Chitta Baral}, \bibinfo{person}{Tanmay Rajpurohit}, \bibinfo{person}{Oyvind Tafjord}, \bibinfo{person}{Ashish Sabharwal}, \bibinfo{person}{Peter Clark}, {et~al\mbox{.}}} \bibinfo{year}{2022}\natexlab{a}.
\newblock \showarticletitle{LILA: A Unified Benchmark for Mathematical Reasoning}. In \bibinfo{booktitle}{\emph{EMNLP}}. \bibinfo{pages}{5807--5832}.
\newblock


\bibitem[Mishra et~al\mbox{.}(2022b)]%
        {mishra2022numglue}
\bibfield{author}{\bibinfo{person}{Swaroop Mishra}, \bibinfo{person}{Arindam Mitra}, \bibinfo{person}{Neeraj Varshney}, \bibinfo{person}{Bhavdeep Sachdeva}, \bibinfo{person}{Peter Clark}, \bibinfo{person}{Chitta Baral}, {and} \bibinfo{person}{Ashwin Kalyan}.} \bibinfo{year}{2022}\natexlab{b}.
\newblock \showarticletitle{NUMGLUE: A Suite of Fundamental yet Challenging Mathematical Reasoning Tasks}. In \bibinfo{booktitle}{\emph{ACL}}. \bibinfo{pages}{3505--3523}.
\newblock


\bibitem[Mitra and Baral(2016)]%
        {mitra2016learning}
\bibfield{author}{\bibinfo{person}{Arindam Mitra} {and} \bibinfo{person}{Chitta Baral}.} \bibinfo{year}{2016}\natexlab{}.
\newblock \showarticletitle{Learning to use formulas to solve simple arithmetic problems}. In \bibinfo{booktitle}{\emph{ACL}}. \bibinfo{pages}{2144--2153}.
\newblock


\bibitem[Mo and Xin(2023)]%
        {mo2023tree}
\bibfield{author}{\bibinfo{person}{Shentong Mo} {and} \bibinfo{person}{Miao Xin}.} \bibinfo{year}{2023}\natexlab{}.
\newblock \showarticletitle{Tree of Uncertain Thoughts Reasoning for Large Language Models}.
\newblock \bibinfo{journal}{\emph{arXiv}} (\bibinfo{year}{2023}).
\newblock


\bibitem[Muffo et~al\mbox{.}({[n.\,d.]})]%
        {muffo_evaluating_nodate}
\bibfield{author}{\bibinfo{person}{Matteo Muffo}, \bibinfo{person}{Aldo Cocco}, {and} \bibinfo{person}{Enrico Bertino}.} \bibinfo{year}{[n.\,d.]}\natexlab{}.
\newblock \showarticletitle{Evaluating {Transformer} {Language} {Models} on {Arithmetic} {Operations} {Using} {Number} {Decomposition}}.
\newblock  (\bibinfo{year}{[n.\,d.]}).
\newblock


\bibitem[Naik et~al\mbox{.}(2023)]%
        {naik2023diversity}
\bibfield{author}{\bibinfo{person}{Ranjita Naik}, \bibinfo{person}{Varun Chandrasekaran}, \bibinfo{person}{Mert Yuksekgonul}, \bibinfo{person}{Hamid Palangi}, {and} \bibinfo{person}{Besmira Nushi}.} \bibinfo{year}{2023}\natexlab{}.
\newblock \showarticletitle{Diversity of Thought Improves Reasoning Abilities of Large Language Models}.
\newblock \bibinfo{journal}{\emph{arXiv}} (\bibinfo{year}{2023}).
\newblock


\bibitem[NAKAMOTO et~al\mbox{.}(2023)]%
        {nakamoto2023enhancing}
\bibfield{author}{\bibinfo{person}{Ryosuke NAKAMOTO}, \bibinfo{person}{Brendan Flanagan}, \bibinfo{person}{Taisei Yamauchi}, \bibinfo{person}{Dai Yilling}, \bibinfo{person}{Kyosuke Takami}, {and} \bibinfo{person}{Horoaki Ogata}.} \bibinfo{year}{2023}\natexlab{}.
\newblock \showarticletitle{Enhancing Automated Scoring of Math Self-Explanation Quality using LLM-Generated Datasets: A Semi-Supervised Approach}.
\newblock  (\bibinfo{year}{2023}).
\newblock


\bibitem[Nelson(1980)]%
        {nelson1980socratic}
\bibfield{author}{\bibinfo{person}{Leonard Nelson}.} \bibinfo{year}{1980}\natexlab{}.
\newblock \showarticletitle{The socratic method}.
\newblock \bibinfo{journal}{\emph{Thinking: The Journal of Philosophy for Children}} \bibinfo{volume}{2}, \bibinfo{number}{2}, \bibinfo{pages}{34--38}.
\newblock


\bibitem[Newell et~al\mbox{.}(1957)]%
        {10.1145/1455567.1455605}
\bibfield{author}{\bibinfo{person}{A. Newell}, \bibinfo{person}{J.~C. Shaw}, {and} \bibinfo{person}{H.~A. Simon}.} \bibinfo{year}{1957}\natexlab{}.
\newblock \showarticletitle{Empirical Explorations of the Logic Theory Machine: A Case Study in Heuristic}. \bibinfo{pages}{218–230}.
\newblock


\bibitem[Nogueira et~al\mbox{.}(2021)]%
        {nogueira2021investigating}
\bibfield{author}{\bibinfo{person}{Rodrigo Nogueira}, \bibinfo{person}{Zhiying Jiang}, {and} \bibinfo{person}{Jimmy Lin}.} \bibinfo{year}{2021}\natexlab{}.
\newblock \showarticletitle{Investigating the limitations of transformers with simple arithmetic tasks}.
\newblock \bibinfo{journal}{\emph{arXiv}} (\bibinfo{year}{2021}).
\newblock


\bibitem[Noorbakhsh et~al\mbox{.}(2021)]%
        {noorbakhsh2021pretrained}
\bibfield{author}{\bibinfo{person}{Kimia Noorbakhsh}, \bibinfo{person}{Modar Sulaiman}, \bibinfo{person}{Mahdi Sharifi}, \bibinfo{person}{Kallol Roy}, {and} \bibinfo{person}{Pooyan Jamshidi}.} \bibinfo{year}{2021}\natexlab{}.
\newblock \showarticletitle{Pretrained Language Models are Symbolic Mathematics Solvers too!}
\newblock \bibinfo{journal}{\emph{arXiv}} (\bibinfo{year}{2021}).
\newblock


\bibitem[Nye et~al\mbox{.}(2021)]%
        {nye_show_2021}
\bibfield{author}{\bibinfo{person}{Maxwell Nye}, \bibinfo{person}{Anders~Johan Andreassen}, \bibinfo{person}{Guy Gur-Ari}, \bibinfo{person}{Henryk Michalewski}, \bibinfo{person}{Jacob Austin}, \bibinfo{person}{David Bieber}, \bibinfo{person}{David Dohan}, \bibinfo{person}{Aitor Lewkowycz}, \bibinfo{person}{Maarten Bosma}, \bibinfo{person}{David Luan}, \bibinfo{person}{Charles Sutton}, {and} \bibinfo{person}{Augustus Odena}.} \bibinfo{year}{2021}\natexlab{}.
\newblock \bibinfo{title}{Show {Your} {Work}: {Scratchpads} for {Intermediate} {Computation} with {Language} {Models}}.
\newblock
\newblock


\bibitem[Olausson et~al\mbox{.}(2023)]%
        {olausson2023linc}
\bibfield{author}{\bibinfo{person}{Theo~X Olausson}, \bibinfo{person}{Alex Gu}, \bibinfo{person}{Benjamin Lipkin}, \bibinfo{person}{Cedegao~E Zhang}, \bibinfo{person}{Armando Solar-Lezama}, \bibinfo{person}{Joshua~B Tenenbaum}, {and} \bibinfo{person}{Roger Levy}.} \bibinfo{year}{2023}\natexlab{}.
\newblock \showarticletitle{LINC: A neurosymbolic approach for logical reasoning by combining language models with first-order logic provers}.
\newblock \bibinfo{journal}{\emph{arXiv}} (\bibinfo{year}{2023}).
\newblock


\bibitem[OpenAI(2023)]%
        {openai2023gpt4}
\bibfield{author}{\bibinfo{person}{OpenAI}.} \bibinfo{year}{2023}\natexlab{}.
\newblock \bibinfo{title}{GPT-4 Technical Report}.
\newblock
\newblock
\showeprint[arxiv]{2303.08774}


\bibitem[{OpenAI}(2024)]%
        {openai_o1_system_card}
\bibfield{author}{\bibinfo{person}{{OpenAI}}.} \bibinfo{year}{2024}\natexlab{}.
\newblock \bibinfo{title}{{OpenAI O1 System Card}}.
\newblock
\newblock
\urldef\tempurl%
\url{https://openai.com/index/openai-o1-system-card/}
\showURL{%
\tempurl}


\bibitem[Paranjape et~al\mbox{.}(2023)]%
        {paranjape2023art}
\bibfield{author}{\bibinfo{person}{Bhargavi Paranjape}, \bibinfo{person}{Scott Lundberg}, \bibinfo{person}{Sameer Singh}, \bibinfo{person}{Hannaneh Hajishirzi}, \bibinfo{person}{Luke Zettlemoyer}, {and} \bibinfo{person}{Marco~Tulio Ribeiro}.} \bibinfo{year}{2023}\natexlab{}.
\newblock \bibinfo{title}{ART: Automatic multi-step reasoning and tool-use for large language models}.
\newblock
\newblock
\showeprint[arxiv]{2303.09014}


\bibitem[Parisi et~al\mbox{.}(2022)]%
        {parisi2022talm}
\bibfield{author}{\bibinfo{person}{Aaron Parisi}, \bibinfo{person}{Yao Zhao}, {and} \bibinfo{person}{Noah Fiedel}.} \bibinfo{year}{2022}\natexlab{}.
\newblock \bibinfo{title}{TALM: Tool Augmented Language Models}.
\newblock
\newblock
\showeprint[arxiv]{2205.12255}


\bibitem[Paster et~al\mbox{.}(2023)]%
        {paster2023openwebmath}
\bibfield{author}{\bibinfo{person}{Keiran Paster}, \bibinfo{person}{Marco~Dos Santos}, \bibinfo{person}{Zhangir Azerbayev}, {and} \bibinfo{person}{Jimmy Ba}.} \bibinfo{year}{2023}\natexlab{}.
\newblock \bibinfo{title}{OpenWebMath: An Open Dataset of High-Quality Mathematical Web Text}.
\newblock
\newblock
\showeprint[arxiv]{2310.06786}


\bibitem[Patel et~al\mbox{.}(2021)]%
        {patel2021nlp}
\bibfield{author}{\bibinfo{person}{Arkil Patel}, \bibinfo{person}{Satwik Bhattamishra}, {and} \bibinfo{person}{Navin Goyal}.} \bibinfo{year}{2021}\natexlab{}.
\newblock \showarticletitle{Are NLP Models really able to Solve Simple Math Word Problems?}. In \bibinfo{booktitle}{\emph{NAACL}}. \bibinfo{pages}{2080--2094}.
\newblock


\bibitem[Peng et~al\mbox{.}(2021)]%
        {pengMathBERTPretrainedModel2021}
\bibfield{author}{\bibinfo{person}{Shuai Peng}, \bibinfo{person}{Ke Yuan}, \bibinfo{person}{Liangcai Gao}, {and} \bibinfo{person}{Zhi Tang}.} \bibinfo{year}{2021}\natexlab{}.
\newblock \bibinfo{title}{{{MathBERT}}: {{A}} Pre-Trained Model for Mathematical Formula Understanding}.
\newblock
\newblock


\bibitem[Pitis et~al\mbox{.}(2023)]%
        {pitis_boosted_2023}
\bibfield{author}{\bibinfo{person}{Silviu Pitis}, \bibinfo{person}{Michael~R Zhang}, \bibinfo{person}{Andrew Wang}, {and} \bibinfo{person}{Jimmy Ba}.} \bibinfo{year}{2023}\natexlab{}.
\newblock \showarticletitle{Boosted prompt ensembles for large language models}.
\newblock \bibinfo{journal}{\emph{arXiv}} (\bibinfo{year}{2023}).
\newblock


\bibitem[Polu and Sutskever(2020a)]%
        {polu2020generative}
\bibfield{author}{\bibinfo{person}{Stanislas Polu} {and} \bibinfo{person}{Ilya Sutskever}.} \bibinfo{year}{2020}\natexlab{a}.
\newblock \bibinfo{title}{Generative Language Modeling for Automated Theorem Proving}.
\newblock
\newblock
\showeprint[arxiv]{2009.03393}


\bibitem[Polu and Sutskever(2020b)]%
        {DBLP:journals/corr/abs-2009-03393}
\bibfield{author}{\bibinfo{person}{Stanislas Polu} {and} \bibinfo{person}{Ilya Sutskever}.} \bibinfo{year}{2020}\natexlab{b}.
\newblock \showarticletitle{Generative Language Modeling for Automated Theorem Proving}.
\newblock \bibinfo{journal}{\emph{CoRR}}  \bibinfo{volume}{abs/2009.03393} (\bibinfo{year}{2020}).
\newblock


\bibitem[Qi et~al\mbox{.}(2023)]%
        {qi2023art}
\bibfield{author}{\bibinfo{person}{Jingyuan Qi}, \bibinfo{person}{Zhiyang Xu}, \bibinfo{person}{Ying Shen}, \bibinfo{person}{Minqian Liu}, \bibinfo{person}{dingnan jin}, \bibinfo{person}{Qifan Wang}, {and} \bibinfo{person}{Lifu Huang}.} \bibinfo{year}{2023}\natexlab{}.
\newblock \showarticletitle{The Art of SOCRATIC QUESTIONING: Recursive Thinking with Large Language Models}. In \bibinfo{booktitle}{\emph{EMNLP}}.
\newblock


\bibitem[Qian et~al\mbox{.}(2023)]%
        {qian2023creator}
\bibfield{author}{\bibinfo{person}{Cheng Qian}, \bibinfo{person}{Chi Han}, \bibinfo{person}{Yi~R Fung}, \bibinfo{person}{Yujia Qin}, \bibinfo{person}{Zhiyuan Liu}, {and} \bibinfo{person}{Heng Ji}.} \bibinfo{year}{2023}\natexlab{}.
\newblock \showarticletitle{Creator: Tool creation for disentangling abstract and concrete reasoning of large language models}.
\newblock \bibinfo{journal}{\emph{arXiv}} (\bibinfo{year}{2023}).
\newblock


\bibitem[Qiao et~al\mbox{.}(2024)]%
        {qiao2024wemathdoeslargemultimodal}
\bibfield{author}{\bibinfo{person}{Runqi Qiao}, \bibinfo{person}{Qiuna Tan}, \bibinfo{person}{Guanting Dong}, \bibinfo{person}{Minhui Wu}, \bibinfo{person}{Chong Sun}, \bibinfo{person}{Xiaoshuai Song}, \bibinfo{person}{Zhuoma GongQue}, \bibinfo{person}{Shanglin Lei}, \bibinfo{person}{Zhe Wei}, \bibinfo{person}{Miaoxuan Zhang}, \bibinfo{person}{Runfeng Qiao}, \bibinfo{person}{Yifan Zhang}, \bibinfo{person}{Xiao Zong}, \bibinfo{person}{Yida Xu}, \bibinfo{person}{Muxi Diao}, \bibinfo{person}{Zhimin Bao}, \bibinfo{person}{Chen Li}, {and} \bibinfo{person}{Honggang Zhang}.} \bibinfo{year}{2024}\natexlab{}.
\newblock \bibinfo{title}{We-Math: Does Your Large Multimodal Model Achieve Human-like Mathematical Reasoning?}
\newblock
\newblock
\showeprint[arxiv]{2407.01284}


\bibitem[Qiao et~al\mbox{.}(2023)]%
        {qiao-etal-2023-reasoning}
\bibfield{author}{\bibinfo{person}{Shuofei Qiao}, \bibinfo{person}{Yixin Ou}, \bibinfo{person}{Ningyu Zhang}, \bibinfo{person}{Xiang Chen}, \bibinfo{person}{Yunzhi Yao}, \bibinfo{person}{Shumin Deng}, \bibinfo{person}{Chuanqi Tan}, \bibinfo{person}{Fei Huang}, {and} \bibinfo{person}{Huajun Chen}.} \bibinfo{year}{2023}\natexlab{}.
\newblock \showarticletitle{Reasoning with Language Model Prompting: A Survey}. In \bibinfo{booktitle}{\emph{ACL}}. \bibinfo{pages}{5368--5393}.
\newblock


\bibitem[Qin et~al\mbox{.}(2020)]%
        {qin2020semantically}
\bibfield{author}{\bibinfo{person}{Jinghui Qin}, \bibinfo{person}{Lihui Lin}, \bibinfo{person}{Xiaodan Liang}, \bibinfo{person}{Rumin Zhang}, {and} \bibinfo{person}{Liang Lin}.} \bibinfo{year}{2020}\natexlab{}.
\newblock \showarticletitle{Semantically-Aligned Universal Tree-Structured Solver for Math Word Problems}. In \bibinfo{booktitle}{\emph{EMNLP}}. \bibinfo{pages}{3780--3789}.
\newblock


\bibitem[Qiu et~al\mbox{.}(2020)]%
        {qiu2020pre}
\bibfield{author}{\bibinfo{person}{Xipeng Qiu}, \bibinfo{person}{Tianxiang Sun}, \bibinfo{person}{Yige Xu}, \bibinfo{person}{Yunfan Shao}, \bibinfo{person}{Ning Dai}, {and} \bibinfo{person}{Xuanjing Huang}.} \bibinfo{year}{2020}\natexlab{}.
\newblock \showarticletitle{Pre-trained models for natural language processing: A survey}.
\newblock \bibinfo{journal}{\emph{Science China Technological Sciences}} \bibinfo{volume}{63}, \bibinfo{number}{10} (\bibinfo{year}{2020}), \bibinfo{pages}{1872--1897}.
\newblock


\bibitem[Radford et~al\mbox{.}(2018)]%
        {radford_gpt-1_2018}
\bibfield{author}{\bibinfo{person}{Alec Radford}, \bibinfo{person}{Karthik Narasimhan}, \bibinfo{person}{Tim Salimans}, {and} \bibinfo{person}{Ilya Sutskever}.} \bibinfo{year}{2018}\natexlab{}.
\newblock \showarticletitle{{Improving} {Language} {Understanding} by {Generative} {Pre}-{Training}}.
\newblock  (\bibinfo{date}{June} \bibinfo{year}{2018}), \bibinfo{pages}{12}.
\newblock


\bibitem[Radford et~al\mbox{.}(2019)]%
        {radford_gpt-2_2019}
\bibfield{author}{\bibinfo{person}{Alec Radford}, \bibinfo{person}{Jeffrey Wu}, \bibinfo{person}{Rewon Child}, \bibinfo{person}{David Luan}, \bibinfo{person}{Dario Amodei}, {and} \bibinfo{person}{Ilya Sutskever}.} \bibinfo{year}{2019}\natexlab{}.
\newblock \showarticletitle{{Language} {Models} are {Unsupervised} {Multitask} {Learners}}.
\newblock  (\bibinfo{date}{Feb.} \bibinfo{year}{2019}), \bibinfo{pages}{24}.
\newblock


\bibitem[Rafailov et~al\mbox{.}(2024)]%
        {rafailov2024direct}
\bibfield{author}{\bibinfo{person}{Rafael Rafailov}, \bibinfo{person}{Archit Sharma}, \bibinfo{person}{Eric Mitchell}, \bibinfo{person}{Christopher~D Manning}, \bibinfo{person}{Stefano Ermon}, {and} \bibinfo{person}{Chelsea Finn}.} \bibinfo{year}{2024}\natexlab{}.
\newblock \showarticletitle{Direct preference optimization: Your language model is secretly a reward model}.
\newblock \bibinfo{journal}{\emph{NeurIPS}}  \bibinfo{volume}{36} (\bibinfo{year}{2024}).
\newblock


\bibitem[Rawte et~al\mbox{.}(2023)]%
        {rawte2023survey}
\bibfield{author}{\bibinfo{person}{Vipula Rawte}, \bibinfo{person}{Amit Sheth}, {and} \bibinfo{person}{Amitava Das}.} \bibinfo{year}{2023}\natexlab{}.
\newblock \showarticletitle{A survey of hallucination in large foundation models}.
\newblock \bibinfo{journal}{\emph{arXiv}} (\bibinfo{year}{2023}).
\newblock


\bibitem[Raﬀel et~al\mbox{.}({[n.\,d.]})]%
        {rael_exploring_nodate}
\bibfield{author}{\bibinfo{person}{Colin Raﬀel}, \bibinfo{person}{Noam Shazeer}, \bibinfo{person}{Adam Roberts}, \bibinfo{person}{Katherine Lee}, \bibinfo{person}{Sharan Narang}, \bibinfo{person}{Michael Matena}, \bibinfo{person}{Yanqi Zhou}, \bibinfo{person}{Wei Li}, {and} \bibinfo{person}{Peter~J Liu}.} \bibinfo{year}{[n.\,d.]}\natexlab{}.
\newblock \showarticletitle{Exploring the {Limits} of {Transfer} {Learning} with a {Uniﬁed} {Text}-to-{Text} {Transformer}}.
\newblock  (\bibinfo{year}{[n.\,d.]}).
\newblock


\bibitem[Roy and Roth(2015)]%
        {roy2015solving}
\bibfield{author}{\bibinfo{person}{Subhro Roy} {and} \bibinfo{person}{Dan Roth}.} \bibinfo{year}{2015}\natexlab{}.
\newblock \showarticletitle{Solving General Arithmetic Word Problems}. In \bibinfo{booktitle}{\emph{EMNLP}}. \bibinfo{pages}{1743--1752}.
\newblock


\bibitem[Roy and Roth(2017)]%
        {roy2017unit}
\bibfield{author}{\bibinfo{person}{Subhro Roy} {and} \bibinfo{person}{Dan Roth}.} \bibinfo{year}{2017}\natexlab{}.
\newblock \showarticletitle{Unit dependency graph and its application to arithmetic word problem solving}. In \bibinfo{booktitle}{\emph{AAAI}}, Vol.~\bibinfo{volume}{31}.
\newblock


\bibitem[Roy and Roth(2018)]%
        {roy2018mapping}
\bibfield{author}{\bibinfo{person}{Subhro Roy} {and} \bibinfo{person}{Dan Roth}.} \bibinfo{year}{2018}\natexlab{}.
\newblock \showarticletitle{Mapping to declarative knowledge for word problem solving}.
\newblock \bibinfo{journal}{\emph{TACL}}  \bibinfo{volume}{6} (\bibinfo{year}{2018}), \bibinfo{pages}{159--172}.
\newblock


\bibitem[Roy et~al\mbox{.}(2015)]%
        {roy2015reasoning}
\bibfield{author}{\bibinfo{person}{Subhro Roy}, \bibinfo{person}{Tim Vieira}, {and} \bibinfo{person}{Dan Roth}.} \bibinfo{year}{2015}\natexlab{}.
\newblock \showarticletitle{Reasoning about quantities in natural language}.
\newblock \bibinfo{journal}{\emph{TACL}}  \bibinfo{volume}{3} (\bibinfo{year}{2015}), \bibinfo{pages}{1--13}.
\newblock


\bibitem[Rudnicki(1992)]%
        {rudnicki1992overview}
\bibfield{author}{\bibinfo{person}{Piotr Rudnicki}.} \bibinfo{year}{1992}\natexlab{}.
\newblock \showarticletitle{An overview of the Mizar project}. In \bibinfo{booktitle}{\emph{Proceedings of the 1992 Workshop on Types for Proofs and Programs}}. \bibinfo{pages}{311--330}.
\newblock


\bibitem[Saxton et~al\mbox{.}(2019)]%
        {saxton2018analysing}
\bibfield{author}{\bibinfo{person}{David Saxton}, \bibinfo{person}{Edward Grefenstette}, \bibinfo{person}{Felix Hill}, {and} \bibinfo{person}{Pushmeet Kohli}.} \bibinfo{year}{2019}\natexlab{}.
\newblock \showarticletitle{Analysing Mathematical Reasoning Abilities of Neural Models}. In \bibinfo{booktitle}{\emph{ICLR}}.
\newblock


\bibitem[Scharpf et~al\mbox{.}(2022)]%
        {scharpf2022mining}
\bibfield{author}{\bibinfo{person}{Philipp Scharpf}, \bibinfo{person}{Moritz Schubotz}, {and} \bibinfo{person}{Bela Gipp}.} \bibinfo{year}{2022}\natexlab{}.
\newblock \showarticletitle{Mining mathematical documents for question answering via unsupervised formula labeling}. In \bibinfo{booktitle}{\emph{ICDL}}. \bibinfo{pages}{1--11}.
\newblock


\bibitem[Schick et~al\mbox{.}(2023)]%
        {schick2023toolformer}
\bibfield{author}{\bibinfo{person}{Timo Schick}, \bibinfo{person}{Jane Dwivedi-Yu}, \bibinfo{person}{Roberto Dessì}, \bibinfo{person}{Roberta Raileanu}, \bibinfo{person}{Maria Lomeli}, \bibinfo{person}{Luke Zettlemoyer}, \bibinfo{person}{Nicola Cancedda}, {and} \bibinfo{person}{Thomas Scialom}.} \bibinfo{year}{2023}\natexlab{}.
\newblock \bibinfo{title}{Toolformer: Language Models Can Teach Themselves to Use Tools}.
\newblock
\newblock
\showeprint[arxiv]{2302.04761}


\bibitem[Schulman et~al\mbox{.}(2017)]%
        {schulman2017proximal}
\bibfield{author}{\bibinfo{person}{John Schulman}, \bibinfo{person}{Filip Wolski}, \bibinfo{person}{Prafulla Dhariwal}, \bibinfo{person}{Alec Radford}, {and} \bibinfo{person}{Oleg Klimov}.} \bibinfo{year}{2017}\natexlab{}.
\newblock \showarticletitle{Proximal policy optimization algorithms}.
\newblock \bibinfo{journal}{\emph{arXiv}} (\bibinfo{year}{2017}).
\newblock


\bibitem[Sel et~al\mbox{.}(2023)]%
        {sel2023algorithm}
\bibfield{author}{\bibinfo{person}{Bilgehan Sel}, \bibinfo{person}{Ahmad Al-Tawaha}, \bibinfo{person}{Vanshaj Khattar}, \bibinfo{person}{Lu Wang}, \bibinfo{person}{Ruoxi Jia}, {and} \bibinfo{person}{Ming Jin}.} \bibinfo{year}{2023}\natexlab{}.
\newblock \showarticletitle{Algorithm of thoughts: Enhancing exploration of ideas in large language models}.
\newblock \bibinfo{journal}{\emph{arXiv}} (\bibinfo{year}{2023}).
\newblock


\bibitem[Seo et~al\mbox{.}(2015)]%
        {seo-etal-2015-solving}
\bibfield{author}{\bibinfo{person}{Minjoon Seo}, \bibinfo{person}{Hannaneh Hajishirzi}, \bibinfo{person}{Ali Farhadi}, \bibinfo{person}{Oren Etzioni}, {and} \bibinfo{person}{Clint Malcolm}.} \bibinfo{year}{2015}\natexlab{}.
\newblock \showarticletitle{Solving Geometry Problems: Combining Text and Diagram Interpretation}. In \bibinfo{booktitle}{\emph{EMNLP}}. \bibinfo{pages}{1466--1476}.
\newblock


\bibitem[Shen et~al\mbox{.}(2021)]%
        {shen_generate_2021}
\bibfield{author}{\bibinfo{person}{Jianhao Shen}, \bibinfo{person}{Yichun Yin}, \bibinfo{person}{Lin Li}, \bibinfo{person}{Lifeng Shang}, \bibinfo{person}{Xin Jiang}, \bibinfo{person}{Ming Zhang}, {and} \bibinfo{person}{Qun Liu}.} \bibinfo{year}{2021}\natexlab{}.
\newblock \bibinfo{title}{Generate \& {Rank}: {A} {Multi}-task {Framework} for {Math} {Word} {Problems}}.
\newblock
\newblock


\bibitem[Shi et~al\mbox{.}(2022)]%
        {shi2022language}
\bibfield{author}{\bibinfo{person}{Freda Shi}, \bibinfo{person}{Mirac Suzgun}, \bibinfo{person}{Markus Freitag}, \bibinfo{person}{Xuezhi Wang}, \bibinfo{person}{Suraj Srivats}, \bibinfo{person}{Soroush Vosoughi}, \bibinfo{person}{Hyung~Won Chung}, \bibinfo{person}{Yi Tay}, \bibinfo{person}{Sebastian Ruder}, \bibinfo{person}{Denny Zhou}, {et~al\mbox{.}}} \bibinfo{year}{2022}\natexlab{}.
\newblock \showarticletitle{Language models are multilingual chain-of-thought reasoners}. In \bibinfo{booktitle}{\emph{ICLR}}.
\newblock


\bibitem[Shi et~al\mbox{.}(2015)]%
        {shi2015automatically}
\bibfield{author}{\bibinfo{person}{Shuming Shi}, \bibinfo{person}{Yuehui Wang}, \bibinfo{person}{Chin-Yew Lin}, \bibinfo{person}{Xiaojiang Liu}, {and} \bibinfo{person}{Yong Rui}.} \bibinfo{year}{2015}\natexlab{}.
\newblock \showarticletitle{Automatically solving number word problems by semantic parsing and reasoning}. In \bibinfo{booktitle}{\emph{EMNLP}}. \bibinfo{pages}{1132--1142}.
\newblock


\bibitem[Shi et~al\mbox{.}(2024)]%
        {shihu2024mathllava}
\bibfield{author}{\bibinfo{person}{Wenhao Shi}, \bibinfo{person}{Zhiqiang Hu}, \bibinfo{person}{Yi Bin}, \bibinfo{person}{Junhua Liu}, \bibinfo{person}{Yang Yang}, \bibinfo{person}{See-Kiong Ng}, \bibinfo{person}{Lidong Bing}, {and} \bibinfo{person}{Roy Ka-Wei Lee}.} \bibinfo{year}{2024}\natexlab{}.
\newblock \bibinfo{title}{Math-LLaVA: Bootstrapping Mathematical Reasoning for Multimodal Large Language Models}.
\newblock
\newblock


\bibitem[Shinn et~al\mbox{.}(2023)]%
        {shinn2023reflexion}
\bibfield{author}{\bibinfo{person}{Noah Shinn}, \bibinfo{person}{Federico Cassano}, \bibinfo{person}{Ashwin Gopinath}, \bibinfo{person}{Karthik~R Narasimhan}, {and} \bibinfo{person}{Shunyu Yao}.} \bibinfo{year}{2023}\natexlab{}.
\newblock \showarticletitle{Reflexion: Language agents with verbal reinforcement learning}. In \bibinfo{booktitle}{\emph{NeurIPS}}.
\newblock


\bibitem[Shridhar et~al\mbox{.}(2023)]%
        {shridhar2023screws}
\bibfield{author}{\bibinfo{person}{Kumar Shridhar}, \bibinfo{person}{Harsh Jhamtani}, \bibinfo{person}{Hao Fang}, \bibinfo{person}{Benjamin Van~Durme}, \bibinfo{person}{Jason Eisner}, {and} \bibinfo{person}{Patrick Xia}.} \bibinfo{year}{2023}\natexlab{}.
\newblock \showarticletitle{SCREWS: A Modular Framework for Reasoning with Revisions}.
\newblock \bibinfo{journal}{\emph{arXiv}} (\bibinfo{year}{2023}).
\newblock


\bibitem[Shridhar et~al\mbox{.}(2022)]%
        {shridhar2022automatic}
\bibfield{author}{\bibinfo{person}{Kumar Shridhar}, \bibinfo{person}{Jakub Macina}, \bibinfo{person}{Mennatallah El-Assady}, \bibinfo{person}{Tanmay Sinha}, \bibinfo{person}{Manu Kapur}, {and} \bibinfo{person}{Mrinmaya Sachan}.} \bibinfo{year}{2022}\natexlab{}.
\newblock \showarticletitle{Automatic Generation of Socratic Subquestions for Teaching Math Word Problems}.
\newblock \bibinfo{journal}{\emph{arXiv}} (\bibinfo{year}{2022}).
\newblock


\bibitem[Shum et~al\mbox{.}(2023)]%
        {shum_automatic_2023}
\bibfield{author}{\bibinfo{person}{Kashun Shum}, \bibinfo{person}{Shizhe Diao}, {and} \bibinfo{person}{Tong Zhang}.} \bibinfo{year}{2023}\natexlab{}.
\newblock \showarticletitle{Automatic Prompt Augmentation and Selection with Chain-of-Thought from Labeled Data}. In \bibinfo{booktitle}{\emph{EMNLP}}. \bibinfo{pages}{12113--12139}.
\newblock


\bibitem[Slind and Norrish(2008)]%
        {slind2008brief}
\bibfield{author}{\bibinfo{person}{Konrad Slind} {and} \bibinfo{person}{Michael Norrish}.} \bibinfo{year}{2008}\natexlab{}.
\newblock \showarticletitle{A brief overview of HOL4}. In \bibinfo{booktitle}{\emph{TPHOLs}}. \bibinfo{pages}{28--32}.
\newblock


\bibitem[Song et~al\mbox{.}(2024)]%
        {song2024towards}
\bibfield{author}{\bibinfo{person}{Peiyang Song}, \bibinfo{person}{Kaiyu Yang}, {and} \bibinfo{person}{Anima Anandkumar}.} \bibinfo{year}{2024}\natexlab{}.
\newblock \showarticletitle{Towards large language models as copilots for theorem proving in lean}.
\newblock \bibinfo{journal}{\emph{arXiv}} (\bibinfo{year}{2024}).
\newblock


\bibitem[Spithourakis et~al\mbox{.}(2016)]%
        {spithourakis2016numerically}
\bibfield{author}{\bibinfo{person}{Georgios Spithourakis}, \bibinfo{person}{Isabelle Augenstein}, {and} \bibinfo{person}{Sebastian Riedel}.} \bibinfo{year}{2016}\natexlab{}.
\newblock \showarticletitle{Numerically Grounded Language Models for Semantic Error Correction}. In \bibinfo{booktitle}{\emph{EMNLP}}. \bibinfo{pages}{987--992}.
\newblock


\bibitem[Spithourakis and Riedel(2018)]%
        {spithourakis2018numeracy}
\bibfield{author}{\bibinfo{person}{GP Spithourakis} {and} \bibinfo{person}{S Riedel}.} \bibinfo{year}{2018}\natexlab{}.
\newblock \showarticletitle{Numeracy for language models: Evaluating and improving their ability to predict numbers}. In \bibinfo{booktitle}{\emph{ACL}}, Vol.~\bibinfo{volume}{56}. \bibinfo{pages}{2104--2115}.
\newblock


\bibitem[Sun et~al\mbox{.}(2023)]%
        {sun2023adaplanner}
\bibfield{author}{\bibinfo{person}{Haotian Sun}, \bibinfo{person}{Yuchen Zhuang}, \bibinfo{person}{Lingkai Kong}, \bibinfo{person}{Bo Dai}, {and} \bibinfo{person}{Chao Zhang}.} \bibinfo{year}{2023}\natexlab{}.
\newblock \showarticletitle{AdaPlanner: Adaptive Planning from Feedback with Language Models}.
\newblock \bibinfo{journal}{\emph{arXiv}} (\bibinfo{year}{2023}).
\newblock


\bibitem[Taylor et~al\mbox{.}(2022)]%
        {taylor_galactica_2022}
\bibfield{author}{\bibinfo{person}{Ross Taylor}, \bibinfo{person}{Marcin Kardas}, \bibinfo{person}{Guillem Cucurull}, \bibinfo{person}{Thomas Scialom}, \bibinfo{person}{Anthony Hartshorn}, \bibinfo{person}{Elvis Saravia}, \bibinfo{person}{Andrew Poulton}, \bibinfo{person}{Viktor Kerkez}, {and} \bibinfo{person}{Robert Stojnic}.} \bibinfo{year}{2022}\natexlab{}.
\newblock \bibinfo{title}{Galactica: {A} {Large} {Language} {Model} for {Science}}.
\newblock
\newblock


\bibitem[Team et~al\mbox{.}(2023)]%
        {team2023gemini}
\bibfield{author}{\bibinfo{person}{Gemini Team}, \bibinfo{person}{Rohan Anil}, \bibinfo{person}{Sebastian Borgeaud}, \bibinfo{person}{Jean-Baptiste Alayrac}, \bibinfo{person}{Jiahui Yu}, \bibinfo{person}{Radu Soricut}, \bibinfo{person}{Johan Schalkwyk}, \bibinfo{person}{Andrew~M Dai}, \bibinfo{person}{Anja Hauth}, \bibinfo{person}{Katie Millican}, {et~al\mbox{.}}} \bibinfo{year}{2023}\natexlab{}.
\newblock \showarticletitle{Gemini: a family of highly capable multimodal models}.
\newblock \bibinfo{journal}{\emph{arXiv}} (\bibinfo{year}{2023}).
\newblock


\bibitem[Team(2024)]%
        {qwq-32b-preview}
\bibfield{author}{\bibinfo{person}{Qwen Team}.} \bibinfo{year}{2024}\natexlab{}.
\newblock \bibinfo{title}{QwQ: Reflect Deeply on the Boundaries of the Unknown}.
\newblock
\newblock
\urldef\tempurl%
\url{https://qwenlm.github.io/blog/qwq-32b-preview/}
\showURL{%
\tempurl}


\bibitem[Thoppilan et~al\mbox{.}(2022)]%
        {thoppilan2022lamda}
\bibfield{author}{\bibinfo{person}{Romal Thoppilan}, \bibinfo{person}{Daniel De~Freitas}, \bibinfo{person}{Jamie Hall}, \bibinfo{person}{Noam Shazeer}, \bibinfo{person}{Apoorv Kulshreshtha}, \bibinfo{person}{Heng-Tze Cheng}, \bibinfo{person}{Alicia Jin}, \bibinfo{person}{Taylor Bos}, \bibinfo{person}{Leslie Baker}, \bibinfo{person}{Yu Du}, {et~al\mbox{.}}} \bibinfo{year}{2022}\natexlab{}.
\newblock \showarticletitle{Lamda: Language models for dialog applications}.
\newblock \bibinfo{journal}{\emph{arXiv}} (\bibinfo{year}{2022}).
\newblock


\bibitem[Touvron et~al\mbox{.}(2023a)]%
        {touvron2023llama}
\bibfield{author}{\bibinfo{person}{Hugo Touvron}, \bibinfo{person}{Thibaut Lavril}, \bibinfo{person}{Gautier Izacard}, \bibinfo{person}{Xavier Martinet}, \bibinfo{person}{Marie-Anne Lachaux}, \bibinfo{person}{Timoth{\'e}e Lacroix}, \bibinfo{person}{Baptiste Rozi{\`e}re}, \bibinfo{person}{Naman Goyal}, \bibinfo{person}{Eric Hambro}, \bibinfo{person}{Faisal Azhar}, {et~al\mbox{.}}} \bibinfo{year}{2023}\natexlab{a}.
\newblock \showarticletitle{Llama: Open and efficient foundation language models}.
\newblock \bibinfo{journal}{\emph{arXiv}} (\bibinfo{year}{2023}).
\newblock


\bibitem[Touvron et~al\mbox{.}(2023b)]%
        {touvron2023llama2}
\bibfield{author}{\bibinfo{person}{Hugo Touvron}, \bibinfo{person}{Louis Martin}, \bibinfo{person}{Kevin Stone}, \bibinfo{person}{Peter Albert}, \bibinfo{person}{Amjad Almahairi}, \bibinfo{person}{Yasmine Babaei}, \bibinfo{person}{Nikolay Bashlykov}, \bibinfo{person}{Soumya Batra}, \bibinfo{person}{Prajjwal Bhargava}, \bibinfo{person}{Shruti Bhosale}, {et~al\mbox{.}}} \bibinfo{year}{2023}\natexlab{b}.
\newblock \showarticletitle{Llama 2: Open foundation and fine-tuned chat models}.
\newblock \bibinfo{journal}{\emph{arXiv}} (\bibinfo{year}{2023}).
\newblock


\bibitem[Trung et~al\mbox{.}(2024)]%
        {trung2024reft}
\bibfield{author}{\bibinfo{person}{Luong Trung}, \bibinfo{person}{Xinbo Zhang}, \bibinfo{person}{Zhanming Jie}, \bibinfo{person}{Peng Sun}, \bibinfo{person}{Xiaoran Jin}, {and} \bibinfo{person}{Hang Li}.} \bibinfo{year}{2024}\natexlab{}.
\newblock \showarticletitle{Reft: Reasoning with reinforced fine-tuning}. In \bibinfo{booktitle}{\emph{ACL}}. \bibinfo{pages}{7601--7614}.
\newblock


\bibitem[Upadhyay and Chang(2015)]%
        {upadhyay2015draw}
\bibfield{author}{\bibinfo{person}{Shyam Upadhyay} {and} \bibinfo{person}{Ming-Wei Chang}.} \bibinfo{year}{2015}\natexlab{}.
\newblock \bibinfo{booktitle}{\emph{Draw: A challenging and diverse algebra word problem set}}.
\newblock \bibinfo{type}{{T}echnical {R}eport}. \bibinfo{institution}{Citeseer}.
\newblock


\bibitem[Upadhyay and Chang(2017)]%
        {upadhyay2017annotating}
\bibfield{author}{\bibinfo{person}{Shyam Upadhyay} {and} \bibinfo{person}{Ming-Wei Chang}.} \bibinfo{year}{2017}\natexlab{}.
\newblock \showarticletitle{Annotating Derivations: A New Evaluation Strategy and Dataset for Algebra Word Problems}. In \bibinfo{booktitle}{\emph{EACL}}. \bibinfo{pages}{494--504}.
\newblock


\bibitem[Upadhyay et~al\mbox{.}(2023)]%
        {upadhyay-etal-2023-improving}
\bibfield{author}{\bibinfo{person}{Shriyash Upadhyay}, \bibinfo{person}{Etan Ginsberg}, {and} \bibinfo{person}{Chris Callison-Burch}.} \bibinfo{year}{2023}\natexlab{}.
\newblock \showarticletitle{Improving Mathematics Tutoring With A Code Scratchpad}. In \bibinfo{booktitle}{\emph{BEA}}. \bibinfo{pages}{20--28}.
\newblock


\bibitem[Vaswani et~al\mbox{.}(2017)]%
        {vaswani2017attention}
\bibfield{author}{\bibinfo{person}{Ashish Vaswani}, \bibinfo{person}{Noam Shazeer}, \bibinfo{person}{Niki Parmar}, \bibinfo{person}{Jakob Uszkoreit}, \bibinfo{person}{Llion Jones}, \bibinfo{person}{Aidan~N Gomez}, \bibinfo{person}{{\L}ukasz Kaiser}, {and} \bibinfo{person}{Illia Polosukhin}.} \bibinfo{year}{2017}\natexlab{}.
\newblock \showarticletitle{Attention is all you need}.
\newblock \bibinfo{journal}{\emph{NeurIPS}}  \bibinfo{volume}{30} (\bibinfo{year}{2017}).
\newblock


\bibitem[Wallace et~al\mbox{.}(2019)]%
        {wallace2019nlp}
\bibfield{author}{\bibinfo{person}{Eric Wallace}, \bibinfo{person}{Yizhong Wang}, \bibinfo{person}{Sujian Li}, \bibinfo{person}{Sameer Singh}, {and} \bibinfo{person}{Matt Gardner}.} \bibinfo{year}{2019}\natexlab{}.
\newblock \showarticletitle{Do NLP Models Know Numbers? Probing Numeracy in Embeddings}. In \bibinfo{booktitle}{\emph{EMNLP}}. \bibinfo{pages}{5307--5315}.
\newblock


\bibitem[Wang and Komatsuzaki(2021)]%
        {wang2021gpt}
\bibfield{author}{\bibinfo{person}{Ben Wang} {and} \bibinfo{person}{Aran Komatsuzaki}.} \bibinfo{year}{2021}\natexlab{}.
\newblock \bibinfo{title}{GPT-J-6B: A 6 billion parameter autoregressive language model}.
\newblock
\newblock


\bibitem[Wang et~al\mbox{.}(2021)]%
        {wang2021exploring}
\bibfield{author}{\bibinfo{person}{Cunxiang Wang}, \bibinfo{person}{Boyuan Zheng}, \bibinfo{person}{Yuchen Niu}, {and} \bibinfo{person}{Yue Zhang}.} \bibinfo{year}{2021}\natexlab{}.
\newblock \showarticletitle{Exploring Generalization Ability of Pretrained Language Models on Arithmetic and Logical Reasoning}. In \bibinfo{booktitle}{\emph{NLPCC}}. \bibinfo{pages}{758--769}.
\newblock


\bibitem[Wang et~al\mbox{.}(2024b)]%
        {wang2024measuring}
\bibfield{author}{\bibinfo{person}{Ke Wang}, \bibinfo{person}{Junting Pan}, \bibinfo{person}{Weikang Shi}, \bibinfo{person}{Zimu Lu}, \bibinfo{person}{Mingjie Zhan}, {and} \bibinfo{person}{Hongsheng Li}.} \bibinfo{year}{2024}\natexlab{b}.
\newblock \showarticletitle{Measuring multimodal mathematical reasoning with math-vision dataset}.
\newblock \bibinfo{journal}{\emph{arXiv}} (\bibinfo{year}{2024}).
\newblock


\bibitem[Wang et~al\mbox{.}(2024a)]%
        {Qwen2-VL}
\bibfield{author}{\bibinfo{person}{Peng Wang}, \bibinfo{person}{Shuai Bai}, \bibinfo{person}{Sinan Tan}, \bibinfo{person}{Shijie Wang}, \bibinfo{person}{Zhihao Fan}, \bibinfo{person}{Jinze Bai}, \bibinfo{person}{Keqin Chen}, \bibinfo{person}{Xuejing Liu}, \bibinfo{person}{Jialin Wang}, \bibinfo{person}{Wenbin Ge}, \bibinfo{person}{Yang Fan}, \bibinfo{person}{Kai Dang}, \bibinfo{person}{Mengfei Du}, \bibinfo{person}{Xuancheng Ren}, \bibinfo{person}{Rui Men}, \bibinfo{person}{Dayiheng Liu}, \bibinfo{person}{Chang Zhou}, \bibinfo{person}{Jingren Zhou}, {and} \bibinfo{person}{Junyang Lin}.} \bibinfo{year}{2024}\natexlab{a}.
\newblock \showarticletitle{Qwen2-VL: Enhancing Vision-Language Model's Perception of the World at Any Resolution}.
\newblock \bibinfo{journal}{\emph{arXiv}} (\bibinfo{year}{2024}).
\newblock


\bibitem[Wang et~al\mbox{.}(2023a)]%
        {wang2023making}
\bibfield{author}{\bibinfo{person}{Peiyi Wang}, \bibinfo{person}{Lei Li}, \bibinfo{person}{Liang Chen}, \bibinfo{person}{Feifan Song}, \bibinfo{person}{Binghuai Lin}, \bibinfo{person}{Yunbo Cao}, \bibinfo{person}{Tianyu Liu}, {and} \bibinfo{person}{Zhifang Sui}.} \bibinfo{year}{2023}\natexlab{a}.
\newblock \showarticletitle{Making large language models better reasoners with alignment}.
\newblock \bibinfo{journal}{\emph{arXiv}} (\bibinfo{year}{2023}).
\newblock


\bibitem[Wang et~al\mbox{.}(2023b)]%
        {wang2023cogvlm}
\bibfield{author}{\bibinfo{person}{Weihan Wang}, \bibinfo{person}{Qingsong Lv}, \bibinfo{person}{Wenmeng Yu}, \bibinfo{person}{Wenyi Hong}, \bibinfo{person}{Ji Qi}, \bibinfo{person}{Yan Wang}, \bibinfo{person}{Junhui Ji}, \bibinfo{person}{Zhuoyi Yang}, \bibinfo{person}{Lei Zhao}, \bibinfo{person}{Xixuan Song}, \bibinfo{person}{Jiazheng Xu}, \bibinfo{person}{Bin Xu}, \bibinfo{person}{Juanzi Li}, \bibinfo{person}{Yuxiao Dong}, \bibinfo{person}{Ming Ding}, {and} \bibinfo{person}{Jie Tang}.} \bibinfo{year}{2023}\natexlab{b}.
\newblock \bibinfo{title}{CogVLM: Visual Expert for Pretrained Language Models}.
\newblock
\newblock
\showeprint[arxiv]{2311.03079}


\bibitem[Wang et~al\mbox{.}(2023c)]%
        {wang2023mint}
\bibfield{author}{\bibinfo{person}{Xingyao Wang}, \bibinfo{person}{Zihan Wang}, \bibinfo{person}{Jiateng Liu}, \bibinfo{person}{Yangyi Chen}, \bibinfo{person}{Lifan Yuan}, \bibinfo{person}{Hao Peng}, {and} \bibinfo{person}{Heng Ji}.} \bibinfo{year}{2023}\natexlab{c}.
\newblock \showarticletitle{Mint: Evaluating llms in multi-turn interaction with tools and language feedback}.
\newblock \bibinfo{journal}{\emph{arXiv}} (\bibinfo{year}{2023}).
\newblock


\bibitem[Wang et~al\mbox{.}(2023d)]%
        {wang2022self}
\bibfield{author}{\bibinfo{person}{Xuezhi Wang}, \bibinfo{person}{Jason Wei}, \bibinfo{person}{Dale Schuurmans}, \bibinfo{person}{Quoc~V Le}, \bibinfo{person}{Ed~H Chi}, \bibinfo{person}{Sharan Narang}, \bibinfo{person}{Aakanksha Chowdhery}, {and} \bibinfo{person}{Denny Zhou}.} \bibinfo{year}{2023}\natexlab{d}.
\newblock \showarticletitle{Self-Consistency Improves Chain of Thought Reasoning in Language Models}. In \bibinfo{booktitle}{\emph{ICLR}}.
\newblock


\bibitem[wang and Fu(2024)]%
        {wang2024understanding}
\bibfield{author}{\bibinfo{person}{Yifan wang} {and} \bibinfo{person}{Yun Fu}.} \bibinfo{year}{2024}\natexlab{}.
\newblock \bibinfo{title}{Understanding, Abstracting and Checking: Evoking Complicated Multimodal Reasoning in {LMM}s}.
\newblock
\newblock


\bibitem[Wang et~al\mbox{.}(2017)]%
        {wang2017deep}
\bibfield{author}{\bibinfo{person}{Yan Wang}, \bibinfo{person}{Xiaojiang Liu}, {and} \bibinfo{person}{Shuming Shi}.} \bibinfo{year}{2017}\natexlab{}.
\newblock \showarticletitle{Deep neural solver for math word problems}. In \bibinfo{booktitle}{\emph{EMNLP}}. \bibinfo{pages}{845--854}.
\newblock


\bibitem[Wang et~al\mbox{.}(2023e)]%
        {wang2023generative}
\bibfield{author}{\bibinfo{person}{Zengzhi Wang}, \bibinfo{person}{Rui Xia}, {and} \bibinfo{person}{Pengfei Liu}.} \bibinfo{year}{2023}\natexlab{e}.
\newblock \showarticletitle{Generative AI for Math: Part I--MathPile: A Billion-Token-Scale Pretraining Corpus for Math}.
\newblock \bibinfo{journal}{\emph{arXiv}} (\bibinfo{year}{2023}).
\newblock


\bibitem[Wei et~al\mbox{.}(2022)]%
        {wei2022chain}
\bibfield{author}{\bibinfo{person}{Jason Wei}, \bibinfo{person}{Xuezhi Wang}, \bibinfo{person}{Dale Schuurmans}, \bibinfo{person}{Maarten Bosma}, \bibinfo{person}{Fei Xia}, \bibinfo{person}{Ed Chi}, \bibinfo{person}{Quoc~V Le}, \bibinfo{person}{Denny Zhou}, {et~al\mbox{.}}} \bibinfo{year}{2022}\natexlab{}.
\newblock \showarticletitle{Chain-of-thought prompting elicits reasoning in large language models}.
\newblock \bibinfo{journal}{\emph{NeurIPS}}  \bibinfo{volume}{35} (\bibinfo{year}{2022}), \bibinfo{pages}{24824--24837}.
\newblock


\bibitem[Welleck et~al\mbox{.}(2021)]%
        {welleck2021naturalproofs}
\bibfield{author}{\bibinfo{person}{Sean Welleck}, \bibinfo{person}{Jiacheng Liu}, \bibinfo{person}{Ronan Le~Bras}, \bibinfo{person}{Hannaneh Hajishirzi}, \bibinfo{person}{Yejin Choi}, {and} \bibinfo{person}{Kyunghyun Cho}.} \bibinfo{year}{2021}\natexlab{}.
\newblock \showarticletitle{NaturalProofs: Mathematical Theorem Proving in Natural Language}. In \bibinfo{booktitle}{\emph{NeurIPS Datasets and Benchmarks Track (Round 1)}}.
\newblock


\bibitem[Welleck et~al\mbox{.}(2022)]%
        {welleck2022naturalprover}
\bibfield{author}{\bibinfo{person}{Sean Welleck}, \bibinfo{person}{Jiacheng Liu}, \bibinfo{person}{Ximing Lu}, \bibinfo{person}{Hannaneh Hajishirzi}, {and} \bibinfo{person}{Yejin Choi}.} \bibinfo{year}{2022}\natexlab{}.
\newblock \showarticletitle{Naturalprover: Grounded mathematical proof generation with language models}.
\newblock \bibinfo{journal}{\emph{NeurIPS}}  \bibinfo{volume}{35} (\bibinfo{year}{2022}), \bibinfo{pages}{4913--4927}.
\newblock


\bibitem[Welleck and Saha(2023)]%
        {llmstep}
\bibfield{author}{\bibinfo{person}{Sean Welleck} {and} \bibinfo{person}{Rahul Saha}.} \bibinfo{year}{2023}\natexlab{}.
\newblock \showarticletitle{llmstep: LLM proofstep suggestions in Lean}. In \bibinfo{booktitle}{\emph{NeurIPS}}.
\newblock


\bibitem[Wenzel et~al\mbox{.}(2008)]%
        {wenzel2008isabelle}
\bibfield{author}{\bibinfo{person}{Makarius Wenzel}, \bibinfo{person}{Lawrence~C Paulson}, {and} \bibinfo{person}{Tobias Nipkow}.} \bibinfo{year}{2008}\natexlab{}.
\newblock \showarticletitle{The isabelle framework}. In \bibinfo{booktitle}{\emph{TPHOLs}}. \bibinfo{pages}{33--38}.
\newblock


\bibitem[Wu et~al\mbox{.}(2021)]%
        {wu2020int}
\bibfield{author}{\bibinfo{person}{Yuhuai Wu}, \bibinfo{person}{Albert Jiang}, \bibinfo{person}{Jimmy Ba}, {and} \bibinfo{person}{Roger~Baker Grosse}.} \bibinfo{year}{2021}\natexlab{}.
\newblock \showarticletitle{INT: An Inequality Benchmark for Evaluating Generalization in Theorem Proving}. In \bibinfo{booktitle}{\emph{ICLR}}.
\newblock


\bibitem[Wu et~al\mbox{.}(2022)]%
        {DBLP:conf/nips/WuJLRSJS22}
\bibfield{author}{\bibinfo{person}{Yuhuai Wu}, \bibinfo{person}{Albert~Qiaochu Jiang}, \bibinfo{person}{Wenda Li}, \bibinfo{person}{Markus~N. Rabe}, \bibinfo{person}{Charles Staats}, \bibinfo{person}{Mateja Jamnik}, {and} \bibinfo{person}{Christian Szegedy}.} \bibinfo{year}{2022}\natexlab{}.
\newblock \showarticletitle{Autoformalization with Large Language Models}. In \bibinfo{booktitle}{\emph{NeurIPS}}.
\newblock


\bibitem[Wu et~al\mbox{.}({[n.\,d.]})]%
        {wu_lime_nodate}
\bibfield{author}{\bibinfo{person}{Yuhuai Wu}, \bibinfo{person}{Markus Rabe}, {and} \bibinfo{person}{Wenda Li}.} \bibinfo{year}{[n.\,d.]}\natexlab{}.
\newblock \showarticletitle{{LIME}: {Learning} {Inductive} {Bias} for {Primitives} of {Mathematical} {Reasoning}}.
\newblock  (\bibinfo{year}{[n.\,d.]}).
\newblock


\bibitem[Xia et~al\mbox{.}(2024)]%
        {xia2024evaluating}
\bibfield{author}{\bibinfo{person}{Shijie Xia}, \bibinfo{person}{Xuefeng Li}, \bibinfo{person}{Yixin Liu}, \bibinfo{person}{Tongshuang Wu}, {and} \bibinfo{person}{Pengfei Liu}.} \bibinfo{year}{2024}\natexlab{}.
\newblock \showarticletitle{Evaluating Mathematical Reasoning Beyond Accuracy}.
\newblock \bibinfo{journal}{\emph{arXiv}} (\bibinfo{year}{2024}).
\newblock


\bibitem[Xiang et~al\mbox{.}(2024)]%
        {xiang2024atomthink}
\bibfield{author}{\bibinfo{person}{Kun Xiang}, \bibinfo{person}{Zhili Liu}, \bibinfo{person}{Zihao Jiang}, \bibinfo{person}{Yunshuang Nie}, \bibinfo{person}{Runhui Huang}, \bibinfo{person}{Haoxiang Fan}, \bibinfo{person}{Hanhui Li}, \bibinfo{person}{Weiran Huang}, \bibinfo{person}{Yihan Zeng}, \bibinfo{person}{Jianhua Han}, {et~al\mbox{.}}} \bibinfo{year}{2024}\natexlab{}.
\newblock \showarticletitle{AtomThink: A Slow Thinking Framework for Multimodal Mathematical Reasoning}.
\newblock \bibinfo{journal}{\emph{arXiv}} (\bibinfo{year}{2024}).
\newblock


\bibitem[Xu et~al\mbox{.}(2023)]%
        {xu_wizardlm_2023}
\bibfield{author}{\bibinfo{person}{Can Xu}, \bibinfo{person}{Qingfeng Sun}, \bibinfo{person}{Kai Zheng}, \bibinfo{person}{Xiubo Geng}, \bibinfo{person}{Pu Zhao}, \bibinfo{person}{Jiazhan Feng}, \bibinfo{person}{Chongyang Tao}, {and} \bibinfo{person}{Daxin Jiang}.} \bibinfo{year}{2023}\natexlab{}.
\newblock \showarticletitle{Wizardlm: Empowering large language models to follow complex instructions}.
\newblock \bibinfo{journal}{\emph{arXiv}} (\bibinfo{year}{2023}).
\newblock


\bibitem[Xu et~al\mbox{.}(2024b)]%
        {xu2024llava}
\bibfield{author}{\bibinfo{person}{Guowei Xu}, \bibinfo{person}{Peng Jin}, \bibinfo{person}{Li Hao}, \bibinfo{person}{Yibing Song}, \bibinfo{person}{Lichao Sun}, {and} \bibinfo{person}{Li Yuan}.} \bibinfo{year}{2024}\natexlab{b}.
\newblock \showarticletitle{LLaVA-o1: Let Vision Language Models Reason Step-by-Step}.
\newblock \bibinfo{journal}{\emph{arXiv}} (\bibinfo{year}{2024}).
\newblock


\bibitem[Xu et~al\mbox{.}(2024a)]%
        {xu2024faithful}
\bibfield{author}{\bibinfo{person}{Jundong Xu}, \bibinfo{person}{Hao Fei}, \bibinfo{person}{Liangming Pan}, \bibinfo{person}{Qian Liu}, \bibinfo{person}{Mong-Li Lee}, {and} \bibinfo{person}{Wynne Hsu}.} \bibinfo{year}{2024}\natexlab{a}.
\newblock \showarticletitle{Faithful Logical Reasoning via Symbolic Chain-of-Thought}.
\newblock \bibinfo{journal}{\emph{arXiv}} (\bibinfo{year}{2024}).
\newblock


\bibitem[Yan et~al\mbox{.}(2024)]%
        {yan2024survey}
\bibfield{author}{\bibinfo{person}{Yibo Yan}, \bibinfo{person}{Jiamin Su}, \bibinfo{person}{Jianxiang He}, \bibinfo{person}{Fangteng Fu}, \bibinfo{person}{Xu Zheng}, \bibinfo{person}{Yuanhuiyi Lyu}, \bibinfo{person}{Kun Wang}, \bibinfo{person}{Shen Wang}, \bibinfo{person}{Qingsong Wen}, {and} \bibinfo{person}{Xuming Hu}.} \bibinfo{year}{2024}\natexlab{}.
\newblock \showarticletitle{A Survey of Mathematical Reasoning in the Era of Multimodal Large Language Model: Benchmark, Method \& Challenges}.
\newblock \bibinfo{journal}{\emph{arXiv}} (\bibinfo{year}{2024}).
\newblock


\bibitem[Yang et~al\mbox{.}(2024)]%
        {yang2024qwen2}
\bibfield{author}{\bibinfo{person}{An Yang}, \bibinfo{person}{Beichen Zhang}, \bibinfo{person}{Binyuan Hui}, \bibinfo{person}{Bofei Gao}, \bibinfo{person}{Bowen Yu}, \bibinfo{person}{Chengpeng Li}, \bibinfo{person}{Dayiheng Liu}, \bibinfo{person}{Jianhong Tu}, \bibinfo{person}{Jingren Zhou}, \bibinfo{person}{Junyang Lin}, {et~al\mbox{.}}} \bibinfo{year}{2024}\natexlab{}.
\newblock \showarticletitle{Qwen2. 5-math technical report: Toward mathematical expert model via self-improvement}.
\newblock \bibinfo{journal}{\emph{arXiv}} (\bibinfo{year}{2024}).
\newblock


\bibitem[Yang and Deng(2019)]%
        {yang2019learning}
\bibfield{author}{\bibinfo{person}{Kaiyu Yang} {and} \bibinfo{person}{Jia Deng}.} \bibinfo{year}{2019}\natexlab{}.
\newblock \showarticletitle{Learning to prove theorems via interacting with proof assistants}. In \bibinfo{booktitle}{\emph{ICML}}. \bibinfo{pages}{6984--6994}.
\newblock


\bibitem[Yang et~al\mbox{.}(2023)]%
        {yang2023gpt}
\bibfield{author}{\bibinfo{person}{Zhen Yang}, \bibinfo{person}{Ming Ding}, \bibinfo{person}{Qingsong Lv}, \bibinfo{person}{Zhihuan Jiang}, \bibinfo{person}{Zehai He}, \bibinfo{person}{Yuyi Guo}, \bibinfo{person}{Jinfeng Bai}, {and} \bibinfo{person}{Jie Tang}.} \bibinfo{year}{2023}\natexlab{}.
\newblock \showarticletitle{GPT Can Solve Mathematical Problems Without a Calculator}.
\newblock \bibinfo{journal}{\emph{arXiv}} (\bibinfo{year}{2023}).
\newblock


\bibitem[Yang et~al\mbox{.}(2022)]%
        {yang2022logicsolver}
\bibfield{author}{\bibinfo{person}{Zhicheng Yang}, \bibinfo{person}{Jinghui Qin}, \bibinfo{person}{Jiaqi Chen}, \bibinfo{person}{Liang Lin}, {and} \bibinfo{person}{Xiaodan Liang}.} \bibinfo{year}{2022}\natexlab{}.
\newblock \showarticletitle{Logicsolver: Towards interpretable math word problem solving with logical prompt-enhanced learning}.
\newblock \bibinfo{journal}{\emph{arXiv}} (\bibinfo{year}{2022}).
\newblock


\bibitem[Yao et~al\mbox{.}(2023b)]%
        {yao2023tree}
\bibfield{author}{\bibinfo{person}{Shunyu Yao}, \bibinfo{person}{Dian Yu}, \bibinfo{person}{Jeffrey Zhao}, \bibinfo{person}{Izhak Shafran}, \bibinfo{person}{Thomas~L Griffiths}, \bibinfo{person}{Yuan Cao}, {and} \bibinfo{person}{Karthik Narasimhan}.} \bibinfo{year}{2023}\natexlab{b}.
\newblock \showarticletitle{Tree of thoughts: Deliberate problem solving with large language models}.
\newblock \bibinfo{journal}{\emph{arXiv}} (\bibinfo{year}{2023}).
\newblock


\bibitem[Yao et~al\mbox{.}(2023a)]%
        {yao2023beyond}
\bibfield{author}{\bibinfo{person}{Yao Yao}, \bibinfo{person}{Zuchao Li}, {and} \bibinfo{person}{Hai Zhao}.} \bibinfo{year}{2023}\natexlab{a}.
\newblock \showarticletitle{Beyond Chain-of-Thought, Effective Graph-of-Thought Reasoning in Large Language Models}.
\newblock \bibinfo{journal}{\emph{arXiv}} (\bibinfo{year}{2023}).
\newblock


\bibitem[Yin et~al\mbox{.}(2024)]%
        {yin2024mumath}
\bibfield{author}{\bibinfo{person}{Shuo Yin}, \bibinfo{person}{Weihao You}, \bibinfo{person}{Zhilong Ji}, \bibinfo{person}{Guoqiang Zhong}, {and} \bibinfo{person}{Jinfeng Bai}.} \bibinfo{year}{2024}\natexlab{}.
\newblock \showarticletitle{MuMath-Code: Combining Tool-Use Large Language Models with Multi-perspective Data Augmentation for Mathematical Reasoning}.
\newblock \bibinfo{journal}{\emph{arXiv}} (\bibinfo{year}{2024}).
\newblock


\bibitem[Ying et~al\mbox{.}(2024)]%
        {ying2024internlmmath}
\bibfield{author}{\bibinfo{person}{Huaiyuan Ying}, \bibinfo{person}{Shuo Zhang}, \bibinfo{person}{Linyang Li}, \bibinfo{person}{Zhejian Zhou}, \bibinfo{person}{Yunfan Shao}, \bibinfo{person}{Zhaoye Fei}, \bibinfo{person}{Yichuan Ma}, \bibinfo{person}{Jiawei Hong}, \bibinfo{person}{Kuikun Liu}, \bibinfo{person}{Ziyi Wang}, \bibinfo{person}{Yudong Wang}, \bibinfo{person}{Zijian Wu}, \bibinfo{person}{Shuaibin Li}, \bibinfo{person}{Fengzhe Zhou}, \bibinfo{person}{Hongwei Liu}, \bibinfo{person}{Songyang Zhang}, \bibinfo{person}{Wenwei Zhang}, \bibinfo{person}{Hang Yan}, \bibinfo{person}{Xipeng Qiu}, \bibinfo{person}{Jiayu Wang}, \bibinfo{person}{Kai Chen}, {and} \bibinfo{person}{Dahua Lin}.} \bibinfo{year}{2024}\natexlab{}.
\newblock \bibinfo{title}{InternLM-Math: Open Math Large Language Models Toward Verifiable Reasoning}.
\newblock
\newblock


\bibitem[Yoran et~al\mbox{.}(2023)]%
        {yoran2023answering}
\bibfield{author}{\bibinfo{person}{Ori Yoran}, \bibinfo{person}{Tomer Wolfson}, \bibinfo{person}{Ben Bogin}, \bibinfo{person}{Uri Katz}, \bibinfo{person}{Daniel Deutch}, {and} \bibinfo{person}{Jonathan Berant}.} \bibinfo{year}{2023}\natexlab{}.
\newblock \showarticletitle{Answering questions by meta-reasoning over multiple chains of thought}.
\newblock \bibinfo{journal}{\emph{arXiv}} (\bibinfo{year}{2023}).
\newblock


\bibitem[Yu et~al\mbox{.}(2023a)]%
        {yu2023thought}
\bibfield{author}{\bibinfo{person}{Junchi Yu}, \bibinfo{person}{Ran He}, {and} \bibinfo{person}{Rex Ying}.} \bibinfo{year}{2023}\natexlab{a}.
\newblock \showarticletitle{Thought Propagation: An Analogical Approach to Complex Reasoning with Large Language Models}.
\newblock \bibinfo{journal}{\emph{arXiv}} (\bibinfo{year}{2023}).
\newblock


\bibitem[Yu et~al\mbox{.}(2023b)]%
        {yu2023metamath}
\bibfield{author}{\bibinfo{person}{Longhui Yu}, \bibinfo{person}{Weisen Jiang}, \bibinfo{person}{Han Shi}, \bibinfo{person}{Jincheng Yu}, \bibinfo{person}{Zhengying Liu}, \bibinfo{person}{Yu Zhang}, \bibinfo{person}{James~T Kwok}, \bibinfo{person}{Zhenguo Li}, \bibinfo{person}{Adrian Weller}, {and} \bibinfo{person}{Weiyang Liu}.} \bibinfo{year}{2023}\natexlab{b}.
\newblock \showarticletitle{Metamath: Bootstrap your own mathematical questions for large language models}.
\newblock \bibinfo{journal}{\emph{arXiv}} (\bibinfo{year}{2023}).
\newblock


\bibitem[Yuan et~al\mbox{.}(2024)]%
        {yuan2024advancing}
\bibfield{author}{\bibinfo{person}{Lifan Yuan}, \bibinfo{person}{Ganqu Cui}, \bibinfo{person}{Hanbin Wang}, \bibinfo{person}{Ning Ding}, \bibinfo{person}{Xingyao Wang}, \bibinfo{person}{Jia Deng}, \bibinfo{person}{Boji Shan}, \bibinfo{person}{Huimin Chen}, \bibinfo{person}{Ruobing Xie}, \bibinfo{person}{Yankai Lin}, {et~al\mbox{.}}} \bibinfo{year}{2024}\natexlab{}.
\newblock \showarticletitle{Advancing llm reasoning generalists with preference trees}.
\newblock \bibinfo{journal}{\emph{arXiv}} (\bibinfo{year}{2024}).
\newblock


\bibitem[Yuan et~al\mbox{.}(2023a)]%
        {yuan2023scaling}
\bibfield{author}{\bibinfo{person}{Zheng Yuan}, \bibinfo{person}{Hongyi Yuan}, \bibinfo{person}{Chengpeng Li}, \bibinfo{person}{Guanting Dong}, \bibinfo{person}{Chuanqi Tan}, {and} \bibinfo{person}{Chang Zhou}.} \bibinfo{year}{2023}\natexlab{a}.
\newblock \showarticletitle{Scaling relationship on learning mathematical reasoning with large language models}.
\newblock \bibinfo{journal}{\emph{arXiv}} (\bibinfo{year}{2023}).
\newblock


\bibitem[Yuan et~al\mbox{.}(2023b)]%
        {yuan2023well}
\bibfield{author}{\bibinfo{person}{Zheng Yuan}, \bibinfo{person}{Hongyi Yuan}, \bibinfo{person}{Chuanqi Tan}, \bibinfo{person}{Wei Wang}, {and} \bibinfo{person}{Songfang Huang}.} \bibinfo{year}{2023}\natexlab{b}.
\newblock \showarticletitle{How well do Large Language Models perform in Arithmetic tasks?}
\newblock \bibinfo{journal}{\emph{arXiv}} (\bibinfo{year}{2023}).
\newblock


\bibitem[Yue et~al\mbox{.}(2024)]%
        {yue2024mmmumassivemultidisciplinemultimodal}
\bibfield{author}{\bibinfo{person}{Xiang Yue}, \bibinfo{person}{Yuansheng Ni}, \bibinfo{person}{Kai Zhang}, \bibinfo{person}{Tianyu Zheng}, \bibinfo{person}{Ruoqi Liu}, \bibinfo{person}{Ge Zhang}, \bibinfo{person}{Samuel Stevens}, \bibinfo{person}{Dongfu Jiang}, \bibinfo{person}{Weiming Ren}, \bibinfo{person}{Yuxuan Sun}, \bibinfo{person}{Cong Wei}, \bibinfo{person}{Botao Yu}, \bibinfo{person}{Ruibin Yuan}, \bibinfo{person}{Renliang Sun}, \bibinfo{person}{Ming Yin}, \bibinfo{person}{Boyuan Zheng}, \bibinfo{person}{Zhenzhu Yang}, \bibinfo{person}{Yibo Liu}, \bibinfo{person}{Wenhao Huang}, \bibinfo{person}{Huan Sun}, \bibinfo{person}{Yu Su}, {and} \bibinfo{person}{Wenhu Chen}.} \bibinfo{year}{2024}\natexlab{}.
\newblock \bibinfo{title}{MMMU: A Massive Multi-discipline Multimodal Understanding and Reasoning Benchmark for Expert AGI}.
\newblock
\newblock
\showeprint[arxiv]{2311.16502}


\bibitem[Yue et~al\mbox{.}(2023)]%
        {yue2023mammoth}
\bibfield{author}{\bibinfo{person}{Xiang Yue}, \bibinfo{person}{Xingwei Qu}, \bibinfo{person}{Ge Zhang}, \bibinfo{person}{Yao Fu}, \bibinfo{person}{Wenhao Huang}, \bibinfo{person}{Huan Sun}, \bibinfo{person}{Yu Su}, {and} \bibinfo{person}{Wenhu Chen}.} \bibinfo{year}{2023}\natexlab{}.
\newblock \showarticletitle{Mammoth: Building math generalist models through hybrid instruction tuning}.
\newblock \bibinfo{journal}{\emph{arXiv}} (\bibinfo{year}{2023}).
\newblock


\bibitem[Zelikman et~al\mbox{.}(2024)]%
        {zelikman2024quiet}
\bibfield{author}{\bibinfo{person}{Eric Zelikman}, \bibinfo{person}{Georges Harik}, \bibinfo{person}{Yijia Shao}, \bibinfo{person}{Varuna Jayasiri}, \bibinfo{person}{Nick Haber}, {and} \bibinfo{person}{Noah~D Goodman}.} \bibinfo{year}{2024}\natexlab{}.
\newblock \showarticletitle{Quiet-star: Language models can teach themselves to think before speaking}.
\newblock \bibinfo{journal}{\emph{arXiv}} (\bibinfo{year}{2024}).
\newblock


\bibitem[Zelikman et~al\mbox{.}(2022)]%
        {zelikman2022star}
\bibfield{author}{\bibinfo{person}{Eric Zelikman}, \bibinfo{person}{Yuhuai Wu}, \bibinfo{person}{Jesse Mu}, {and} \bibinfo{person}{Noah Goodman}.} \bibinfo{year}{2022}\natexlab{}.
\newblock \showarticletitle{Star: Bootstrapping reasoning with reasoning}.
\newblock \bibinfo{journal}{\emph{NeurIPS}}  \bibinfo{volume}{35} (\bibinfo{year}{2022}), \bibinfo{pages}{15476--15488}.
\newblock


\bibitem[Zeng et~al\mbox{.}(2022)]%
        {zeng2022socratic}
\bibfield{author}{\bibinfo{person}{Andy Zeng}, \bibinfo{person}{Maria Attarian}, \bibinfo{person}{Brian Ichter}, \bibinfo{person}{Krzysztof Choromanski}, \bibinfo{person}{Adrian Wong}, \bibinfo{person}{Stefan Welker}, \bibinfo{person}{Federico Tombari}, \bibinfo{person}{Aveek Purohit}, \bibinfo{person}{Michael Ryoo}, \bibinfo{person}{Vikas Sindhwani}, {et~al\mbox{.}}} \bibinfo{year}{2022}\natexlab{}.
\newblock \showarticletitle{Socratic models: Composing zero-shot multimodal reasoning with language}.
\newblock \bibinfo{journal}{\emph{arXiv}} (\bibinfo{year}{2022}).
\newblock


\bibitem[Zhang et~al\mbox{.}(2024b)]%
        {zhang2024accessing}
\bibfield{author}{\bibinfo{person}{Di Zhang}, \bibinfo{person}{Xiaoshui Huang}, \bibinfo{person}{Dongzhan Zhou}, \bibinfo{person}{Yuqiang Li}, {and} \bibinfo{person}{Wanli Ouyang}.} \bibinfo{year}{2024}\natexlab{b}.
\newblock \showarticletitle{Accessing gpt-4 level mathematical olympiad solutions via monte carlo tree self-refine with llama-3 8b}.
\newblock \bibinfo{journal}{\emph{arXiv}} (\bibinfo{year}{2024}).
\newblock


\bibitem[Zhang et~al\mbox{.}(2023b)]%
        {zhang2023interpretable}
\bibfield{author}{\bibinfo{person}{Mengxue Zhang}, \bibinfo{person}{Zichao Wang}, \bibinfo{person}{Zhichao Yang}, \bibinfo{person}{Weiqi Feng}, {and} \bibinfo{person}{Andrew Lan}.} \bibinfo{year}{2023}\natexlab{b}.
\newblock \showarticletitle{Interpretable math word problem solution generation via step-by-step planning}.
\newblock \bibinfo{journal}{\emph{arXiv}} (\bibinfo{year}{2023}).
\newblock


\bibitem[Zhang et~al\mbox{.}(2022)]%
        {Zhang2022}
\bibfield{author}{\bibinfo{person}{Ming-Liang Zhang}, \bibinfo{person}{Fei Yin}, \bibinfo{person}{Yi-Han Hao}, {and} \bibinfo{person}{Cheng-Lin Liu}.} \bibinfo{year}{2022}\natexlab{}.
\newblock \showarticletitle{Plane Geometry Diagram Parsing}. In \bibinfo{booktitle}{\emph{IJCAI}}. \bibinfo{pages}{1636--1643}.
\newblock


\bibitem[Zhang et~al\mbox{.}(2023d)]%
        {Zhang2023PGPS}
\bibfield{author}{\bibinfo{person}{Ming-Liang Zhang}, \bibinfo{person}{Fei Yin}, {and} \bibinfo{person}{Cheng-Lin Liu}.} \bibinfo{year}{2023}\natexlab{d}.
\newblock \showarticletitle{A multi-modal neural geometric solver with textual clauses parsed from diagram}. In \bibinfo{booktitle}{\emph{IJCAI}}. \bibinfo{pages}{3374--3382}.
\newblock


\bibitem[Zhang et~al\mbox{.}(2025)]%
        {zhang2025mathverse}
\bibfield{author}{\bibinfo{person}{Renrui Zhang}, \bibinfo{person}{Dongzhi Jiang}, \bibinfo{person}{Yichi Zhang}, \bibinfo{person}{Haokun Lin}, \bibinfo{person}{Ziyu Guo}, \bibinfo{person}{Pengshuo Qiu}, \bibinfo{person}{Aojun Zhou}, \bibinfo{person}{Pan Lu}, \bibinfo{person}{Kai-Wei Chang}, \bibinfo{person}{Yu Qiao}, {et~al\mbox{.}}} \bibinfo{year}{2025}\natexlab{}.
\newblock \showarticletitle{Mathverse: Does your multi-modal llm truly see the diagrams in visual math problems?}. In \bibinfo{booktitle}{\emph{ECCV}}. \bibinfo{pages}{169--186}.
\newblock


\bibitem[Zhang et~al\mbox{.}(2023a)]%
        {zhang_instruction_2023}
\bibfield{author}{\bibinfo{person}{Shengyu Zhang}, \bibinfo{person}{Linfeng Dong}, \bibinfo{person}{Xiaoya Li}, \bibinfo{person}{Sen Zhang}, \bibinfo{person}{Xiaofei Sun}, \bibinfo{person}{Shuhe Wang}, \bibinfo{person}{Jiwei Li}, \bibinfo{person}{Runyi Hu}, \bibinfo{person}{Tianwei Zhang}, \bibinfo{person}{Fei Wu}, {et~al\mbox{.}}} \bibinfo{year}{2023}\natexlab{a}.
\newblock \showarticletitle{Instruction tuning for large language models: A survey}.
\newblock \bibinfo{journal}{\emph{arXiv}} (\bibinfo{year}{2023}).
\newblock


\bibitem[Zhang et~al\mbox{.}(2020a)]%
        {zhangLanguageEmbeddingsCapture2020}
\bibfield{author}{\bibinfo{person}{Xikun Zhang}, \bibinfo{person}{Deepak Ramachandran}, \bibinfo{person}{Ian Tenney}, \bibinfo{person}{Yanai Elazar}, {and} \bibinfo{person}{Dan Roth}.} \bibinfo{year}{2020}\natexlab{a}.
\newblock \showarticletitle{Do language embeddings capture scales?}. In \bibinfo{booktitle}{\emph{EMNLP}}. \bibinfo{pages}{4889–4896}.
\newblock


\bibitem[Zhang et~al\mbox{.}(2020b)]%
        {zhang2020language}
\bibfield{author}{\bibinfo{person}{Xikun Zhang}, \bibinfo{person}{Deepak Ramachandran}, \bibinfo{person}{Ian Tenney}, \bibinfo{person}{Yanai Elazar}, {and} \bibinfo{person}{Dan Roth}.} \bibinfo{year}{2020}\natexlab{b}.
\newblock \showarticletitle{Do Language Embeddings capture Scales?}. In \bibinfo{booktitle}{\emph{BlackboxNLP Workshop on Analyzing and Interpreting Neural Networks for NLP}}. \bibinfo{pages}{292--299}.
\newblock


\bibitem[Zhang et~al\mbox{.}(2023c)]%
        {zhang2023cumulative}
\bibfield{author}{\bibinfo{person}{Yifan Zhang}, \bibinfo{person}{Jingqin Yang}, \bibinfo{person}{Yang Yuan}, {and} \bibinfo{person}{Andrew Chi-Chih Yao}.} \bibinfo{year}{2023}\natexlab{c}.
\newblock \showarticletitle{Cumulative reasoning with large language models}.
\newblock \bibinfo{journal}{\emph{arXiv}} (\bibinfo{year}{2023}).
\newblock


\bibitem[Zhang et~al\mbox{.}(2024a)]%
        {zhang2024learn}
\bibfield{author}{\bibinfo{person}{Zhihan Zhang}, \bibinfo{person}{Tao Ge}, \bibinfo{person}{Zhenwen Liang}, \bibinfo{person}{Wenhao Yu}, \bibinfo{person}{Dian Yu}, \bibinfo{person}{Mengzhao Jia}, \bibinfo{person}{Dong Yu}, {and} \bibinfo{person}{Meng Jiang}.} \bibinfo{year}{2024}\natexlab{a}.
\newblock \showarticletitle{Learn Beyond The Answer: Training Language Models with Reflection for Mathematical Reasoning}.
\newblock \bibinfo{journal}{\emph{arXiv}} (\bibinfo{year}{2024}).
\newblock


\bibitem[Zhang et~al\mbox{.}(2023e)]%
        {zhang2023automatic}
\bibfield{author}{\bibinfo{person}{Zhuosheng Zhang}, \bibinfo{person}{Aston Zhang}, \bibinfo{person}{Mu Li}, {and} \bibinfo{person}{Alex Smola}.} \bibinfo{year}{2023}\natexlab{e}.
\newblock \showarticletitle{Automatic Chain of Thought Prompting in Large Language Models}. In \bibinfo{booktitle}{\emph{ICLR}}.
\newblock


\bibitem[Zhao et~al\mbox{.}(2024b)]%
        {zhao2024exploring}
\bibfield{author}{\bibinfo{person}{Jun Zhao}, \bibinfo{person}{Jingqi Tong}, \bibinfo{person}{Yurong Mou}, \bibinfo{person}{Ming Zhang}, \bibinfo{person}{Qi Zhang}, {and} \bibinfo{person}{Xuan-Jing Huang}.} \bibinfo{year}{2024}\natexlab{b}.
\newblock \showarticletitle{Exploring the Compositional Deficiency of Large Language Models in Mathematical Reasoning Through Trap Problems}. In \bibinfo{booktitle}{\emph{EMNLP}}. \bibinfo{pages}{16361--16376}.
\newblock


\bibitem[Zhao et~al\mbox{.}(2023a)]%
        {zhao2023verify}
\bibfield{author}{\bibinfo{person}{Ruochen Zhao}, \bibinfo{person}{Xingxuan Li}, \bibinfo{person}{Shafiq Joty}, \bibinfo{person}{Chengwei Qin}, {and} \bibinfo{person}{Lidong Bing}.} \bibinfo{year}{2023}\natexlab{a}.
\newblock \showarticletitle{Verify-and-edit: A knowledge-enhanced chain-of-thought framework}.
\newblock \bibinfo{journal}{\emph{arXiv}} (\bibinfo{year}{2023}).
\newblock


\bibitem[Zhao et~al\mbox{.}(2023b)]%
        {zhao2023survey}
\bibfield{author}{\bibinfo{person}{Wayne~Xin Zhao}, \bibinfo{person}{Kun Zhou}, \bibinfo{person}{Junyi Li}, \bibinfo{person}{Tianyi Tang}, \bibinfo{person}{Xiaolei Wang}, \bibinfo{person}{Yupeng Hou}, \bibinfo{person}{Yingqian Min}, \bibinfo{person}{Beichen Zhang}, \bibinfo{person}{Junjie Zhang}, \bibinfo{person}{Zican Dong}, {et~al\mbox{.}}} \bibinfo{year}{2023}\natexlab{b}.
\newblock \showarticletitle{A survey of large language models}.
\newblock \bibinfo{journal}{\emph{arXiv}} (\bibinfo{year}{2023}).
\newblock


\bibitem[Zhao et~al\mbox{.}(2022)]%
        {zhao2022multihiertt}
\bibfield{author}{\bibinfo{person}{Yilun Zhao}, \bibinfo{person}{Yunxiang Li}, \bibinfo{person}{Chenying Li}, {and} \bibinfo{person}{Rui Zhang}.} \bibinfo{year}{2022}\natexlab{}.
\newblock \showarticletitle{MultiHiertt: Numerical Reasoning over Multi Hierarchical Tabular and Textual Data}. In \bibinfo{booktitle}{\emph{ACL}}. \bibinfo{pages}{6588--6600}.
\newblock


\bibitem[Zhao et~al\mbox{.}(2024c)]%
        {zhao2024marco}
\bibfield{author}{\bibinfo{person}{Yu Zhao}, \bibinfo{person}{Huifeng Yin}, \bibinfo{person}{Bo Zeng}, \bibinfo{person}{Hao Wang}, \bibinfo{person}{Tianqi Shi}, \bibinfo{person}{Chenyang Lyu}, \bibinfo{person}{Longyue Wang}, \bibinfo{person}{Weihua Luo}, {and} \bibinfo{person}{Kaifu Zhang}.} \bibinfo{year}{2024}\natexlab{c}.
\newblock \showarticletitle{Marco-o1: Towards open reasoning models for open-ended solutions}.
\newblock \bibinfo{journal}{\emph{arXiv}} (\bibinfo{year}{2024}).
\newblock


\bibitem[Zhao et~al\mbox{.}(2024a)]%
        {zhao2024stepwise}
\bibfield{author}{\bibinfo{person}{Zilong Zhao}, \bibinfo{person}{Yao Rong}, \bibinfo{person}{Dongyang Guo}, \bibinfo{person}{Emek G{\"o}zl{\"u}kl{\"u}}, \bibinfo{person}{Emir G{\"u}lboy}, {and} \bibinfo{person}{Enkelejda Kasneci}.} \bibinfo{year}{2024}\natexlab{a}.
\newblock \showarticletitle{Stepwise Self-Consistent Mathematical Reasoning with Large Language Models}.
\newblock \bibinfo{journal}{\emph{arXiv}} (\bibinfo{year}{2024}).
\newblock


\bibitem[Zheng et~al\mbox{.}(2023)]%
        {zheng2021minif2f}
\bibfield{author}{\bibinfo{person}{Kunhao Zheng}, \bibinfo{person}{Jesse~Michael Han}, {and} \bibinfo{person}{Stanislas Polu}.} \bibinfo{year}{2023}\natexlab{}.
\newblock \showarticletitle{miniF2F: a cross-system benchmark for formal Olympiad-level mathematics}. In \bibinfo{booktitle}{\emph{ICLR}}.
\newblock


\bibitem[Zhong et~al\mbox{.}(2023)]%
        {zhong2023agieval}
\bibfield{author}{\bibinfo{person}{Wanjun Zhong}, \bibinfo{person}{Ruixiang Cui}, \bibinfo{person}{Yiduo Guo}, \bibinfo{person}{Yaobo Liang}, \bibinfo{person}{Shuai Lu}, \bibinfo{person}{Yanlin Wang}, \bibinfo{person}{Amin Saied}, \bibinfo{person}{Weizhu Chen}, {and} \bibinfo{person}{Nan Duan}.} \bibinfo{year}{2023}\natexlab{}.
\newblock \bibinfo{title}{AGIEval: A Human-Centric Benchmark for Evaluating Foundation Models}.
\newblock
\newblock
\showeprint[arxiv]{2304.06364}~[cs.CL]


\bibitem[Zhou et~al\mbox{.}(2023d)]%
        {zhou2023solving}
\bibfield{author}{\bibinfo{person}{Aojun Zhou}, \bibinfo{person}{Ke Wang}, \bibinfo{person}{Zimu Lu}, \bibinfo{person}{Weikang Shi}, \bibinfo{person}{Sichun Luo}, \bibinfo{person}{Zipeng Qin}, \bibinfo{person}{Shaoqing Lu}, \bibinfo{person}{Anya Jia}, \bibinfo{person}{Linqi Song}, \bibinfo{person}{Mingjie Zhan}, {et~al\mbox{.}}} \bibinfo{year}{2023}\natexlab{d}.
\newblock \showarticletitle{Solving challenging math word problems using gpt-4 code interpreter with code-based self-verification}.
\newblock \bibinfo{journal}{\emph{arXiv}} (\bibinfo{year}{2023}).
\newblock


\bibitem[Zhou et~al\mbox{.}(2023e)]%
        {zhou2023language}
\bibfield{author}{\bibinfo{person}{Andy Zhou}, \bibinfo{person}{Kai Yan}, \bibinfo{person}{Michal Shlapentokh-Rothman}, \bibinfo{person}{Haohan Wang}, {and} \bibinfo{person}{Yu-Xiong Wang}.} \bibinfo{year}{2023}\natexlab{e}.
\newblock \showarticletitle{Language agent tree search unifies reasoning acting and planning in language models}.
\newblock \bibinfo{journal}{\emph{arXiv}} (\bibinfo{year}{2023}).
\newblock


\bibitem[Zhou et~al\mbox{.}(2022)]%
        {zhouTeachingAlgorithmicReasoning2022}
\bibfield{author}{\bibinfo{person}{Hattie Zhou}, \bibinfo{person}{Azade Nova}, \bibinfo{person}{Hugo Larochelle}, \bibinfo{person}{Aaron Courville}, \bibinfo{person}{Behnam Neyshabur}, {and} \bibinfo{person}{Hanie Sedghi}.} \bibinfo{year}{2022}\natexlab{}.
\newblock \bibinfo{title}{Teaching Algorithmic Reasoning via In-Context Learning}.
\newblock
\newblock


\bibitem[Zhou et~al\mbox{.}(2015)]%
        {zhou2015learn}
\bibfield{author}{\bibinfo{person}{Lipu Zhou}, \bibinfo{person}{Shuaixiang Dai}, {and} \bibinfo{person}{Liwei Chen}.} \bibinfo{year}{2015}\natexlab{}.
\newblock \showarticletitle{Learn to solve algebra word problems using quadratic programming}. In \bibinfo{booktitle}{\emph{EMNLP}}. \bibinfo{pages}{817--822}.
\newblock


\bibitem[Zhou et~al\mbox{.}(2023a)]%
        {DBLP:conf/icml/ZhouJWCS23}
\bibfield{author}{\bibinfo{person}{Wangchunshu Zhou}, \bibinfo{person}{Yuchen~Eleanor Jiang}, \bibinfo{person}{Ethan Wilcox}, \bibinfo{person}{Ryan Cotterell}, {and} \bibinfo{person}{Mrinmaya Sachan}.} \bibinfo{year}{2023}\natexlab{a}.
\newblock \showarticletitle{Controlled Text Generation with Natural Language Instructions}. In \bibinfo{booktitle}{\emph{ICML}}, Vol.~\bibinfo{volume}{202}. \bibinfo{pages}{42602--42613}.
\newblock


\bibitem[Zhou et~al\mbox{.}(2023b)]%
        {zhou2023isr}
\bibfield{author}{\bibinfo{person}{Zhehua Zhou}, \bibinfo{person}{Jiayang Song}, \bibinfo{person}{Kunpeng Yao}, \bibinfo{person}{Zhan Shu}, {and} \bibinfo{person}{Lei Ma}.} \bibinfo{year}{2023}\natexlab{b}.
\newblock \showarticletitle{ISR-LLM: Iterative Self-Refined Large Language Model for Long-Horizon Sequential Task Planning}.
\newblock \bibinfo{journal}{\emph{arXiv}} (\bibinfo{year}{2023}).
\newblock


\bibitem[Zhou et~al\mbox{.}(2023c)]%
        {zhou2023mathattack}
\bibfield{author}{\bibinfo{person}{Zihao Zhou}, \bibinfo{person}{Qiufeng Wang}, \bibinfo{person}{Mingyu Jin}, \bibinfo{person}{Jie Yao}, \bibinfo{person}{Jianan Ye}, \bibinfo{person}{Wei Liu}, \bibinfo{person}{Wei Wang}, \bibinfo{person}{Xiaowei Huang}, {and} \bibinfo{person}{Kaizhu Huang}.} \bibinfo{year}{2023}\natexlab{c}.
\newblock \bibinfo{title}{MathAttack: Attacking Large Language Models Towards Math Solving Ability}.
\newblock
\newblock
\showeprint[arxiv]{2309.01686}~[cs.CL]


\bibitem[Zhu et~al\mbox{.}(2021)]%
        {zhu2021tat}
\bibfield{author}{\bibinfo{person}{Fengbin Zhu}, \bibinfo{person}{Wenqiang Lei}, \bibinfo{person}{Youcheng Huang}, \bibinfo{person}{Chao Wang}, \bibinfo{person}{Shuo Zhang}, \bibinfo{person}{Jiancheng Lv}, \bibinfo{person}{Fuli Feng}, {and} \bibinfo{person}{Tat-Seng Chua}.} \bibinfo{year}{2021}\natexlab{}.
\newblock \showarticletitle{TAT-QA: A Question Answering Benchmark on a Hybrid of Tabular and Textual Content in Finance}. In \bibinfo{booktitle}{\emph{ACL}}. \bibinfo{pages}{3277--3287}.
\newblock


\bibitem[Zhu et~al\mbox{.}(2022)]%
        {zhu2022solving}
\bibfield{author}{\bibinfo{person}{Xinyu Zhu}, \bibinfo{person}{Junjie Wang}, \bibinfo{person}{Lin Zhang}, \bibinfo{person}{Yuxiang Zhang}, \bibinfo{person}{Ruyi Gan}, \bibinfo{person}{Jiaxing Zhang}, {and} \bibinfo{person}{Yujiu Yang}.} \bibinfo{year}{2022}\natexlab{}.
\newblock \showarticletitle{Solving math word problems via cooperative reasoning induced language models}.
\newblock \bibinfo{journal}{\emph{arXiv}} (\bibinfo{year}{2022}).
\newblock


\end{thebibliography}
